\documentclass[conference]{IEEEtran}
\usepackage{times}

\usepackage[numbers,sort&compress]{natbib}
\usepackage{multicol}
\usepackage[bookmarks=true]{hyperref}
\usepackage{wrapfig}
\usepackage{url}            % simple URL typesetting
\usepackage{booktabs}       % professional-quality tables
\usepackage{amsfonts}       % blackboard math symbols
\usepackage{nicefrac}       % compact symbols for 1/2, etc.
\usepackage{microtype}      % microtypography
\usepackage{xcolor}         % colors
\usepackage{graphicx}
\usepackage{subcaption}
\usepackage{xspace}
\usepackage{algorithm}
\usepackage{algpseudocode}
\usepackage{amsmath}
\let\labelindent\relax % <-- Insert this line before enumitem
\usepackage{enumitem}
\usepackage{changepage} 
\usepackage{tcolorbox}
\usepackage{mdframed}

\usepackage{array}
\newcolumntype{L}[1]{>{\raggedright\arraybackslash}m{#1}}
\usepackage{multirow}
\usepackage{textcomp}
\usepackage[utf8]{inputenc} % allow utf-8 input
\usepackage[T1]{fontenc}    % use 8-bit T1 fonts
\usepackage{booktabs}
\usepackage{longtable}
\usepackage{stfloats}
\newcolumntype{M}[1]{>{\centering\arraybackslash}m{#1}}

\usepackage[symbol]{footmisc}

\definecolor{darkred}{rgb}{0.6, 0.1, 0.1}
\algrenewcommand{\algorithmiccomment}[1]{\textcolor{darkred}{\texttt{// #1}}}

\newcommand{\methodname}{\textsc{VLMgineer}\xspace}
\newcommand{\benchmarkname}{\textsc{RoboToolBench}\xspace}
\newcommand{\NumBenchmarkTasks}{12\xspace}
\newcommand{\PercentImprovementOverHuman}{64.7\%\xspace}
\newcommand{\PercentImprovementOverExisting}{24.3\%\xspace}

\newcommand{\appendixCaseStudy}{\textsc{A.1}\xspace}

\usepackage[addedmarkup=uline, defaultcolor=magenta, todonotes={textsize=scriptsize, textwidth=2cm}, authormarkuptext=name, commandnameprefix=always, xcolor, final]{changes}
\definechangesauthor[name={JS}, color=blue]{js}
\newcommand{\js}[1]{\chcomment[id=js]{#1}}
\newcommand{\jsadd}[1]{\chadded[id=js]{#1}}

\definechangesauthor[name={TL}, color=pink]{tl}

\definechangesauthor[name={JG}, color=orange]{jg}

\definechangesauthor[name={NF}, color=red]{nf}
\newcommand{\nf}[1]{\chcomment[id=nf]{#1}}
\newcommand{\nfadd}[1]{\chadded[id=nf]{#1}}

\begin{document}

\title{\methodname: \\Vision Language Models as Robotic Toolsmiths}

% You will get a Paper-ID when submitting a pdf file to the conference system
\author{George Jiayuan Gao$^{*}$, Tianyu Li$^{*}$, Junyao Shi, Yihan Li$^{\dag}$, Zizhe Zhang$^{\dag}$, Nadia Figueroa, Dinesh Jayaraman
\\
\href{https://vlmgineer.github.io/}{\textbf{vlmgineer.github.io/release}}
\\
\footnotemark{Email correspondence: \{gegao, tianyu\}@seas.upenn.edu. $^{*}$ \jsadd{and $^{\dagger}$ denote} equal contribution. }%
% \\[-4.0ex]
}

%\author{\authorblockN{Michael Shell}
%\authorblockA{School of Electrical and\\Computer Engineering\\
%Georgia Institute of Technology\\
%Atlanta, Georgia 30332--0250\\
%Email: mshell@ece.gatech.edu}
%\and
%\authorblockA{Twentieth Century Fox\\
%\and
%\authorblockN{James Kirk\\ and Montgomery Scott}
%\authorblockA{Starfleet Academy\\
%San Francisco, California 96678-2391\\

% avoiding spaces at the end of the author lines is not a problem with
% conference papers because we don't use \thanks or \IEEEmembership

%Homer Simpson\authorrefmark{2},
%Montgomery Scott\authorrefmark{3} and
%Eldon Tyrell\authorrefmark{4}}
%Telephone: (800) 555--1212, Fax: (888) 555--1212}
%\authorblockA{\authorrefmark{4}Tyrell Inc., 123 Replicant Street, Los Angeles, California 90210--4321}}

% \maketitle

\twocolumn[{%
    \renewcommand\twocolumn[1][]{#1}%
    \maketitle
    \vspace{-5pt}
    \begin{center}
            \includegraphics[width=1.0\linewidth]{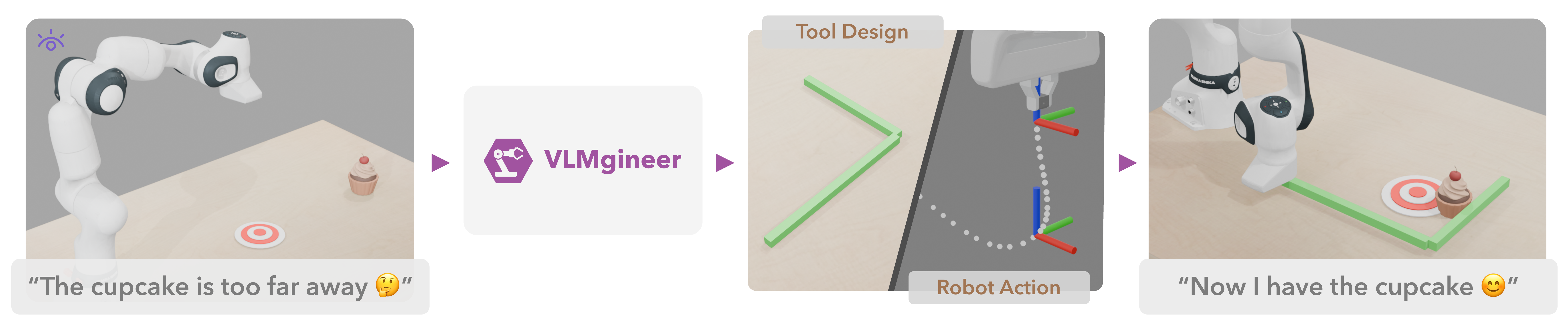}
            \captionof{figure}{Given a manipulation task that lies outside the robot’s capabilities, \textbf{\methodname} first prompts a vision language model to generate a tool and action. We then employ evolutionary search in simulation to refine the tool’s geometry and synthesize the corresponding robot motion plan. Finally, the robot, equipped with the automatically designed tool, successfully completes the task.}\label{fig:concept}
            \vspace{0.5em}
    \end{center}}]

\begin{abstract}
Tool design and use reflect the ability to understand and manipulate the physical world through creativity, planning, and foresight. As such, these capabilities are often regarded as measurable indicators of intelligence across biological species. While much of today's research on robotic intelligence focuses on generating better controllers, inventing smarter tools offers a complementary form of physical intelligence: shifting the onus of problem-solving onto the tool's design. Given the vast and impressive common-sense, reasoning, and creative capabilities of today's foundation models, we investigate whether these models can provide
%the following question: can we use these models as 
useful priors to automatically design and effectively wield such tools? We present \methodname, a framework that harnesses the code generation abilities of vision language models (VLMs) together with evolutionary search to iteratively co-design physical tools and the action plans that operate them to perform a task. We evaluate \methodname on a diverse new benchmark of everyday manipulation scenarios that demand creative tool design and use. Across this suite, \methodname consistently discovers tools and policies that solve tasks more effectively and innovatively, transforming challenging robotics problems into straightforward executions. It also outperforms VLM-generated designs from human specifications and existing human-crafted tools for everyday tasks.
To facilitate future research on automated tool invention, we will release our benchmark and code. Project Website: \href{https://vlmgineer.github.io/release}{vlmgineer.github.io}.
\end{abstract}

\IEEEpeerreviewmaketitle

\section{Introduction}

Humans exhibit a remarkable ability to design and utilize tools, fundamentally extending their capabilities to accomplish tasks otherwise beyond their reach through creativity, planning, and foresight. This capacity for tool creation and usage represents one of our most distinctive cognitive adaptations, and therefore is widely regarded as a marker of cognitive complexity. 
% In principle, 

\jsadd{Much of today's robotics research concentrates on enabling complex robot motions to utilize standard pre-designed robot end-effectors or tools}~\cite{RoboCook2023,qitooluse2024,car2024platoplanningllmsaffordances,shaw2024bimanual,chen2024vegetablepeelingcasestudy}\nf{Do not cite papers on arxiv not peer-reviewed}. 
\jsadd{In this work, we pursue an alternative form of physical intelligence: inventing tools that simplify downstream control, thereby shifting the primary problem-solving burden from devising control strategies to designing the tool's geometry. This requires solving a complex co-design problem over tool designs and the motions that wield those tools. The search space is dauntingly large, and it is clear that the design procedure would benefit from suitable priors to guide this search. 
%Achieving comparable versatility in robots demands a coupled approach: the shape of a tool and the motions that wield it should be co-designed --- each constraining and enabling the other. 
}

\jsadd{One potential source of such priors might be large vision-language models (VLMs) pre-trained on web-scale data. State-of-the-art VLMs have been shown to possess impressive common-sense, reasoning, and creative abilities, alongside extraordinary capabilities in code generation, visual comprehension, and in-context learning.} When combined with evolutionary search methods, VLMs have successfully crafted human-level reward functions for reinforcement learning~\cite{yu2023language,ma2023eureka}, 3D graphics~\cite{huang2024blenderalchemy}, articulations of in-the-wild objects~\cite{le2024articulateanything}, intricate 3D sculptural designs~\cite{bloxnet2024}, and developing advanced algorithms to solve mathematics and science problems~\cite{romera2024funsearch,aglietti2024funbo,alphaevolve2025}.
% , and have even enabled large language models (LLMs) to see and hear without training~\cite{ashutosh2025llmsheartraining}. 
%\jsadd{These insights motivate us to ask: can today's VLMs provide the inductive biases needed to design \textit{more innovative and action-efficient} tools?}

%In the wake of these results, 
%we ask: 
So, \textit{can today's VLMs also guide the design of innovative and action-efficient physical tools for robots}? We introduce \methodname, a \jsadd{fully autonomous}
% \nf{Where do you show the general-purposeness? You only show results on a single robot for tabletop tasks. I would just keep "fully autonomous framework"}
framework that leverages VLMs to jointly evolve both tool design and manipulation strategies for robots. Our method demonstrates unprecedented efficacy in developing specialized tools for diverse manipulation tasks, through an evolutionary search process guided by VLM-generated tool geometries and action plans. Compared to prior more limited investigations of tool design\nf{Which one? add citations}, that largely consider parameter optimization for a manually designed parametric template, \methodname works off-the-shelf for new tasks without task-specific templates, prompts, or examples.
%\js{A slight nod to issues with prior work, can be taken out}.
% Furthermore, it does not require manual parameter specifications or human intervention for tool and action design. 
\jsadd{Furthermore, compared to prior works that use VLMs, \methodname does not require in-context few-shot examples.} To facilitate future research and benchmarking, we also introduce \benchmarkname, a comprehensive simulation suite comprising \NumBenchmarkTasks diverse robotic tool-use \nfadd{manipulation} tasks specifically designed to evaluate tool \nfadd{design} and policy optimization methods.

% In summary, we make the following contributions:
% \begin{itemize}[leftmargin=*]
%     \item \textbf{\methodname, a novel evolutionary optimization framework}
    % that leverages the creative capabilities of VLMs to simultaneously optimize both tool morphology and manipulation strategies for general-purpose robotic applications, eliminating manual parameter specifications and human intervention. \methodname 
    % that automatically discovers innovative tools to solve robotics task more efficiently.
    % , transforming challenging robotics problems into straightforward executions.
    % \item \textbf{\benchmarkname, a comprehensive simulation benchmark} consisting of \NumBenchmarkTasks robotic tool-use tasks designed explicitly for evaluating robotic tool and policy designs. 
    % To facilitate future research, we will open-source this benchmark.
    
    % \item \textbf{Significantly exceeds current state-of-the-art computational design approaches}, achieving an average normalized improvement of \PercentImprovementOverComputational.
    % \item \textbf{Readily incorporates human preferences and feedback without requiring gradient updates}, enabling iterative improvements that closely align robotic designs and behaviors with nuanced human preferences.
% \end{itemize}
% \jsadd{Junyao: should we also stress somewhere that we don't need any in-context few shot examples, unlike many previous methods that use VLMs.} \nf{Yes, you can mention this above before "To facilitate future research...".. like "Furthermore, compared to prior works that use VLMs, \methodname does not require in-context ..."}
% \\
\textbf{Our fully autonomous approach demonstrates superior task performance over designs generated with human specifications and human-crafted everyday tools.} When evaluated on \benchmarkname, \methodname achieves an average normalized improvement of \nfadd{\PercentImprovementOverHuman} over VLM-generated designs from human language specifications and outperforms existing human-crafted tools by an average normalized improvement of \nfadd{\PercentImprovementOverExisting}.
% Our experimental results across four distinct manipulation tasks (GetCube, LiftPlate, MoveBall, and ScoopSpheres) demonstrate that \textsc{VLMgineer} significantly outperforms both default gripper configurations and human-specified tool designs. Quantitative evaluation shows our approach achieving near-perfect task objective scores (0.9579-0.9998) compared to human baselines (0.417-0.9958) while maintaining competitive efficiency in terms of distance traversed. Furthermore, our evolutionary search mechanism consistently improves upon initially sampled designs, demonstrating the value of iterative refinement in tool-action co-design.
Our results serve to validate both the physical design intelligence enshrined in VLMs pre-trained on web-scale data, and also present the promise of more adaptable and capable robotics systems that can ingeniously create and use tools.
%By bridging the gap between tool design and policy learning, our work presents the promise of more adaptable and capable robotic systems that can ingeniously create and use tools. 

%This form of ``mechanical intelligence'' not only enhances the practical utility of robots but also provides insights into the co-evolution of morphological and behavioral adaptations in biological systems.

\section{Related Work}
\label{sec:related}

\textbf{\jsadd{Task-Specific}
%\js{Adding this here for clarification, @dinesh do you like it?} 
Computational Agent and Tool Design.} 
Previous research has extensively investigated methods for optimizing robot morphology, end-effectors, and tool designs for robot manipulation through various computational approaches, ranging from model-based optimization~\cite{allen2022physical},
reinforcement learning (RL)~\cite{li2021learning,guolearning,he2024morph}, data-driven generative models~\cite{wu2019imagine,Fit2Form2020,DGDM2024,liu2024paperbot}, and differentiable simulation~\cite{li2023learning}. 
Others have explored robot design for locomotion
using evolutionary algorithms~\cite{jelisavcic2019lamarckian,hejna2021task,walker2021evolution,sims2023evolving,dong2023sard,dong2023herd},
stochastic optimization~\cite{exarchos2022task}, and graph search~\cite{10.1145/3414685.3417831}.
% Manual pre-definition of optimization parameters.
% Limited in range of tasks performed: all. Usually one or two tasks.
% Does not learn policy: all. Use pre-defined trajectories or policies. 
However, these existing approaches typically require manual \jsadd{task-specific} pre-definition of a handful of optimization parameters,
% restrict optimization to a narrow set of tasks (often just one or two),
rely on fixed trajectories or pre-defined control policies, and tend to suffer from low sample efficiency.
In contrast, we introduce a VLM-driven approach that simultaneously optimizes both tool design and manipulation strategy, 
% policies,
enabling generalization across diverse manipulation tasks without requiring manual parameter specifications.
% and significantly improving sample efficiency compared to prior methods.

% \subsection{Model-free policy design}
% \jsadd{Is this section necessary? If so, I can flesh it out more.} 

% \jsadd{\textbf{RL to generate actions.} I can cite PPO, Eureka.}

% \jsadd{\textbf{LLM to generate actions.} I can cite as zero-shot trajectory generator, code as policies, Keypoint Action Tokens, MOKA, ReKep, ...}

% \jsadd{Ours: Prompt a VLM to generate 7D trajectories (6dof + gripper).}

\textbf{Robot Learning for Tool-Based Tasks.} To learn effective tool usage, some have employed learned or simulated dynamics models for tool manipulation optimization~\cite{xie2019improvisation,allen2020rapid,girdhar2020forward,lin2022diffskill,lin2022planning}. 
Another prevalent approach involves learning tool and object affordances — understanding the functions of objects and tool-object interactions~\cite{fang2020learning,qin2020keto,brawer2020causal,xu2021deep,noguchi2021tool,RoboCook2023}. Recently, large language models have been leveraged for creative tool use~\cite{xu2023creative}. 
While these methods all assume that suitable tools already exist in the environment, we instead address the more challenging scenario where a general-purpose robot must concurrently 
invent the right tool design and the appropriate way to use it for a task.
%optimize both the tool's design and its manipulation strategies.

\textbf{Joint Optimization of Morphology and Control.} 
% Jointly addressing tool design and control problems has often involved formulating nonlinear programs to solve task and motion planning (TAMP) given predefined design parameter space, which are particularly effective for sequential manipulation over extended horizons
A common strategy for jointly addressing tool design and manipulation problems is to formulate them as nonlinear programs within a predefined parameter space including both design and manipulation variables, a method that has demonstrated particular effectiveness in sequential manipulation tasks involving extended horizons~\cite{toussaint2018differentiable,toussaint2021co}. However, given our objective to deploy \methodname in any arbitrary environment without manual specification of design parameters, we rely on the underexplored
% underestimated 
physical creativity of VLMs. Approaches using RL~\cite{wang2023cuco,wang2023preco,pmlr-v100-luck20a,yuan2021transform2act}, gradient-based optimization~\cite{spielberg2019learning}, Bayesian optimization, evolutionary algorithms~\cite{cheney2018scalable,mertan2024investigatingprematureconvergencecooptimization,ringel2025text2robot}, or a combination of them~\cite{liao2019data,schaff2019jointly,ha2019reinforcement,bhatia2021evolution,pathak2019learning} have been proposed for joint morphology and control learning for robot locomotion tasks, in particular with soft or modular robots. Studies on joint robot or tool and policy design through RL~\cite{chen2020hardware,RoboticToolDesign2023}, differentiable simulation~\cite{DiffHand2021}, and model-based optimization~\cite{kawaharazuka2020tool} have also demonstrated effectiveness in tool manipulation.
% \js{Do I need to expand here? Describe individual methods in more details?}
However, since these methods still all require manual specifications of the design space, they require significant human efforts to scale beyond a few tasks. 

Recent work has explored LLM-aided evolutionary search for robot design \jsadd{in conjunction with RL-based policy optimization} in locomotion~\cite{qiu2024robomorph,chen2025largelanguagemodelsnatural,songlaser2025}, demonstrating the potential of using LLMs to unlock more performant robot design. Unlike prior work, our work targets open-world VLM-guided design of both tools and actions for manipulation without human-in-the-loop parameter specification. \textbf{\methodname leverages the surprising physical intuition and creativity of VLMs to automatically create design solutions using evolutionary search.} It can easily be scaled to a wide range of tasks, and it is \nfadd{much more efficient in terms of samples, time, and compute }\jsadd{than prior RL-based methods.}\nf{Much more efficient than what?}

% \jsadd{Note, I am still QUITE WORRIED about LASeR~\cite{songlaser2025}, 70 percent of this paper is describing a very general algorithm quite similar to \methodname. It only used soft robot as an experiment to validate their method. I can imagine angry NeurIPS reviewers blaming us for not having novelty...}

% \subsection{Evolutionary Search for Design}

% BloxNet~\cite{bloxnet2024} designs block-based structures using VLM-in-the-loop guided search.

% None of them use VLM to design robots.

% We present the first system that...

% \section{\nfadd{Problem Setting and} Background}
\begin{figure*}[h!]
    \centering
    \includegraphics[width=0.9\linewidth, trim=0.3in 1.1in 0.3in 1.2in, clip]{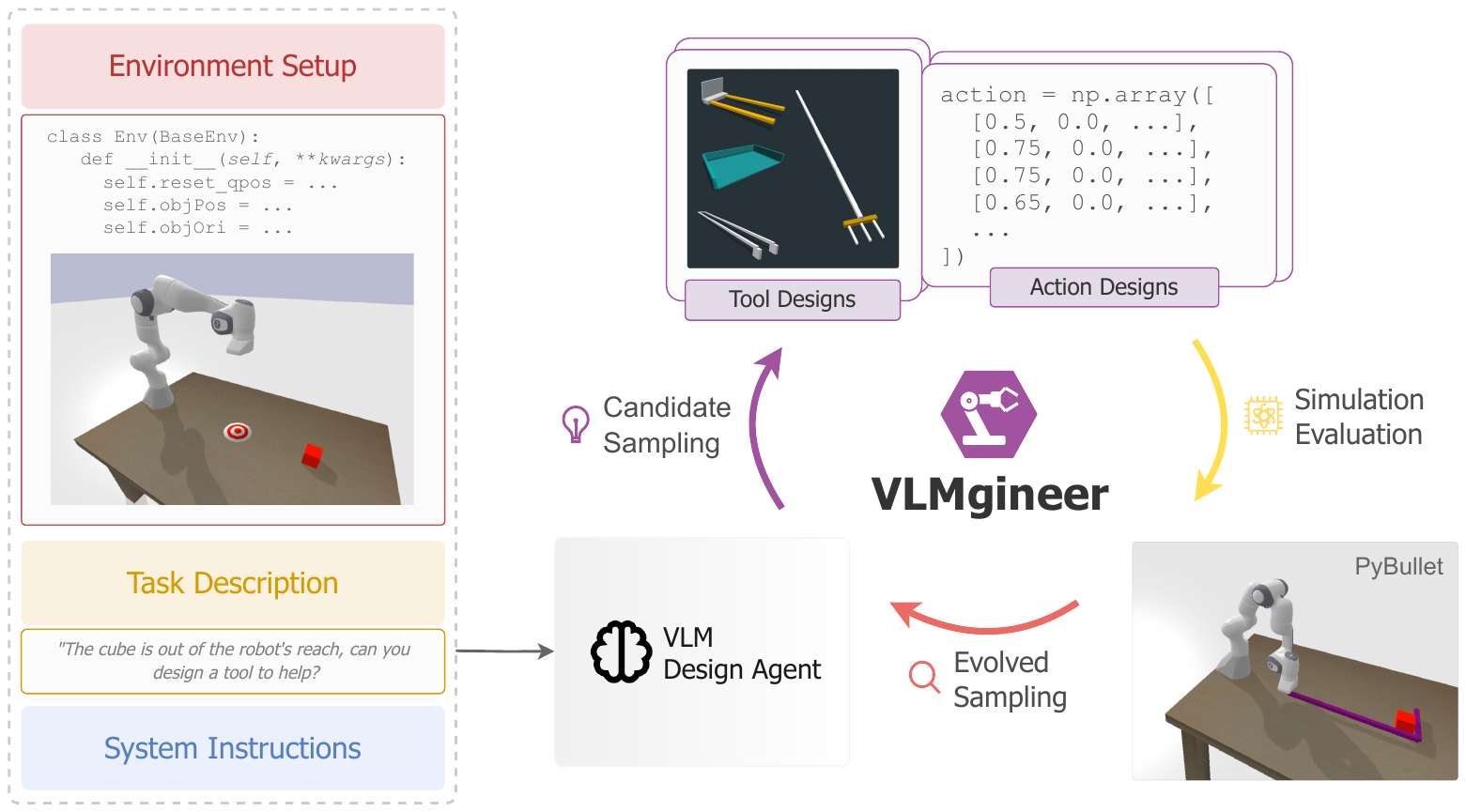}
    \caption{\jsadd{\methodname takes unmodified environment source code, environment image, environmental description, and task description as context to zero-shot generate tool and action designs from a VLM.  It then iteratively refines its tool and action designs through a loop of candidate sampling, simulation-based evaluation, and evolution improvement.}\label{fig:method}}
    \vspace{-10pt}
\end{figure*}

\section{Background}
\label{sec:background}

% \jsadd{Dinesh, should we add some transition here between related work and background, and if so, how?} \nfadd{Yes, you can start this section by clearly posing the problem you are trying to solve with \methodname first like: The goal of \methodname is to ... }

% \nfadd{Then say, at its core \methodname is an evolutionary search algorithm and transition into the next paragraph... this you can reduce and don't need to be so explicit citing all of the related works. }

The goal of \methodname is to jointly optimize both the physical design of tools and the manipulation strategies needed to accomplish a wide range of robotic tasks, without requiring human-specified design parameters or fixed control policies. At its core, \methodname is an evolutionary search algorithm, enhanced by the creative and commonsense reasoning capabilities of vision–language models, to efficiently discover novel tool–action pairs for challenging manipulation problems.

\textbf{Evolutionary Methods.} Evolutionary algorithms~\cite{langdon2013foundations, doncieux2015evolutionary} have a long-standing history in solving optimization problems, inspired by principles of biological evolution and natural selection. They are particularly effective in black-box optimization with vast optimization spaces, such as open-ended design\js{I'm not sure if this is the correct characterization, someone please check}\js{CITE}.
% \jsadd{These work range from ... to ... CITE.}
% There has been a long tradition of research on \textit{evolutionary algorithms}~\cite{langdon2013foundations,doncieux2015evolutionary} inspired by biological evolution used to solve optimization problems, particularly black-box optimization with vast design spaces such as open-ended design, by mimicking natural selection. 
At their core, these methods maintain a \textbf{population} of candidate solutions, which iteratively evolve through carefully designed mutation and crossover operators. 
% The core of evolutionary methods is a \textbf{pool of potential solutions} that is iteratively evolved using a set of \textbf{mutation and crossover operators}. 
Each iteration evaluates individuals against a \textbf{fitness function}, selecting those with higher fitness while discarding or replacing less successful candidates. 
% In each iteration, \textbf{a fitness function} is used to evaluate the individual quality of each solution, and only the ones with high fitness scores are ``selected''. 
To balance exploitation and exploration, \textbf{crossover} combines promising solutions into offspring, and \textbf{mutation} introduces novel variations.
% To encourage diversity and exploration, \textbf{the crossover operator} produces offspring among the selected candidates, and \textbf{the mutation operator} mutates them. 
% Individuals with lower fitness are then discarded or replaced by new individuals with higher fitness. 
Evolutionary algorithms have proven effective across diverse domains such as program synthesis\js{CITE}, symbolic regression\js{CITE}, algorithm discovery\js{CITE}, and even robot design\js{CITE}\js{Unaddressed Nadia comment}.
% \nfadd{At this point, it should be clear to the reader that \methodname is an evolutionary algorithm that will optimize tool design and actions jointly for a manipulation task contextualizing the "candidate solutions" as tool-action samples. Then highlight that: solving this evolutionary process is challenging due to what you say next:}
\setlength{\textfloatsep}{6pt}% Remove \textfloatsep
\begin{algorithm}
\caption{\methodname: Evolutionary Tool and Action Co-Design with VLMs}
\label{alg:vlmgineer}
\small
\begin{algorithmic}[1]
\Require Environment code $\mathcal{E}$, image render $I$, task description $d_{\text{task}}$, fitness function $\mathcal{F}$, initial prompt \textsc{prompt}, Vision-Language Model \textsc{VLM}
\State \textbf{Hyperparameters:} Number of evolution cycles $n$, population size $K$, top-$k$ selection threshold
% \Statex

\For{$n$ iterations do}
    \State \Comment{\textbf{Sample K designs}}
    \State $D_1, D_2, ..., D_K \sim \textsc{VLM}(\mathcal{E}, I, d_{\text{task}}, \textsc{prompt})$
    
    \State \Comment{\textbf{Evaluate design candidates}}
    \State $s_1 = \mathcal{F}(D_1),\cdots,s_K = \mathcal{F}(D_K)$

    \State \Comment{\textbf{Selection}}
    \State Select top-$k$ designs $\{D_{j_1}, ..., D_{j_k}\}$ with highest $s_j$

    \State \Comment{\textbf{Evolution}}
    \State \textsc{prompt} := \textsc{prompt} : $\textsc{Evolve\_Prompt}(\{D_{j_1},...,D_{j_k}\})$
\EndFor
\State \Return Final design $D^* = \arg\max_{D} \mathcal{F}(D)$ across all iterations
\end{algorithmic}
\end{algorithm}
Nevertheless, their reliance on handcrafted mutation and crossover operators remains a significant limitation—such operators are challenging to design and often inadequately capture essential domain-specific insights\js{Unaddressed Nadia comment}.
% However, a key limitation is the use of handwritten evolution and mutation operators, which are critical to the success of the algorithm yet hard to design and difficult to capture the relevant properties of the domain.

% \nfadd{The next paragraph should be re-written as follows (kind of inverting the flow of what you already have written and making it more concise), example:}

% \nfadd{To alleviate this, we take inspiration from \textbf{Large model-guided evolution} such as Eureka~\cite{ma2023eureka} which employs LLMs to.. }

\textbf{Large Model-Guided Evolution.} To improve the scalability, performance, and automation of evolutionary algorithms, recent work has integrated large models into the evolutionary process, automating initial population generation, mutation and crossover operations.
% To make evolutionary algorithms more scalable, performant, and automated, various previous work have explored using Large Models to automate the crossover and mutation process. 
Leveraging the extensive world knowledge and inductive biases inherent in large models allows for more efficient evolution of candidate solutions and also eliminates the necessity of manually defining allowed mutation operations. 
% By leveraging the Large Model's vast world knowledge and strong inductive biases, they evolve the candidate population more efficiently and without the need to pre-define a set of allowed mutation operations. 
Moreover, some approaches exploit the rich semantic understanding of large models to provide nuanced, semantic feedback beyond simple numerical fitness scores.
% Others also utilized the strong semantic understanding of large models to provide more rich and semantic feedback for each individual candidate aside from the fitness score. 
Specific implementations of these principles in evolutionary algorithms vary according to the domain.
% Each downstream application have their own domain-specific instantiations of these elementary concepts in evolutioanry algorithms. 
For instance, Eureka~\cite{ma2023eureka} employs large language models (LLMs) to guide evolutionary reward design in reinforcement learning.
% For example, Eureka~\cite{ma2023eureka} utiliezes LLM-guided evolutionary reward design for reinforcement learning. 

Eureka generates a \textit{population} of candidate reward functions directly from raw environment code, evaluates RL agents trained with these rewards using a task-specific \textit{fitness} function, and selects the best-performing candidates.
% Eureka's \textit{population} is LLM-generated samples of reward function code given raw environment code as context. 
% It uses a \textit{task fitness function} to evaluate RL agents trained with the generated reward functions, and select the high fitness candidates. 
Although it omits explicit crossover\js{Dinesh, is this true? I did not find it in the paper}, Eureka employs LLM-guided in-context reward \textit{mutation} by proposing an improved reward function from an existing one based on textual feedback.. 
% While it doesn't utilize crossover, it performs LLM-guided in-context reward \textit{mutation} by proposing a improved reward function from an existing one based on textual feedback. 
Drawing inspiration from these successes, we investigate whether vision–language models (VLMs) can similarly offer valuable inductive biases to guide the evolutionary design of robotic tools and manipulation actions.
% Inspired by these insights, we investigate if VLMs offer useful inductive biases to perform large model-guided evolution for robot tool and action design. 

% \vspace{-2.5pt}
\section{Method}
% \vspace{-2.5pt}
\methodname builds upon existing Large Model-guided evolutionary methods to achieve effective tool-action co-design. Specifically, \methodname comprises three algorithmic components: \textbf{(1) Population Generation}: We query the \textsc{VLM} to generate a diverse population of candidate designs $D_i$ (tool-action tuples) using the raw environment code $\mathcal{E}$, an image render $I$, task description $d_{\text{task}}$, and an initial prompt \textsc{prompt} as context. \textbf{(2) Fitness Evaluation}: We assess each candidate design $D_i$ using task-specific fitness functions $\mathcal{F}$, retaining only the top-$k$ designs based on their computed rewards. \textbf{(3) Iterative Evolution}: We iteratively prompt the \textsc{VLM} to produce novel offspring designs through guided mutations and crossovers, progressively enhancing tool-action quality.

% \nfadd{Refer to figure and algo}
% \setlength{\textfloatsep}{5pt}% Remove \textfloatsep
% \begin{algorithm}[h]
% \caption{\methodname: Evolutionary Tool and Action Design with VLMs}
% \label{alg:vlmgineer}
% \small
% \begin{algorithmic}[1]
% \Require Environment code $\mathcal{E}$, image render $I$, task description $d_{\text{task}}$, fitness function $\mathcal{F}$, initial prompt \textsc{prompt}, Vision-Language Model \textsc{VLM}
% \State \textbf{Hyperparameters:} Number of evolution cycles $n$, population size $K$, top-$k$ selection threshold
% % \Statex

% \For{$n$ iterations do}
%     \State \Comment{\textbf{Sample K designs}}
%     \State $D_1, D_2, ..., D_K \sim \textsc{VLM}(\mathcal{E}, I, d_{\text{task}}, \textsc{prompt})$
    
%     \State \Comment{\textbf{Evaluate design candidates}}
%     \State $s_1 = \mathcal{F}(D_1),\cdots,s_K = \mathcal{F}(D_K)$

%     \State \Comment{\textbf{Selection}}
%     \State Select top-$k$ designs $\{D_{j_1}, ..., D_{j_k}\}$ with highest $s_j$

%     \State \Comment{\textbf{Design Reflection}}
%     \State \textsc{prompt} :  = \textsc{prompt} : $\textsc{Evolution Prompt}(\{D_{j_1}, ..., D_{j_k}\})$
% \EndFor

% \State \Return Final design $D^* = \arg\max_{D} \mathcal{F}(D)$ across all iterations
% \end{algorithmic}
% \end{algorithm}

\begin{figure*}[t!]
    \centering
    \includegraphics[width=1.0\linewidth,trim={5cm 3cm 5cm 0},clip]{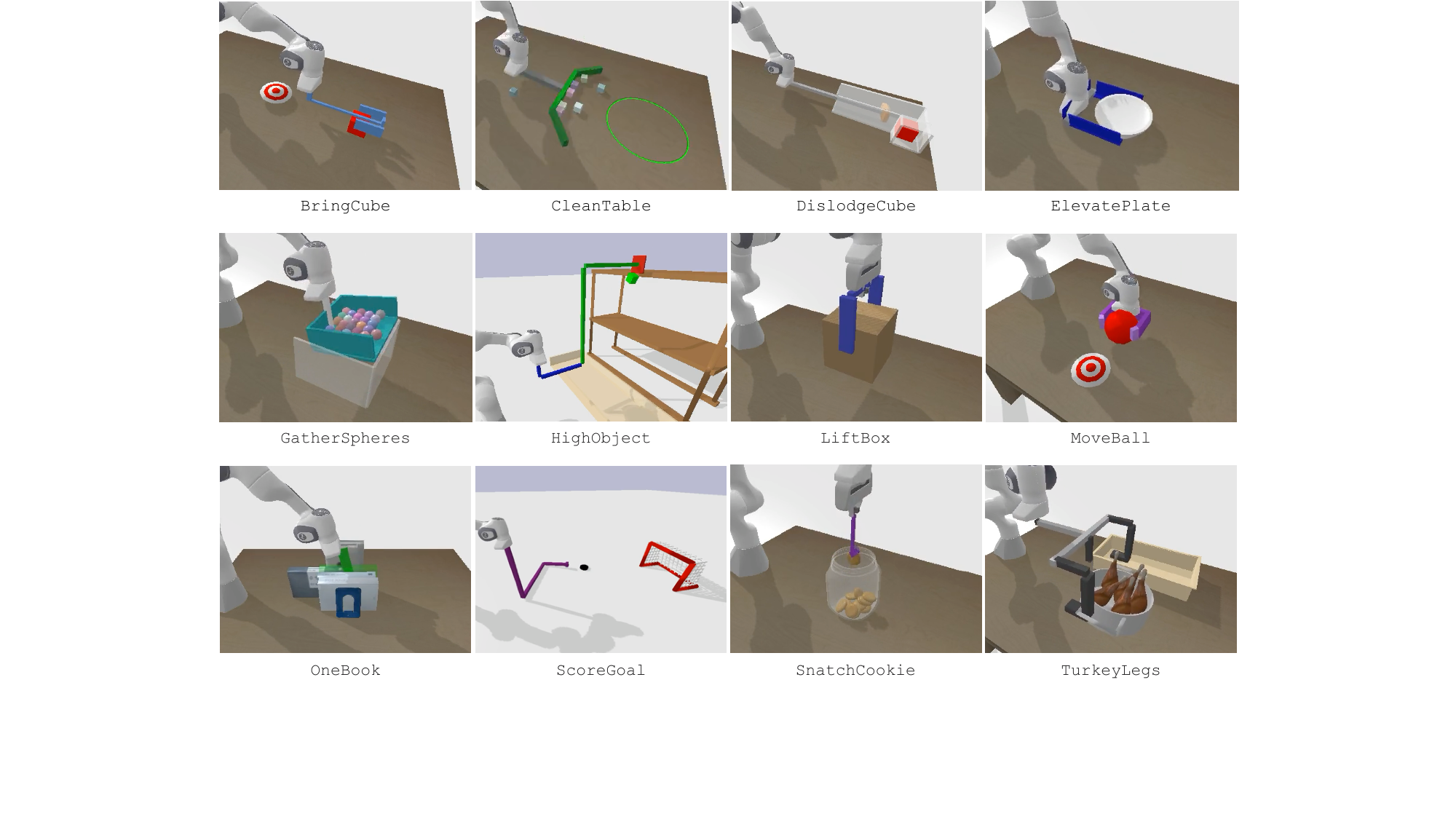}
    \caption{\methodname produces innovative tool designs and their corresponding actions across \NumBenchmarkTasks diverse tasks in \benchmarkname that are challenging to perform using a general-purpose robot arm and gripper.}
    \label{fig:benchmark}
\end{figure*}

\textbf{Joint Tool and Action Candidate Sampling.} While previous approaches of large model-guided evolutionary robot design~\cite{qiu2024robomorph,songlaser2025} typically optimize robot morphology alone, relegating action or control optimization to a subsequent evaluation stage. our approach prompts the VLM to simultaneously generate paired tool designs and corresponding action strategies in a single inference step.
% Our key insight behind 
Joint tool-action sampling allows for a tighter coupling between tools and their associated actions. Rather than sequentially optimizing the tool geometry first and then actions afterward, simultaneous optimization leverages the VLM's inductive biases to smoothly navigate the joint tool–action design space towards the Pareto frontier. Concretely, within each evolution cycle, \methodname prompts the VLM to propose $n$ distinct tool designs along with $m$ candidate action plans per tool, resulting in $n \times m$ total tool-action pairs. This corresponds to a kind of crude VLM-guided policy optimization, which merely selects the best among the $m$ generated action plans. Compared to policy optimization via RL~\cite{ma2023eureka,songlaser2025,qiu2024robomorph}, our action sampling approach, albeit simple, significantly accelerates iteration cycles and reduces computational overhead by exploiting the insight that appropriately designed tools inherently simplify and enhance action plans.

\begin{figure*}[h!]
    \centering
    \includegraphics[width=1.0\linewidth]{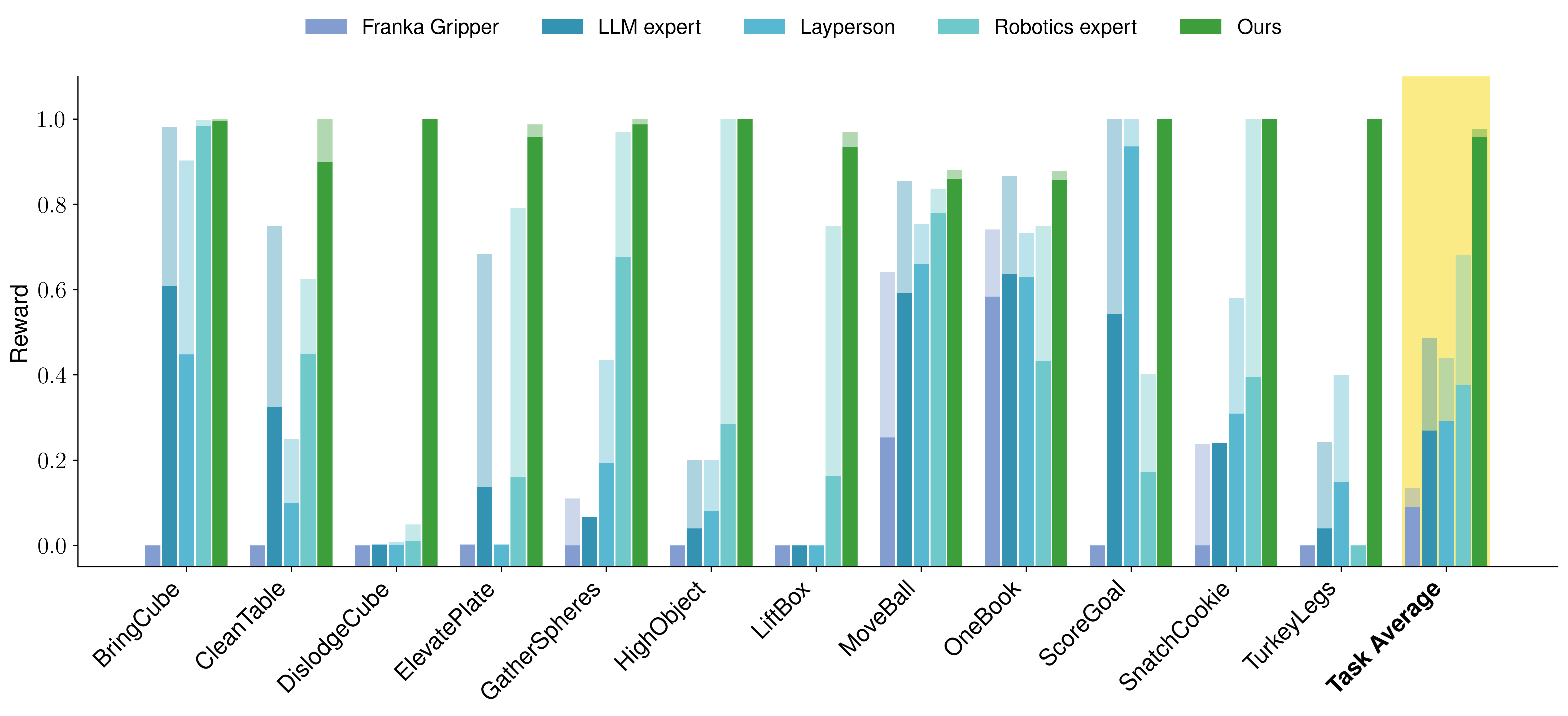}
    \caption{This figure compares the reward of Franka Gripper experiments, 3 Human Prompt experiments, and experiments on our proposed method across 12 tasks. For every method, the bars with the original dark color in the legend indicate the \textit{average} reward of the five runs, while the bars with a paler color visible above them indicate the \textit{best} reward over those runs.}
    \label{fig:best_and_average_bar}
\end{figure*}

\textbf{Specification of How to Do Crossover And Mutation.} A critical part of how \methodname enables effective tool design evolution is the utilization of \textit{inductive in-context \textbf{crossover and mutation}}. We define inductive in-context crossover and mutation as the process of prompting VLMs to introduce random, free-form tool mutations and crossovers, conditioned on previous elite tool candidates, and guided by the model's learned inductive biases for producing better task-solving tools. We use the prompt below to perform inductive in-context crossover and mutation\js{Is showing full prompt necessary? candidate for condensing and cutting?}: \textit{Your design decision is part of a genetic algorithm for tool creation, where each new design is produced either by mutation—changing exactly one aspect (e.g., adjusting a component’s dimension or adding/removing a component)—or by crossover, combining elements from two existing designs. All resulting mutations and crossovers should plausibly enhance task success while preserving design diversity.}

\textbf{Tool Representation Format.} Selecting an appropriate representation for tools—balancing abstraction, design flexibility, and manufacturability—is critical for effective optimization. Prior works have represented objects and tools as meshes~\cite{nair2020toolmacgyvering}, CAD~\cite{thomas2018learningroboticassemblycad}, or blocks~\cite{bloxnet2024}. These representations, however, either introduce excessive complexity and optimization challenges or lack sufficient expressiveness. Inspired by prior work~\cite{le2024articulateanything}, we represent tools in Unified Robot Description Format (URDF). The structured, modular nature of URDF, analogous to code blocks, aligns seamlessly with vision–language models' (VLMs) strengths in code understanding and generation. Concretely, we prompt the VLM to generate URDF-defined tool designs as modular blocks that can be directly integrated into a designated end-effector link of the robot model.

\textbf{Action Representation Format.}\js{Should this be a separate subsection? Or is best to not highlight our simplistic action representation.} 
Building on recent work that leverages VLMs for action generation~\cite{dipalo2024kat, yin2024roboprompt}, we prompt the model to explicitly output action sequences in the form of an $N \times 7$ array, where $N$ denotes the number of waypoints. Each row encodes a 6-DoF pose for the robot end-effector, along with a gripper open/close command.

\section{Robot Tool Design Benchmark}
\label{sec:tool_benchmark}

We propose a comprehensive simulation benchmark  \textbf{\benchmarkname} 
designed explicitly for evaluating robotic tool and policy design. \benchmarkname comprises 12 object manipulation tasks designed to be challenging for the base robot arm to complete without any tools. These task environments are visualized in Fig.~\ref{fig:benchmark}. For several tasks (\texttt{BringCube}, \texttt{CleanTable}, \texttt{GatherSpheres}, \texttt{ScoreGoal}), we took inspiration from the subset of RLBench~\cite{james2019rlbench} tasks that involve tool use --- note, however, that we expect that automated tool design will \textit{replace and improve} the original tools from RLBench. Several other tasks (\texttt{HighObject}, \texttt{ElevatePlate}) are inspired by prior works in computational co-design~\cite{RoboticToolDesign2023} that study task-specific design parameter optimization as discussed in Sec~\ref{sec:related}. Still more task environments are inspired by everyday home scenarios (\texttt{LiftBox}, \texttt{MoveBall}, \texttt{OneBook}, \texttt{SnatchCookie}, \texttt{TurkeyLegs}). Finally, \texttt{DislodgeCube} is inspired by a tool design behavior previously observed in the Caledonian crow~\cite{Jacobs2016-st}, which used tools to retrieve objects in confined spaces. 
We adopt the Franka Panda robot arm as the standard morphology to attach tools to, and implement our environments using PyBullet~\cite{benelot2018}.

\section{Evaluation}
We begin this section with an introduction to our experimental goals and setups, and then analyze the results of our comparison and ablation studies in detail. Our experiments are designed to address the following
questions:
% To showcase \methodname's ability to generate creative and effective tools and usage actions, we evaluate it on our proposed simulation benchmark (Section~\ref{sec:tool_benchmark}) and compare its performance against no tool scenarios, human-specified tool-action designs, and existing human-crafted tools for everyday 
% tasks. Through this, we aim to answer the following questions:
% \jsadd{\textbf{Dinesh:} Don't frame the paper as comparison against baseline, but more like objectively, VLMs can design good tools. All baselines are secondary.}
\textbf{Q1:} Can \methodname effectively discover innovative tools and ways to use them? 
\textbf{Q2:} How does \methodname compare to a human specifying tool designs to a VLM in natural language?
\textbf{Q3:} How do \methodname outputs improve over evolution iterations?
    % \item \jsadd{(Stretch Goal)} \textbf{Q4:} Can \methodname outperform state-of-the-art computational methods that learn to design and use tools for robotic manipulation? \jsadd{(This is~\cite{RoboticToolDesign2023})}
    % \item \jsadd{(Stretch Goal)} \textbf{Q5:} Can \methodname align with human-in-the-loop feedback? 

\paragraph{Baselines.}
%\label{sec:metrics}
%\textbf{Variety Methods.} 
To showcase \methodname's ability to generate creative and effective tools and usage actions, we compare our method with the following baselines: \textbf{(1) Franka Gripper}: We evaluate the performance of the vanilla Franka Panda two-finger gripper without additional tools on \benchmarkname to highlight the inherent limitations of the robot's default morphology; these tasks are after all explicitly designed to be very hard or impossible to perform without the right tools. We derive the no-tool action policy by prompting the VLM to follow an action-sampling procedure analogous to our proposed method, minus the use of any tools. \textbf{(2) Human Prompts}: For these baselines, we ask humans to specify a tool design to the VLM in natural language, following which it attempts to generate that tool and several action plans, as in our method. There is no evolutionary search. We evaluate on humans with varying expertise: "Robotics expert" (a graduate student researching robot learning), "LLM expert" (a graduate student researching LLMs), and "Layperson" (an undergraduate student with no relevant research experience). The procedure on the case study is in Appendix \appendixCaseStudy. \textbf{(3) RLBench Tools}: We evaluate four original tools from the tasks we adapted from RLBench, which are often natural everyday tools for the tasks considered.

\begin{figure*}[t!]
  \centering
  \begin{tabular}{@{}c@{\hspace{1ex}}ccc@{}}
    & {\large{Human}} & {\large{RLBench}} & {\large\textbf{Ours}} \\
    \raisebox{10ex}{\normalsize\rotatebox[origin=c]{90}{{\texttt{BringCube}}}} &
    \begin{subfigure}[b]{0.26\textwidth}
      \centering
      \includegraphics[width=\textwidth]{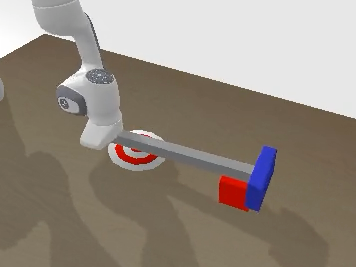}
    \end{subfigure}&
    \begin{subfigure}[b]{0.26\textwidth}
      \centering
      \includegraphics[width=\textwidth]{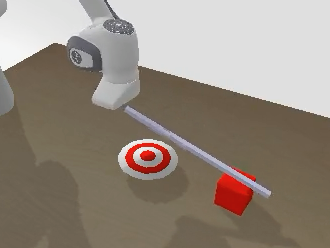}
    \end{subfigure}&
    \begin{subfigure}[b]{0.26\textwidth}
      \centering
      \includegraphics[width=\textwidth]{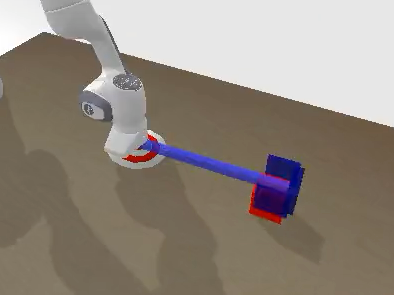}
    \end{subfigure}\\[2ex]
    \raisebox{10ex}{\normalsize\rotatebox[origin=c]{90}{{\texttt{ScoreGoal}}}} &
    \begin{subfigure}[b]{0.26\textwidth}
      \centering
      \includegraphics[width=\textwidth]{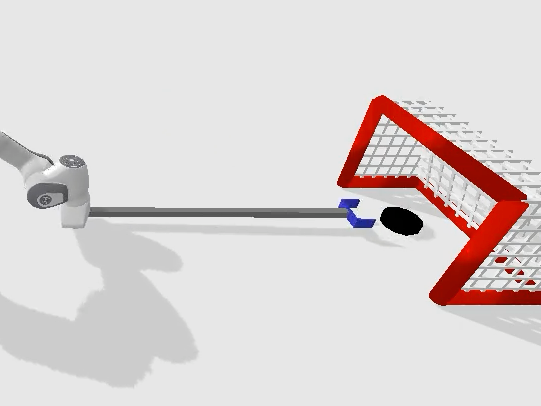}
    \end{subfigure}&
    \begin{subfigure}[b]{0.26\textwidth}
      \centering
      \includegraphics[width=\textwidth]{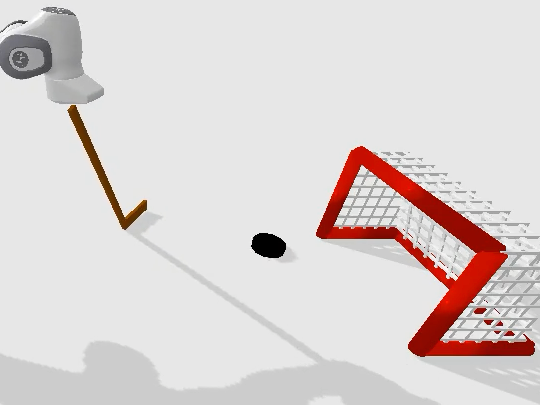}
    \end{subfigure}&
    \begin{subfigure}[b]{0.26\textwidth}
      \centering
      \includegraphics[width=\textwidth]{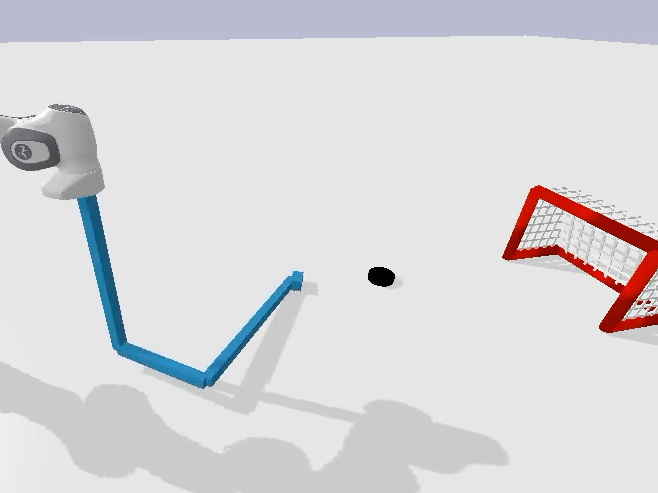}
    \end{subfigure}\\[2ex]
    \raisebox{11ex}{\normalsize\rotatebox[origin=c]{90}{{\texttt{GatherSpheres}}}} &
    \begin{subfigure}[b]{0.26\textwidth}
      \centering
      \includegraphics[width=\textwidth]{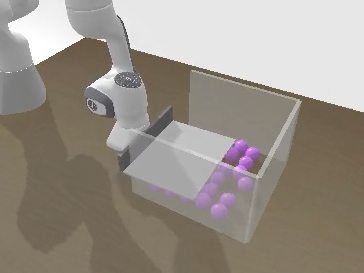}
    \end{subfigure}&
    \begin{subfigure}[b]{0.26\textwidth}
      \centering
      \includegraphics[width=\textwidth]{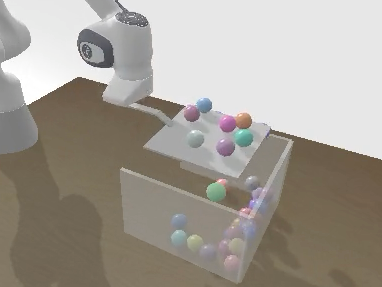}
    \end{subfigure}&
    \begin{subfigure}[b]{0.26\textwidth}
      \centering
      \includegraphics[width=\textwidth]{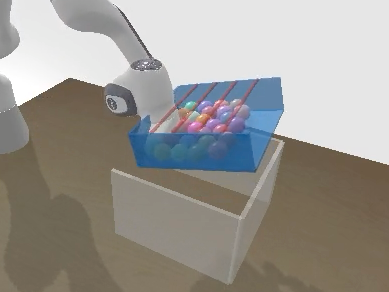}
    \end{subfigure}
  \end{tabular}

  \caption{Qualitative comparison of human-designed, RLBench, and \methodname tools  on three tasks: \texttt{BringCube} (top row), \texttt{ScoreGoal} (middle row), and \texttt{GatherSpheres} (bottom row).}
  \label{fig:qualitative}
\end{figure*}

%\subsection{Experimental Setup}
\paragraph{Evaluation Metrics.} To assess the quality of a tool-action design after each execution, we define these evaluation metrics: \textbf{(1) Task Reward}, which is a set of pre-defined task reward function $R: S\rightarrow r\in[0,1]$ that are unique to each task, where $S$ is its environmental state and $r$ is a normalized reward. These rewards are designed to evaluate the progress made in the task by a certain tool-action pair. \textbf{(2) Distance Traversed}, defined as the total distance traversed (in meters) by the robot end-effector for completing a tool-action pair execution to evaluate the effort efficiency of the design. This is motivated by how better tools tend to reduce the effort needed to complete a task.

% \textbf{Implementation.} We used \texttt{gemini-2.5-pro-preview-03-25} as our VLM at every step throughout the entire evaluation. For tool and action design sampling, we randomize temperatures for each VLM call to obtain greater diversity. \tladd{further explain this part}

We run each baseline approach and our method five times on each task, then report the best and average rewards across those five runs. To more clearly differentiate experiments that resulted in similar rewards, we use the distance traversed as a secondary tie-breaker metric that rewards efficiency.

\subsection{\textbf{Comparing \methodname Results to other Tool Designs}} 

% first thing to talk about the barplot in result
%

% \jsadd{Dinesh: move RLBench away from the bar chart; Add no gripper and human prompt experiment setting to Experiment Setup paragraph.}
% \jsadd{First, analyze Fig. 5. Add average column. under the same page. }
% \jsadd{To show qualitative, on a small subset of tasks (1 or 2), present all the tools synthesized (baseline, humans, ours), replace Fig 4 with this. Have a half page figure of evolution vis.}
The results are summarized in Fig. \ref{fig:best_and_average_bar}. \methodname works consistently well across tasks, in terms of both average and best rewards. We dive into interesting individual method comparisons now. %\textbf{\methodname enhance robots capabilities}. 
As expected, the default Franka Panda two-finger gripper fails on the majority of these tasks. What is perhaps more noteworthy is that \textbf{\methodname outperforms human-prompting!} 
%This is perhaps less expected: \ourmethod outperforms human-specified prompts 
This is true even for expert humans across all tasks and on both metrics (better peak performance and also more reliable). While human prompts occasionally produced strong solutions, their results were less consistent and efficient. In tasks like \texttt{CleanTable} and \texttt{ScoreGoal}, both approaches reached similar peak rewards, but our method did so with significantly shorter paths. For further analysis, Fig. \ref{fig:qualitative} shows example designs from human-prompting and \methodname. Human-designed tools (left column) generally offer suitable forms for task completion; however, \methodname (right column) creates more specialized features that enhance performance. For instance, in task \texttt{ScoreGoal}, our method produces long and bent shapes facilitating simpler, more efficient motions, which the robot just need to move very little along one axis to hit the puck. On the other hand, the straight tool designed from human prompts would require more careful control of the puck. In \texttt{GatherSphere}, our design includes a scoop with side protection and an overhead stripe structure, effectively preventing spheres from bouncing away.

\begin{figure*}[t!]
  \centering
  %–––––––––––––––––––––––––––––––––––––––––––––––––––––––––––––
  % 左侧大图
  \begin{subfigure}[b]{0.45\textwidth}
    \centering
    \includegraphics[width=0.8\linewidth]{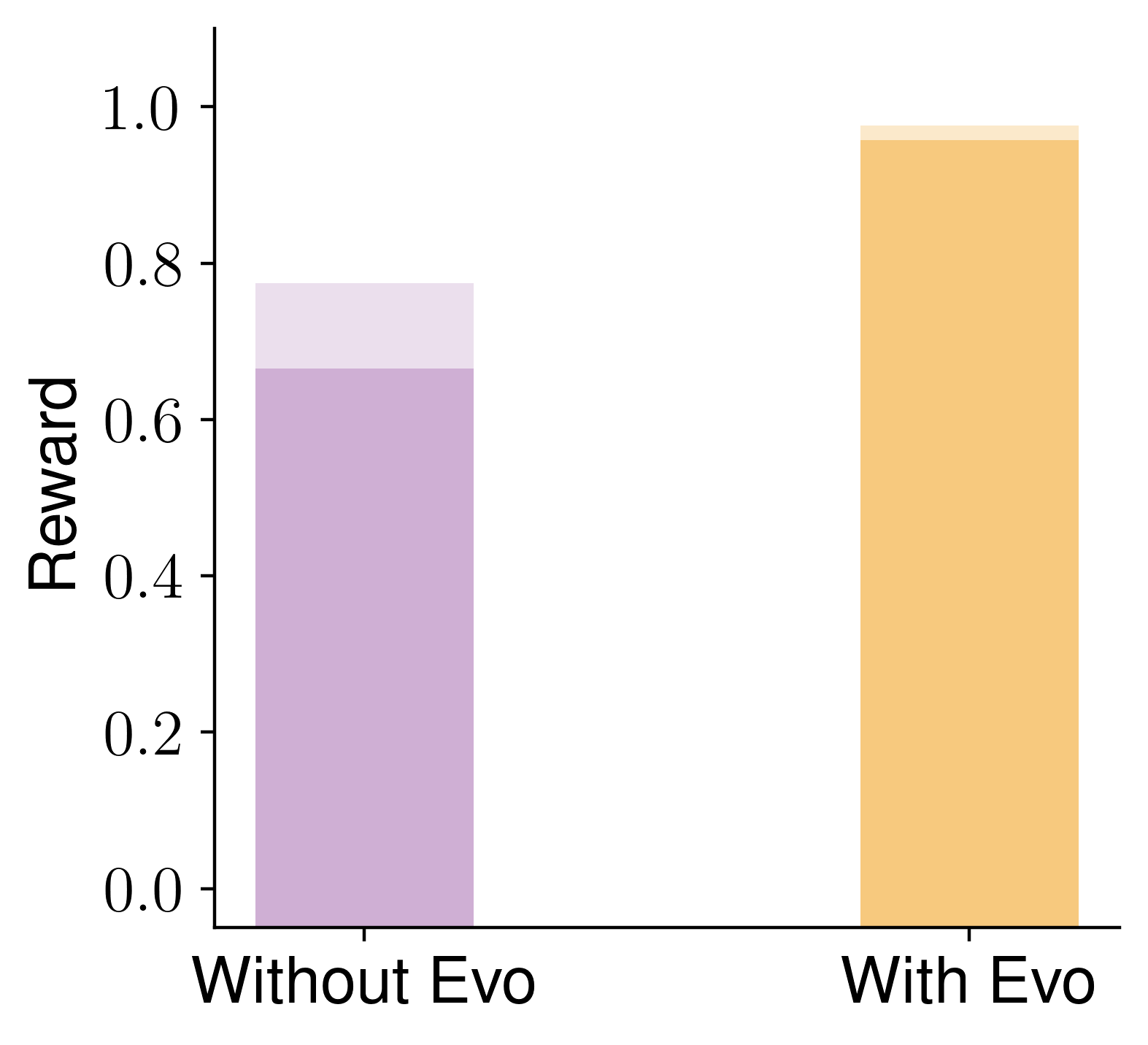}
    \caption{Mean top reward across all evaluated tasks, comparing the Without/With Evo condition.}
    \label{fig:evo_quant}
  \end{subfigure}%
  \quad
  %–––––––––––––––––––––––––––––––––––––––––––––––––––––––––––––
  % 右侧四小图
  \begin{subfigure}[b]{0.45\textwidth}
    \centering
    % 第一行
    \begin{subfigure}[b]{0.5\linewidth}
      \centering
      \includegraphics[width=\linewidth]{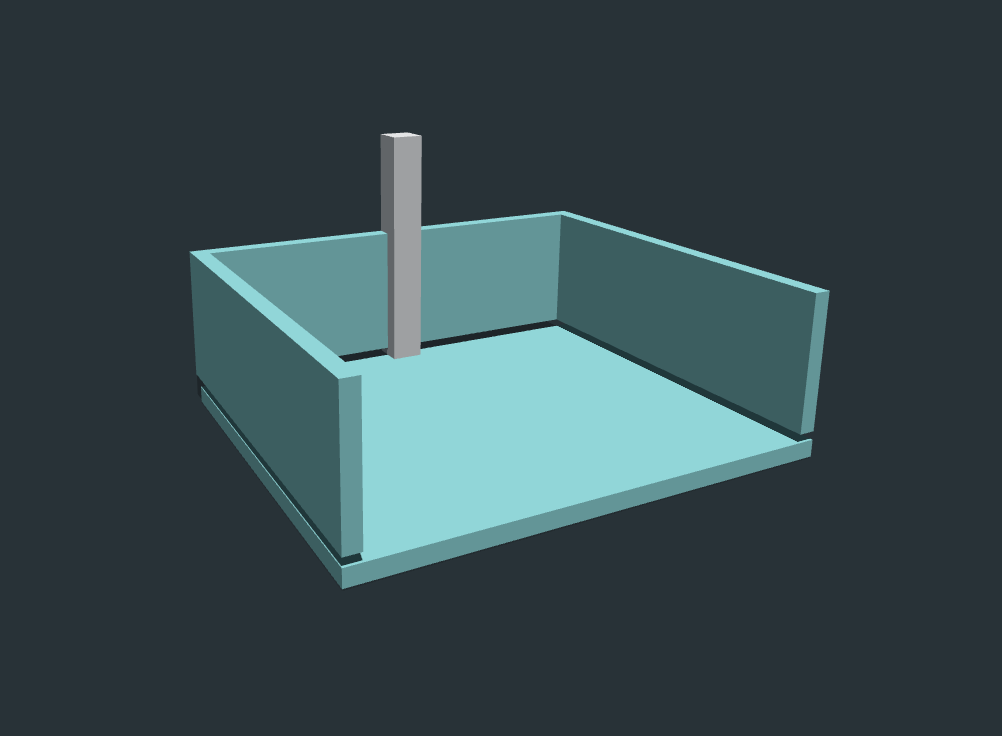}
      % \caption*{GS Before Evo.}
    \end{subfigure}%
    \hfill
    \begin{subfigure}[b]{0.5\linewidth}
      \centering
      \includegraphics[width=\linewidth]{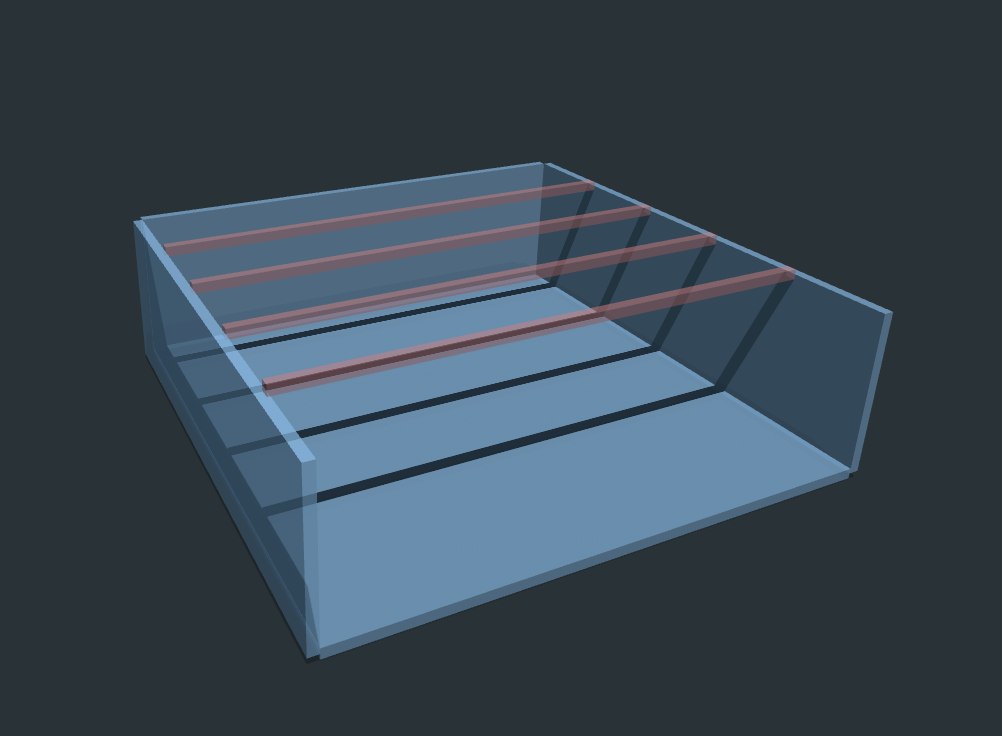}
      % \caption*{GS After Evo.}
    \end{subfigure}
    % 第二行
    \begin{subfigure}[b]{0.5\linewidth}
      \centering
      \includegraphics[width=\linewidth]{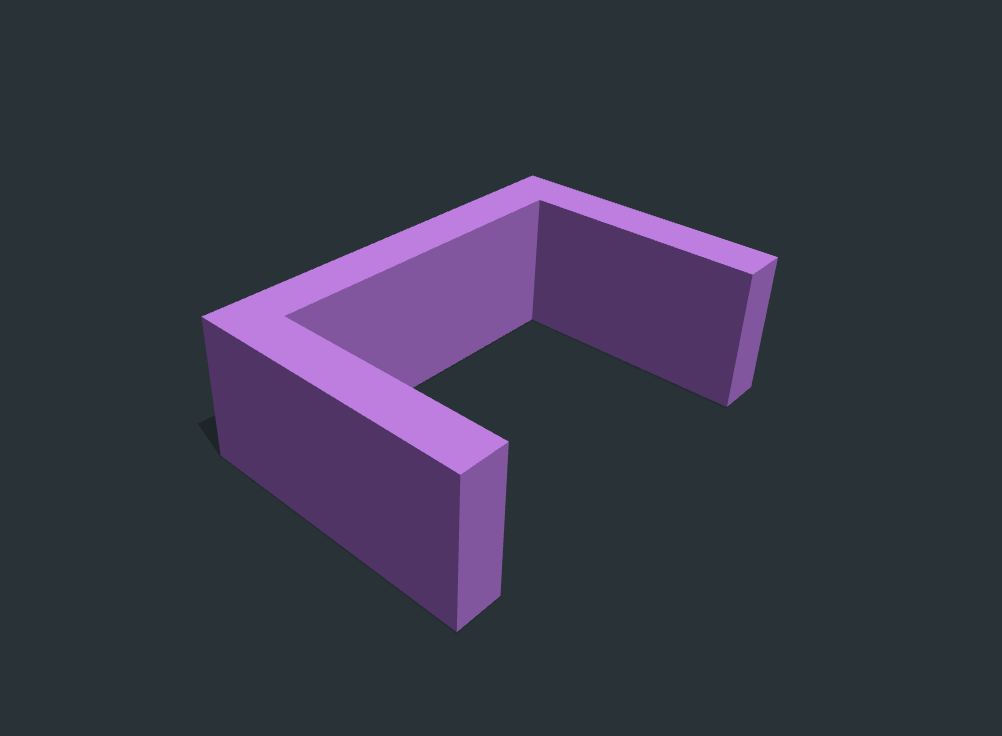}
      \caption*{\normalsize{Before Evo.}}
    \end{subfigure}%
    \hfill
    \begin{subfigure}[b]{0.5\linewidth}
      \centering
      \includegraphics[width=\linewidth]{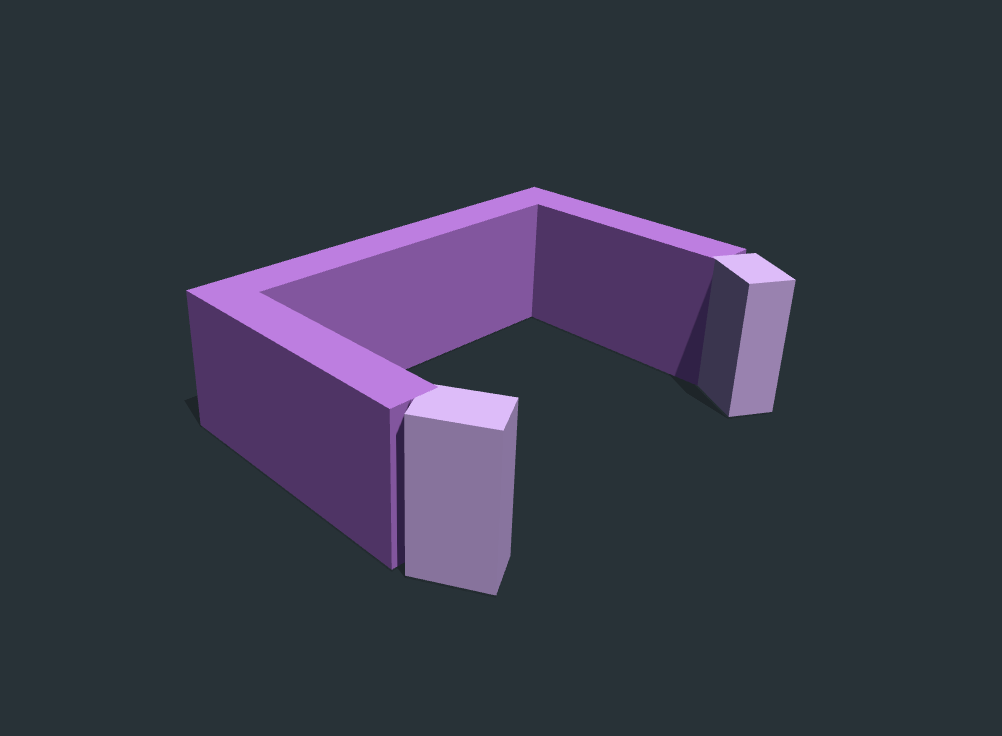}
      \caption*{\normalsize{After Evo.}}
    \end{subfigure}
    \caption{Qualitative comparison before/after evolution on \texttt{GatherSpheres} (top row) and \texttt{MoveBall} (bottom row).}
    \label{fig:evo_qual}
  \end{subfigure}
  %–––––––––––––––––––––––––––––––––––––––––––––––––––––––––––––
  \caption{We present the quantitative (a) and qualitative (b) effectiveness of evolution in tool design.}
  \label{fig:combined_evo}
\end{figure*}
\begin{figure}[h]
    \centering
    \includegraphics[width=0.9\linewidth]{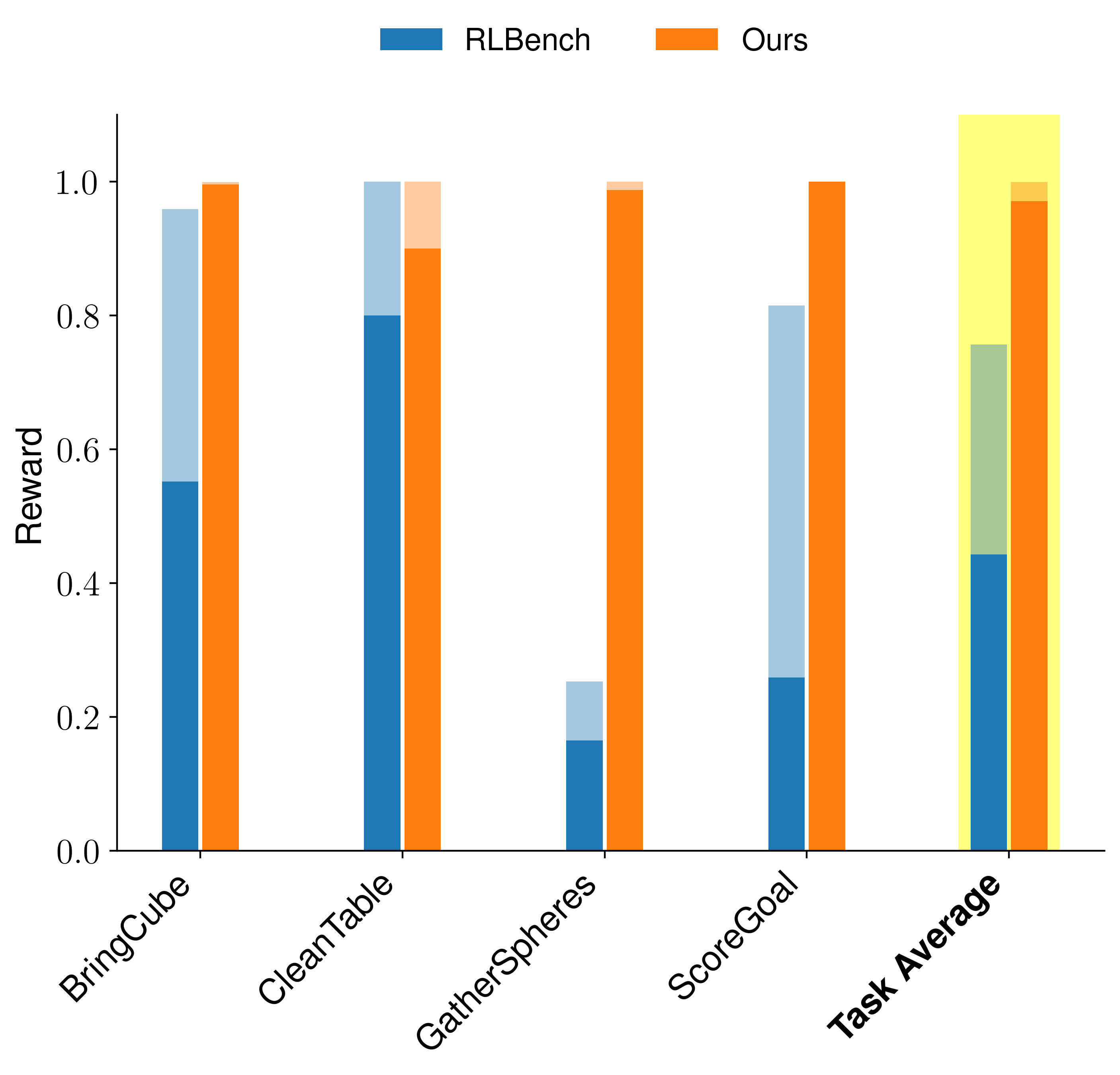}
    \caption{This figure shows the reward of experiments with the original tools in RLBench compared with the experiments on our proposed method across 4 tasks, in which the bars with darker color indicate the average best reward across the five runs, while the bars with paler colors indicate the best reward across the five runs.}
    \label{fig:rlbench_best_and_average_bar}
    \vspace{-2em}
\end{figure}

\textbf{\methodname tools also outperform the RLBench original tools}. On the four RLBench-based tasks in \benchmarkname, we evaluated the standard RLBench Tools (Fig.~\ref{fig:qualitative} middle column). These are often designed to be simple everyday tools that humans might use for those tasks. 
%our \methodname tools on the same set of four tasks.
As shown in Fig. \ref{fig:rlbench_best_and_average_bar}, across every task, \methodname not only attains the highest possible reward across the repeated five runs but does so more reliably (on the average reward) than RLBench Tools. While the rewards in task \texttt{BringCube} and \texttt{CleanTable} are similar to ours, the average rewards over the five repeated runs are generally lower than \methodname. Given that the best rewards over five repeated runs are similar in \texttt{BringCube} and \texttt{CleanTable}, we use distance traversed as the tie breaker. From Fig. \ref{fig:rlbench_best_and_average_bar}, corresponding to the best reward, our method has a lower distance in \texttt{BringCube} and almost the same distance in \texttt{CleanTable}.

Quantitatively inspecting the tools further highlights the advantages of our method over the original tools in RLBench.
% compared to RLBench. 
The RLBench tools (middle column), originally designed for similar but distinct tasks, often underperform due to less optimized features. For example, in \texttt{BringCube}, the RLBench's simple stick, from the original \textit{reach and drag}, provides insufficient lateral control, resulting in inconsistent cube manipulation. Our method’s cage-like structure reliably locks and moves the cube closer, achieving significantly higher rewards. Similarly, in \texttt{ScoreGoal}, RLBench's hockey stick, from their \textit{hockey} task, demands precise, extensive movements, whereas our geometrically optimized tool scores easily with minimal end-effector movement. In \texttt{GatherSpheres}, RLBench’s spatula, from their  \textit{scooping with a spatula} task, lacks effective side control, causing frequent sphere roll-offs, while our design with protective edges achieves more successful scoops and higher task rewards. It is important to note, however, that RLBench tools are not necessarily designed to maximize task ease or performance. Rather, they are chosen to reflect the common tools used by humans for these tasks as a means of benchmarking manipulation policies.

\subsection{Effectiveness of Evolution in Tool Design}
To quantify the necessity of the evolutionary optimization within \methodname, we conducted an ablation study comparing our full evolutionary framework against a sampling-only baseline (\textbf{\methodname w.o. Evolution}), which performs initial tool and action generation without iterative improvement. Fig.~\ref{fig:evo_quant} illustrates the best and average rewards achieved by both configurations across selected tasks.

The results clearly demonstrate the efficacy of evolutionary refinement. \methodname consistently achieves higher task rewards, showcasing its capability to explore and identify superior regions in the joint tool-action design space that remain inaccessible through initial sampling alone. This improvement underscores the value of iterative evolutionary optimization.

To gain qualitative insights into how evolution incrementally enhances designs, we visualize examples from the evolutionary process in Fig.~\ref{fig:evo_qual}. In the \texttt{GatherSpheres} task, the initial scooping tool lacked coverage at the top, allowing spheres to bounce out frequently. Evolutionary iterations addressed this by adding guardrails, significantly improving containment and task success rates. Similarly, in the \texttt{MoveBall} task, the original tool's open-ended design made ball handling challenging. Evolution optimized the geometry by introducing a hugging rim, greatly enhancing control and maneuverability.

These qualitative examples, together with the quantitative evidence, confirm that the evolutionary component of \methodname not only refines initial designs but is critical for achieving robust and effective solutions that maximize performance.

\section{Conclusion}
We propose a new framework for co-optimizing tool design and tool use actions by leveraging the creativity from VLM. By evaluating on \NumBenchmarkTasks different simulation tasks, we demonstrate the capability to design and use tools to solve robotic manipulation tasks. Our results show that we outperform baselines that (1) don't design or use tools and (2) take the specifications directly from humans. We also perform an ablation study to show how our evolutionary module could further boost the performance.

\textbf{Limitations.} While \methodname demonstrates significant advancements in robotic tool and action co-design, several limitations remain that future research should address: (1) Our current framework relies exclusively on simulated environments, potentially impacting the real-world effectiveness and transferability of generated designs. (2) Robot actions are represented as discrete end-effector poses, limiting the handling of complex dynamic tasks requiring precise temporal coordination. (3) Tool representations in URDF format are constrained to simple geometries and limited material properties, and while preliminary results suggest generalization to articulated tools, a comprehensive evaluation is needed. (4) Currently, \methodname is optimized for individual, isolated tasks, and we have not explored multitask optimization or generalization across diverse tasks. Future work should focus on validating designs with real-world robot experiments, enhancing action representations to include dynamic control, and exploring richer, articulated tool designs and multitask generalization.

%% Use plainnat to work nicely with natbib. 
% \clearpage
\bibliographystyle{IEEEtran}
\bibliography{references}

% \clearpage
\newpage
\section{Appendix}

\subsection{Baseline Details}
\subsubsection{Human-Prompted Designs Experiment Implementation}

Each participant underwent the following experimental procedure for each task: (i) We provided a screenshot of the environment and a description of the task, accompanied by a brief Q\&A session to ensure the participants understood the task. (ii) Participants then had five minutes to write a prompt in English specifying their desired tool design and robot action. We instructed participants to be as descriptive as possible while focusing on both the design of the tool and how the robot should use it to accomplish the task. (iii) we integrated their prompt into our standardized request to the VLM (by adding instructions as shown in Appendix \ref{human_spec_instruct}), generating 5 tool described in URDF format along with a batch of 10 samples of action waypoints for each tool. (iv) The VLM outputs were then evaluated in our simulation environment using the same reward metrics described in Section 6. (v) Finally, we evaluated and recorded the best-performing tool and action pair based on the task reward metric for each participant. For a case study, we obtained prompts from three humans coming from three different backgrounds, including an LLM expert (a student with extensive research experience in LLM), a robotics expert (a student with extensive research experience in robotics), and a layperson (with no technical background). This case study will serve as an initial attempt on the concept. In the future, we plan to recruit more human subjects to conduct human study experiments on a larger sample population.

\subsubsection{No-Tool Experiment Implementation}
In the no-tool baseline experiment, we evaluate the robot's performance without any additional tool attachment. The Franka Panda robot uses its original two-finger gripper to perform the task, with the VLM generating action waypoints for the robot end effector pose and gripper open/close, totaling 7 degrees of freedom. The prompt for this baseline is adapted from our proposed prompt by removing the tool design component and associated instructions, while retaining the task description and action generation requirements. We use 5 agents with each generating 10 samples of action waypoints, evaluated using the same metrics introduced in Section 6. The complete no-tool prompt is provided in Appendix \ref{no_tool_instruct}.

\subsubsection{RLBench Experiment Implementation}

In the RLBench experiment, we evaluate the robot's performance with tools from RLbench. We assume the tool is already attached to the end effector without considering the picking step. The tool are scaled to adapt to our tasks which are similar to the ones in RLBench. The prompt for this baseline is also adapted from our proposed prompt by removing the tool design component and associated instructions. We use 5 agents with each generating 10 samples of action waypoints, evaluated using the same metrics introduced in Section 6. The complete no-tool prompt is provided in Appendix \ref{app:rlbench_instruct}.

\subsection{\benchmarkname Details}

Next, we provide detailed descriptions of each task and their corresponding dense reward functions in \benchmarkname.

\begin{table}[H]
  \centering
  \begin{tabular}{@{} L{0.58\linewidth}  L{0.39\linewidth} @{}}
\hline
\vspace{0.6ex}
    % —— BringCube 标题行 ——
    \textbf{\texttt{BringCube}} & \\[0.5ex]
\hline
&\\
    % —— BringCube 描述＋图片 ——
    In this task, a red cube on the desk which is out of the reach of the robot is needed to be brought closer to the target zone.
    
    The reward measures how close the cube is to the target as a fraction of its starting distance, and scales it to 0\textasciitilde1.
      & \includegraphics[width=0.99\linewidth]{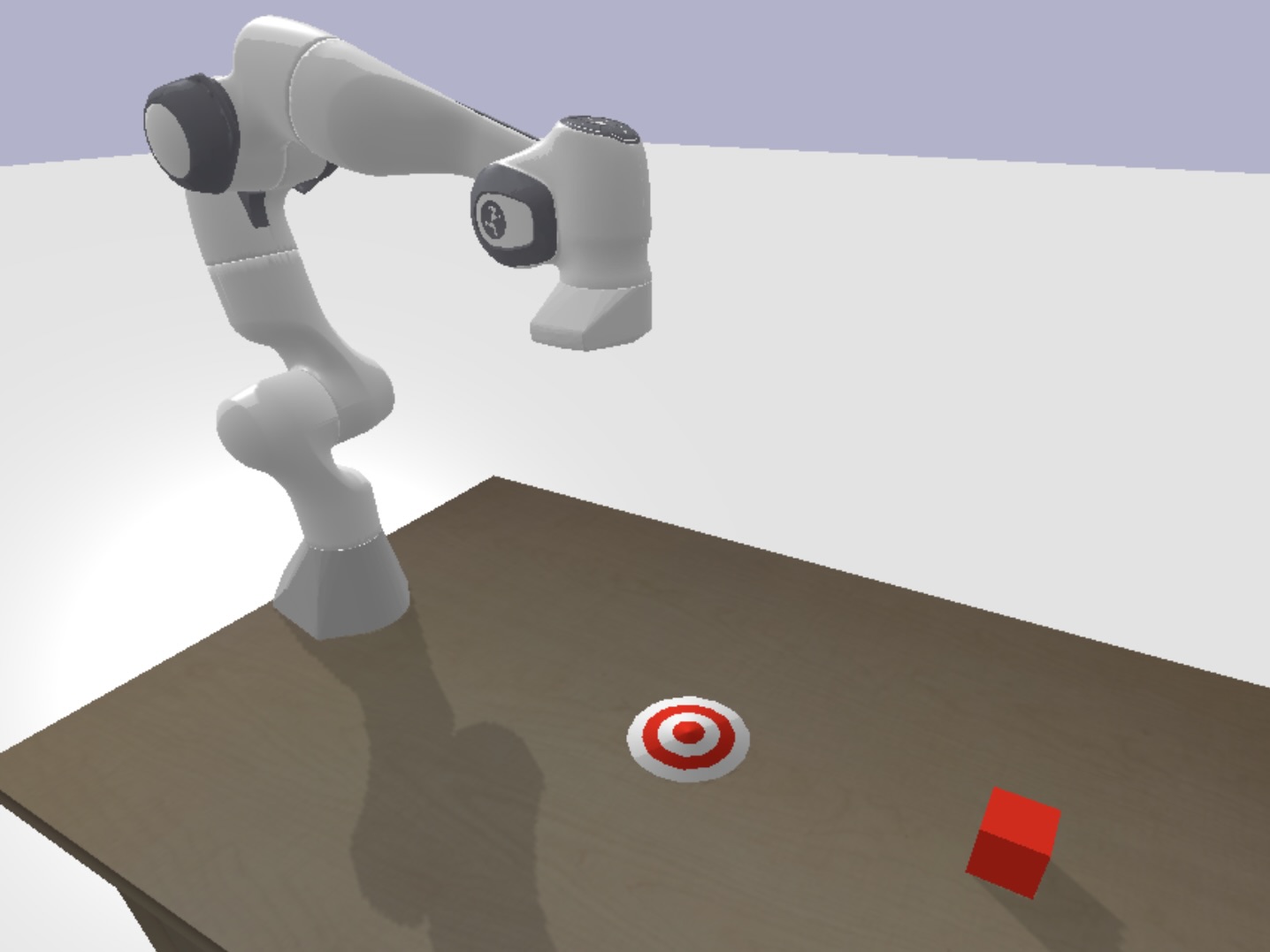}
\end{tabular}
\end{table}
\begin{table}[H]
  \centering
  \begin{tabular}{@{} L{0.58\linewidth}  L{0.39\linewidth} @{}}
\hline
\vspace{0.6ex}
    % —— CleanTable 标题行 ——
    \textbf{\texttt{CleanTable}} & \\[0.5ex]
\hline
&\\
    % —— CleanTable 描述＋图片 ——
    In this task, the colorful cubes representing dusts need to be pushed away from the robot into a circular target zone marked by the green boundary.

    The reward reflects, on average, how far each cube has been pushed toward the goal circle, and scales it to 0\textasciitilde1.
      & \includegraphics[width=0.99\linewidth]{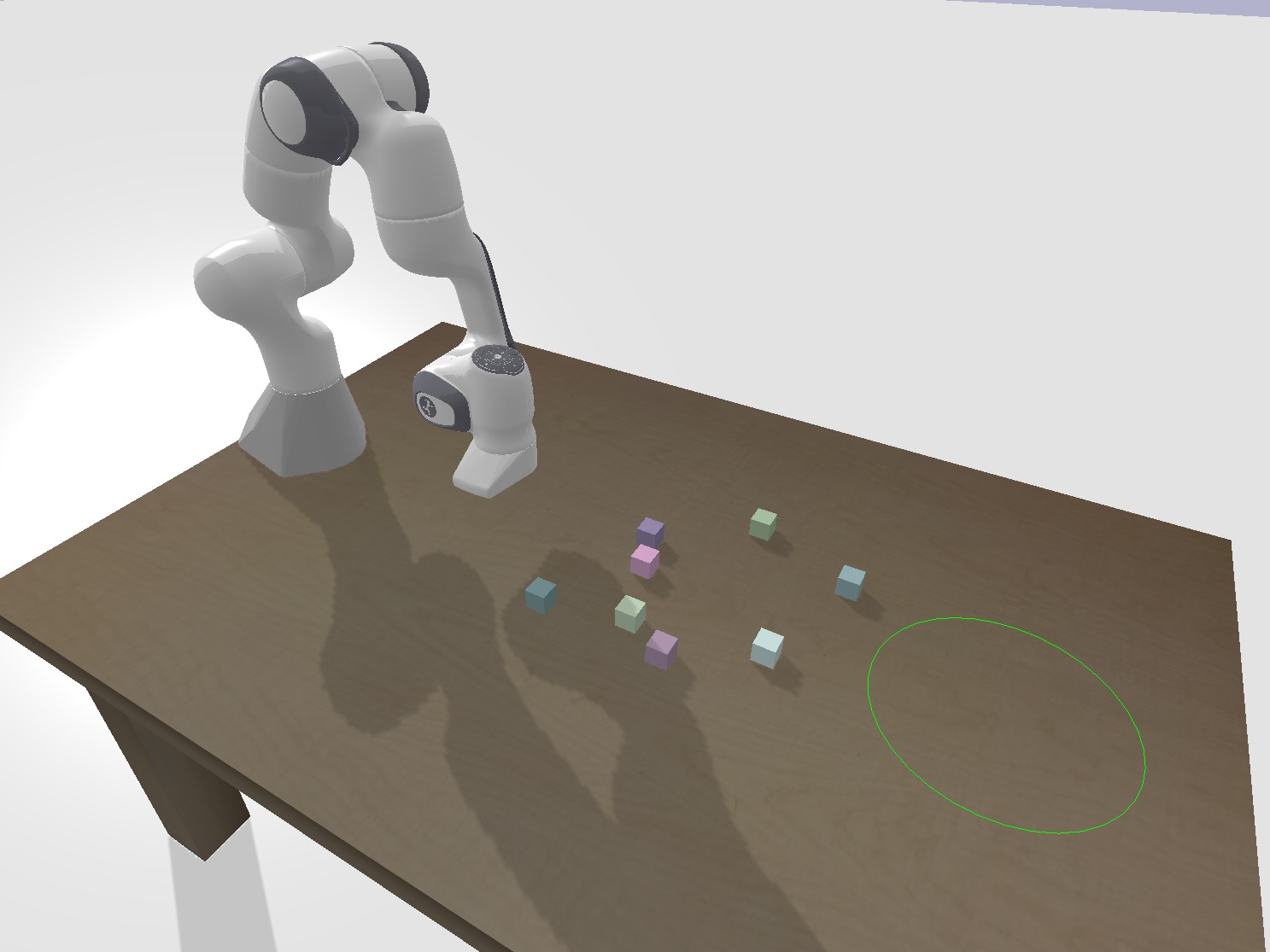}
\end{tabular}
\end{table}
\begin{table}[H]
  \centering
  \begin{tabular}{@{} L{0.58\linewidth}  L{0.39\linewidth} @{}}
\hline
\vspace{0.6ex}
    % —— DislodgeCube 标题行 ——
    \textbf{\texttt{DislodgeCube}} & \\[0.5ex]
\hline
&\\
    % —— DislodgeCube 描述＋图片 ——
    In this task, a red cube is confined within a white, transparent pipe in front of the robot, which has two exits: one opening faces the robot (along negative X) and the other at the front-right corner (along negative Y). The objective is to dislodge the cube through either opening.

    The reward captures the cube’s progress toward either of the two pipe exits by computing two separate, normalized  (on a 0\textasciitilde1 scale) “distance‐to‐exit” scores and then taking the better one.
      & \includegraphics[width=0.99\linewidth]{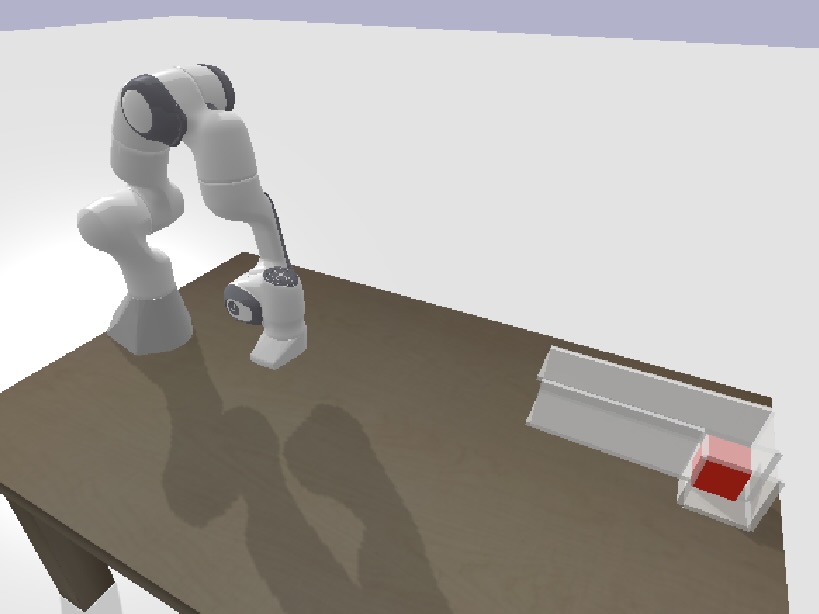}
\end{tabular}
\end{table}
\begin{table}[H]
  \centering
  \begin{tabular}{@{} L{0.58\linewidth}  L{0.39\linewidth} @{}}
\hline
\vspace{0.6ex}
    % —— ElevatePlate 标题行 ——
    \textbf{\texttt{ElevatePlate}} & \\[0.5ex]
\hline
&\\
    % —— ElevatePlate 描述＋图片 ——
    In this task, a white plate placed on the desk in front of the robot needs to be securely lifted up.

    The reward measures how far the plate has moved from its starting position to the desired lifted position, and scales it to 0\textasciitilde1.
      & \includegraphics[width=0.99\linewidth]{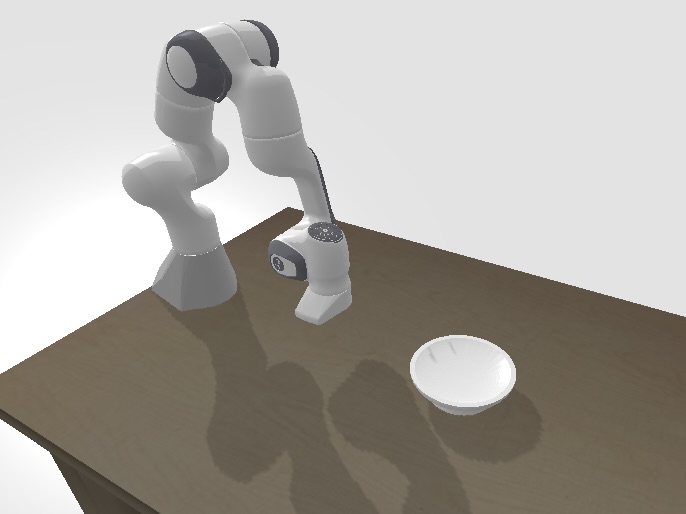}
\end{tabular}
\end{table}
\begin{table}[H]
  \centering
  \begin{tabular}{@{} L{0.58\linewidth}  L{0.39\linewidth} @{}}
\hline
\vspace{0.6ex}
    % —— GatherSpheres 标题行 ——
    \textbf{\texttt{GatherSpheres}} & \\[0.5ex]
\hline
&\\
    % —— GatherSpheres 描述＋图片 ——
    In this task, an open three-walled container filled with small purple spheres is placed before the robot. The objective is to gather and elevate as many spheres as possible above 0.3 m.

    The reward captures, on average, how high the spheres have been lifted up to a specified cap, and scales it to 0\textasciitilde1.
      & \includegraphics[width=0.99\linewidth]{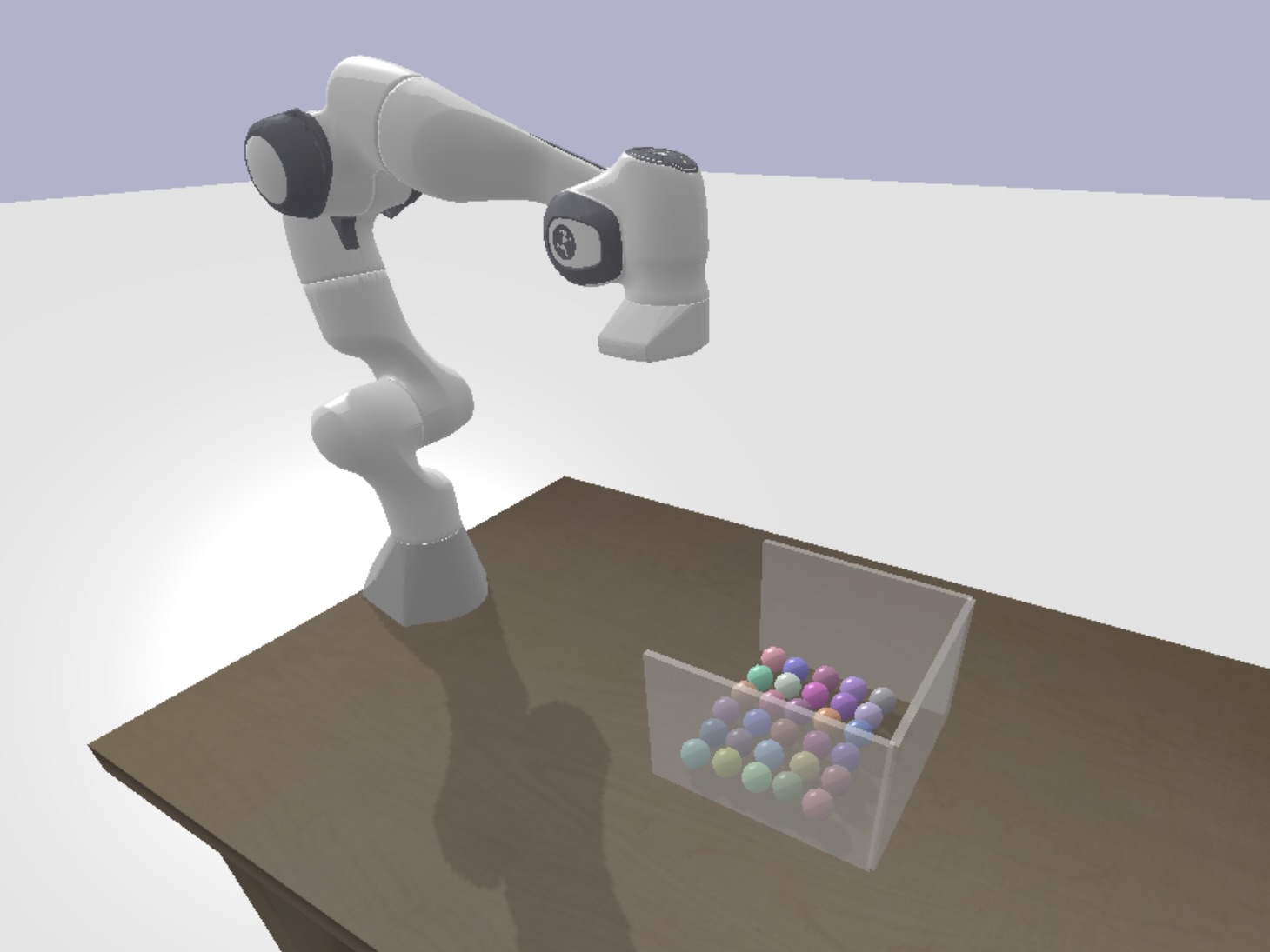}
\end{tabular}
\end{table}
\begin{table}[H]
  \centering
  \begin{tabular}{@{} L{0.58\linewidth}  L{0.39\linewidth} @{}}
\hline
\vspace{0.6ex}
    % —— HighObject 标题行 ——
    \textbf{\texttt{HighObject}} & \\[0.5ex]
\hline
&\\
    % —— HighObject 描述＋图片 ——
    In this task, a green cube sits on the top shelf. The objective is to place it inside the beige box positioned between the shelf and the robot.

    The reward combines a hard “in‐box” check with a smooth distance‐based signal and a bonus for lowering the cube off the shelf, and scales it to 0\textasciitilde1.
      & \includegraphics[width=0.99\linewidth]{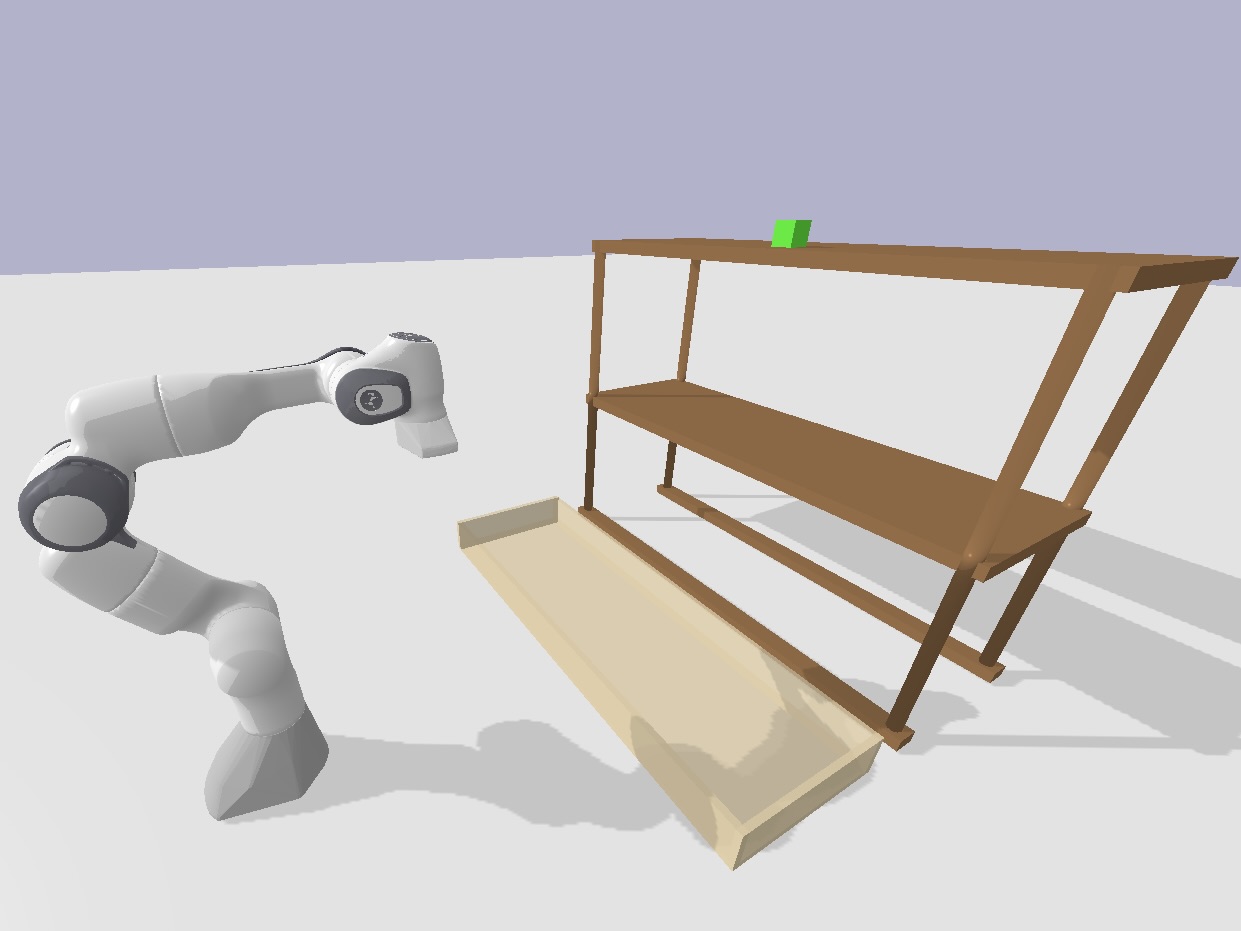}
\end{tabular}
\end{table}
\begin{table}[H]
  \centering
  \begin{tabular}{@{} L{0.58\linewidth}  L{0.39\linewidth} @{}}
\hline
\vspace{0.6ex}
    % —— LiftBox 标题行 ——
    \textbf{\texttt{LiftBox}} & \\[0.5ex]
\hline
&\\
    % —— LiftBox 描述＋图片 ——
    In this task, a brown box on the desk in front of the robot must be lifted above a height threshold of 0.25 m.

    This reward measures how much the box has moved toward its target (lifted) position, and scales it to 0\textasciitilde1.
      & \includegraphics[width=0.99\linewidth]{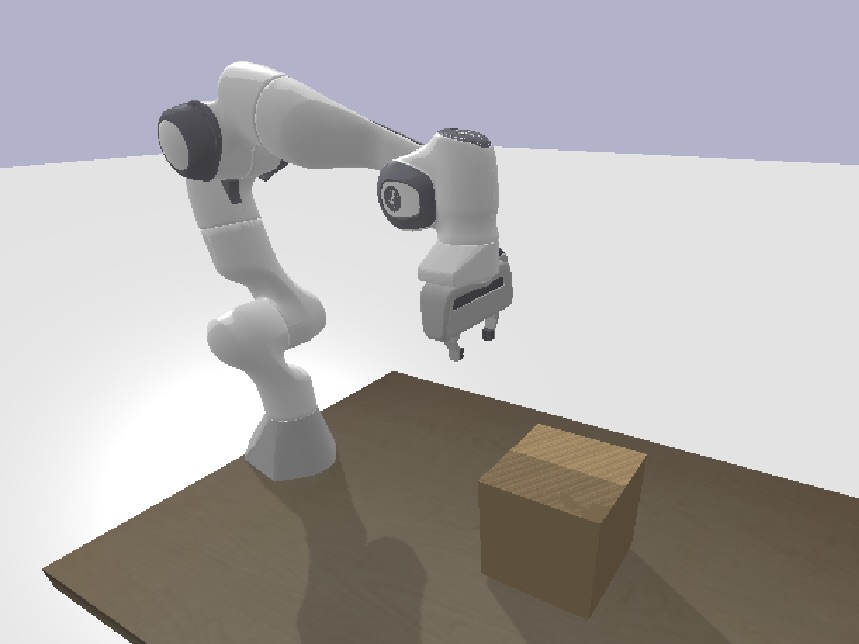}
\end{tabular}
\end{table}
\begin{table}[H]
  \centering
  \begin{tabular}{@{} L{0.58\linewidth}  L{0.39\linewidth} @{}}
\hline
\vspace{0.6ex}
    % —— MoveBall 标题行 ——
    \textbf{\texttt{MoveBall}} & \\[0.5ex]
\hline
&\\
    % —— MoveBall 描述＋图片 ——
    In this task, a red ball on the desk must be moved from the robot’s left side to its right side.

    The reward balances two objectives, getting the ball toward the right‐side target and keeping its speed in check, and scales it to 0\textasciitilde1.
      & \includegraphics[width=0.99\linewidth]{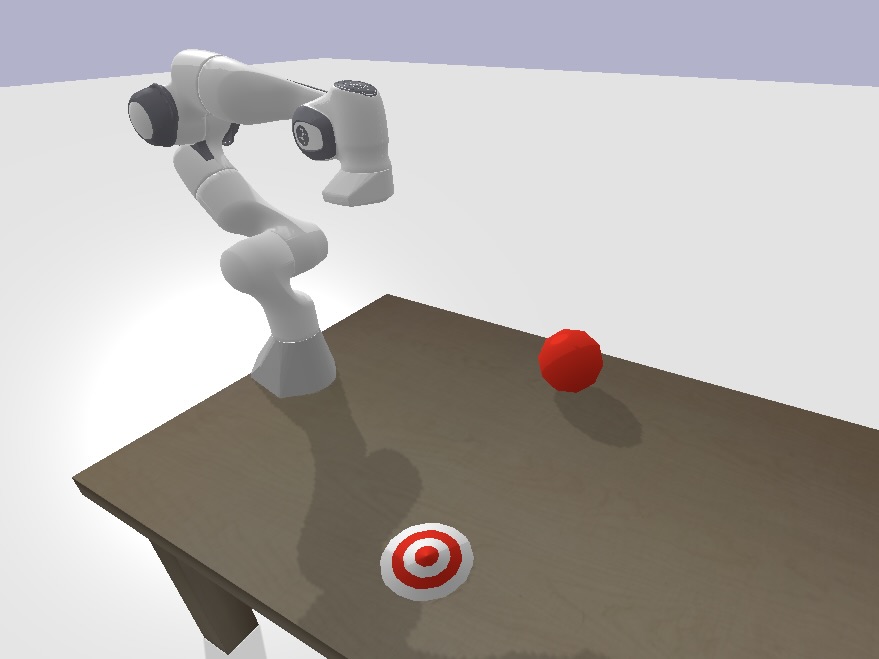}
\end{tabular}
\end{table}
\begin{table}[H]
  \centering
  \begin{tabular}{@{} L{0.58\linewidth}  L{0.39\linewidth} @{}}
\hline
\vspace{0.6ex}
    % —— OneBook 标题行 ——
    \textbf{\texttt{OneBook}} & \\[0.5ex]
\hline
&\\
    % —— OneBook 描述＋图片 ——
    In this task, two book holders with five books between them are in front of the robot. The objective is to pull out the middle (3rd) book while keeping the others in place.

    This reward balances two goals, pulling out the middle book and keeping the others perfectly still, and scales it to 0\textasciitilde1.
      & \includegraphics[width=0.99\linewidth]{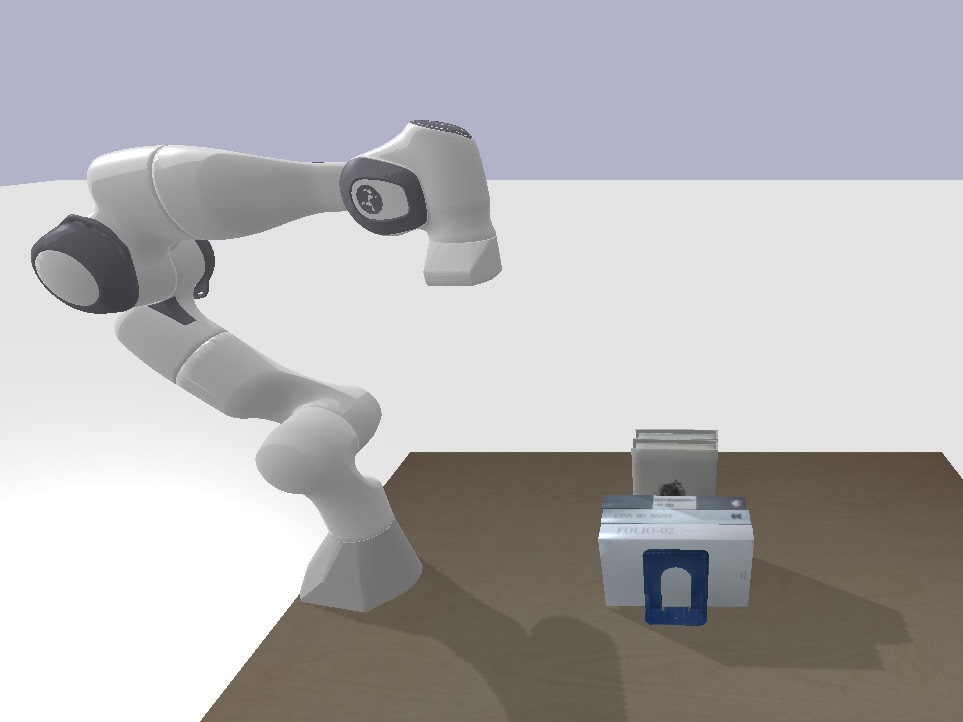}
\end{tabular}
\end{table}
\begin{table}[H]
  \centering
  \begin{tabular}{@{} L{0.58\linewidth}  L{0.39\linewidth} @{}}
\hline
\vspace{0.6ex}
    % —— ScoreGoal 标题行 ——
    \textbf{\texttt{ScoreGoal}} & \\[0.5ex]
\hline
&\\
    % —— ScoreGoal 描述＋图片 ——
    In this task, a hockey puck and a goal are placed on the ground far from the robot. The objective is to place the puck inside the goal.

    The reward gives full credit once the puck is entirely inside the goal’s 3D bounding box, and otherwise scales linearly with how much closer the puck is, horizontally, to the goal than it was at the start, and scales it to 0\textasciitilde1.
      & \includegraphics[width=0.99\linewidth]{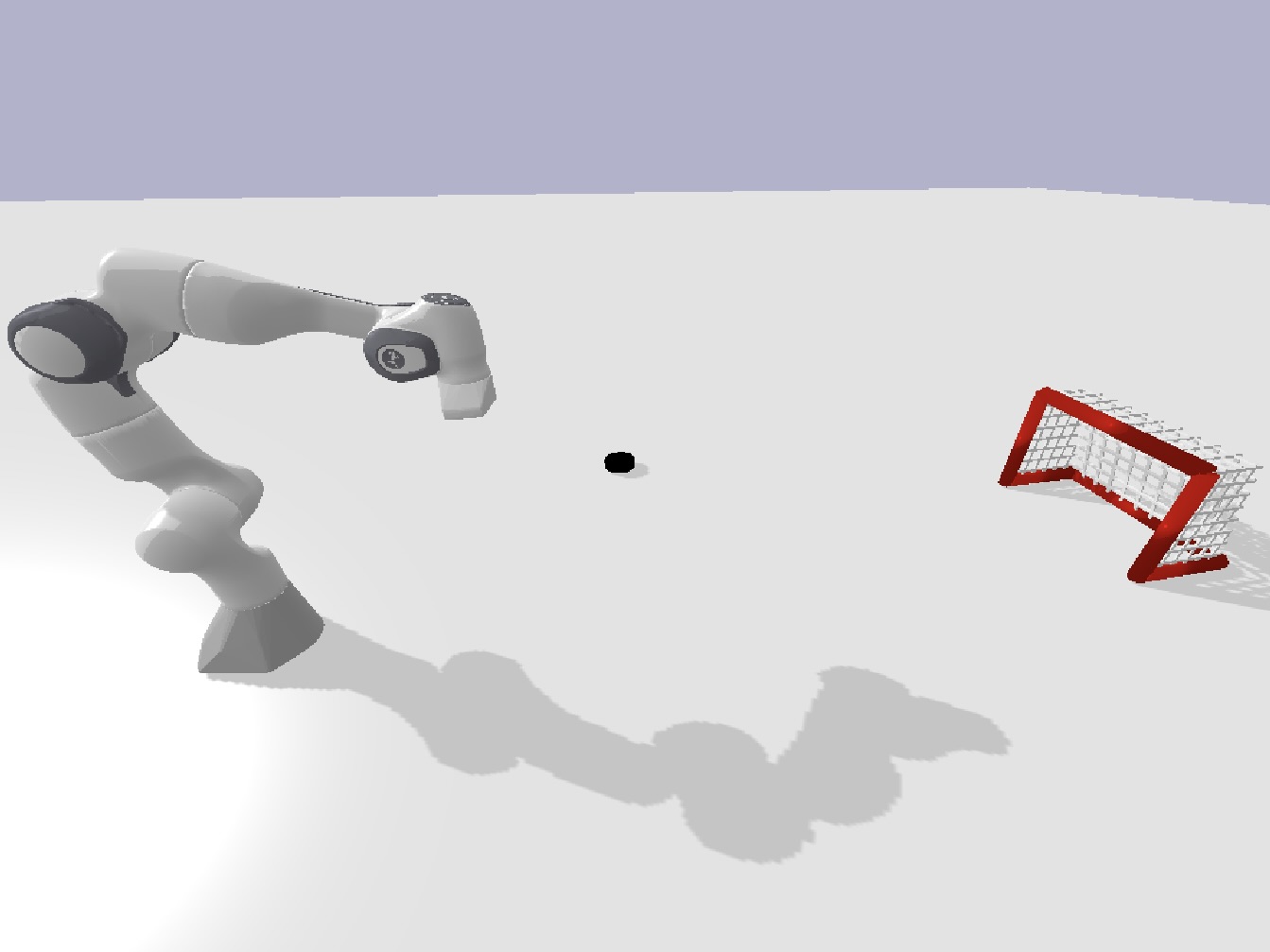}
\end{tabular}
\end{table}
\begin{table}[H]
  \centering
  \begin{tabular}{@{} L{0.58\linewidth}  L{0.39\linewidth} @{}}
\hline
\vspace{0.6ex}
    % —— SnatchCookie 标题行 ——
    \textbf{\texttt{SnatchCookie}} & \\[0.5ex]
\hline
&\\
    % —— SnatchCookie 描述＋图片 ——
    In this task, a transparent jar of cookies sits on the desk in front of the robot. The objective is to take at least one cookie from the jar.

    The reward checks whether any cookie has been lifted out of the jar, and otherwise gives partial credit, from 0 to 1, based on how high the tallest cookie has been raised.
      & \includegraphics[width=0.99\linewidth]{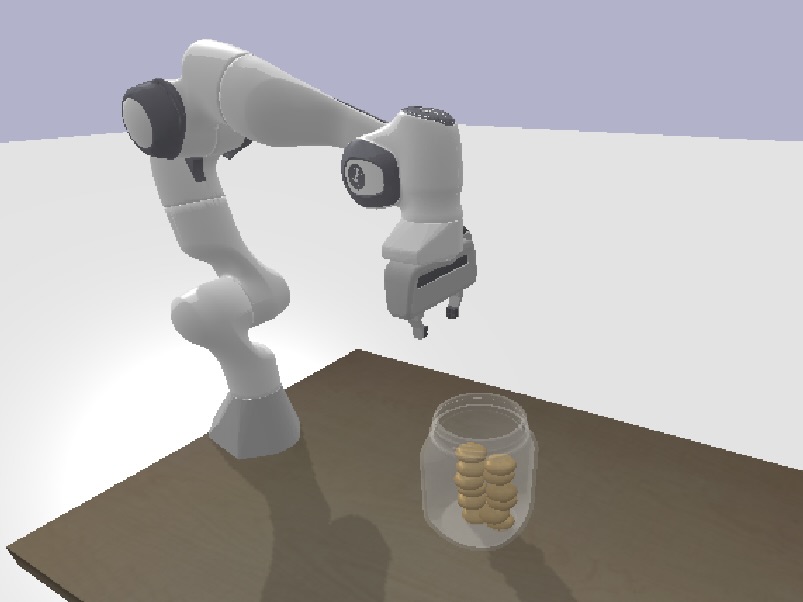}
\end{tabular}
\end{table}
\begin{table}[H]
  \centering
  \begin{tabular}{@{} L{0.58\linewidth}  L{0.39\linewidth} @{}}
\hline
\vspace{0.6ex}
    % —— TurkeyLegs 标题行 ——
    \textbf{\texttt{TurkeyLegs}} & \\[0.5ex]
\hline
&\\
    % —— TurkeyLegs 描述＋图片 ——
    In this task, a silver pot with handles on both sides, full of turkey legs, sits on the desk in front of the robot. To the pot’s left (robot’s perspective) is a chef’s box. The objective is to transfer all turkey legs into the box without moving the pot.

    The reward combines two checks, keeping the pot out of the box and getting each turkey leg into the box, by multiplying, and scales it to 0\textasciitilde1.
      & \includegraphics[width=0.99\linewidth]{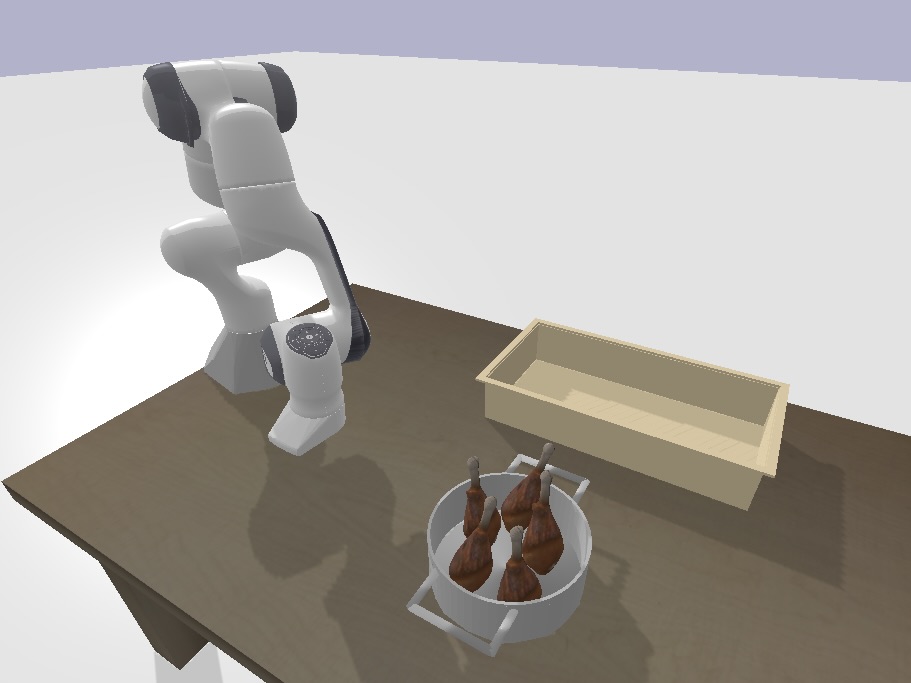}
\end{tabular}
\end{table}

\subsection{\methodname Implementation Details} \label{app:implementation}

\subsubsection{Design Agents Context}

This section describes the context that applies to a single design agent. The design agent is provided with task context as illustrated in Fig. 2, which includes (1) the environment code, (2) a screenshot of the environment, (3) a brief task description, and (4) a total text prompt composed by the prompts in Appendix \ref{app:prompts}. We also provide task-agnostic context, including (i) environment base class, which provides basic task-agnostic functionalities, (ii) environment runner, which establishes the context for how we will use the output tool and action, and (iii) a URDF of the Franka Panda without the tool to indicate where to attach the tool. All task environments are implemented as child classes of the environment base class, ensuring they inherit the fundamental functionalities while allowing for task-specific implementations.

\begin{table*}[h!]
\centering
\caption{Benchmarking Parameters for Different Tasks}
\label{tab:task_configs}
% \footnotesize
\small
\begin{tabular}{@{}lccccccc@{}}
\toprule
\textbf{Task Name}    & $n_{agent}$ & $n_{tool}$ & $n_{action}$ & $k_{top}$ & $reward_{save}$ & $n_{iteration}$ & $k_{sim}$\\
\midrule
BringCube             & 20   & 10   & 10   & 5   & 0.6 & 3 & 100\\
CleanTable            & 20   & 10   & 10   & 5   & 0.6 & 3 & 100\\
DislodgeCube          & 20   & 10   & 10   & 5   & 0.6 & 3 & 100\\
ElevatePlate          & 20   & 10   & 10   & 5   & 0.6 & 3 & 100\\
GatherSpheres         & 20   & 10   & 10   & 5   & 0.6 & 3 & 100\\
HighObject            & 20   & 10   & 10   & 5   & 0.5 & 3 & 100\\
LiftBox               & 30   & 15   & 15   & 5   & 0.1 & 3 & 100\\
MoveBall              & 20   & 10   & 10   & 5   & 0.6 & 3 & 100\\
OneBook               & 20   & 10   & 10   & 5   & 0.4 & 3 & 100\\
ScoreGoal             & 20   & 10   & 10   & 5   & 0.4 & 3 & 100\\
SnatchCookie          & 5    & 5    & 5    & 5   & 0.3 & 3 & 100\\
TurkeyLegs            & 30   & 10   & 15   & 5   & 0.2 & 4 & 100\\
\bottomrule
\end{tabular}
\end{table*}

\subsubsection{Design Agents Queries}

In \methodname, we query VLM for designs by first initializing $n_{agents}$ number of agents in parallel with the same prompts. For each agent, we prompt it to generate $n_{tool}$ number of tool designs, and $n_{action}$ number of action waypoint samples that correspond to \textit{each} tool design. Therefore, the total number of tool-action pairs that are generated via one complete query is $n_{agents} \times n_{tool} \times n_{action}$. The prompt we use to specify this behavior to each agent is presented in Appendix \ref{app:prompts_procedure}.

We explicitly choose this style of querying to maximize time efficiency and design diversity: (1) time efficiency is achieved by reducing the querying algorithmic complexity by using parallel VLM agents. (2) Empirically, we found that design diversity is achieved when we balance dependence and independence between design decisions. Specifically, when a single VLM agent auto-regressively generates $n_{tool} \times n_{action}$ tool-actions pairs, having later design outputs be conditioned on previous design outputs can encourage diversity within that conditional distribution. However, in order to sample from many distinct conditional distributions, as this provides additional diversity, we found that parallel VLM queries that share no history can help with that. Ultimately, we found that optimizing time efficiency and design diversity led to better and faster initial samples as well as evolutions. 

See Appendix \ref{app:params} for details on the values we used for these parameters for benchmarking.

\subsubsection{Design Agent Outputs}

As a part of our prompt to the VLM to query for designs, we specified our desired tool and action formats. For our tool design requirements, please refer to Appendix \ref{app:prompts_tool} for details. For our action design requirements, please refer to Appendix \ref{app:prompts_action} for details. Notably, these prompts are separated into with \& without the Franka gripper usage. During \methodname's initial sampling, some design agents are asked to design tools for the gripper, some are not. This ensures the full capabilities of the default morphology are used. The two types of tools are also specified with different required attachment locations: gripper-using tools are asked to be attached to the two Franka gripper fingers, and non-gripper-using tools are asked to be attached to a ``virtual joint'', which is a joint we set up positioned at the flange of the Franka end effector to make the attachment process more standardized.

\subsubsection{Simulation Evaluation}

From the previous section, we obtain a list of tool-action pairs in the form of URDF designs and action waypoints, respectively. To use these for simulation evaluation, we first merge the tool URDF without modification into a blank Franka Panda URDF (a blank Franka URDF will contain a gripper if gripper usage is enabled, and otherwise will not). For the action waypoints, which are inherently sparse, we implement linear interpolation for the position trajectory and SLERP (Spherical Linear Interpolation) for the orientation trajectory. The Pybullet simulation then executes these interpolated actions in the designated task environment. Finally, the environment returns result metrics for each run with the corresponding samples, allowing for both choosing evolution candidates and for producing the quantitative evaluation of the design performance. To speed up the evaluation, $k_{sim}$ samples are evaluated in parallel Pybullet simulations at a time.

\subsubsection{Evolution}

After evaluating all previous tool-action pairs, we perform selection as follows: (1) For every task, we define two parameters to control the behavior of selection: $reward_{save}$ and $k_{top}$. (2) Using these parameters, we first select the $k_{top}$ number of tool-action pairs with the highest task rewards, and then keep only the pairs that have a reward higher than the $reward_{save}$ threshold, resulting in a set of winner tool-action pairs. We found this selection mechanism empirically allows for the best signals for evolution. 

We then take this winner tool-action pair set and feed it as context into the next design agent query. These previous designs are introduced to the VLM by the ``evolution mission introduction prompt'' in Appendix \ref{app:prompts_intro}, where the VLM is asked to perform mutation and crossover on the previous tools via the rules specified in \ref{app:prompts_evo}. These evolved design samples will be fed into the simulation for evaluation, and the cycle will continue. We define a final $n_{iteration}$ parameter to control the number of iterations that this cycle would go on for.

See Appendix \ref{app:params} for details on the values we used for these parameters for benchmarking.

\subsubsection{\methodname Benchmarking Details}\label{app:params}
When benchmarking \methodname against 
\benchmarkname, we used a different set of parameters for each task, detailed in Table.~\ref{tab:task_configs}. We used \texttt{gemini-2.5-pro-preview-03-25} as our VLM model throughout the entire experiment, and ran PyBullet evaluations on an AMD Ryzen 7 9800X3D 8-Core Processor CPU with 64 GB of RAM. On average, one run of \methodname on one of these tasks should take around 30 minutes.

\subsection{Full Prompts} \label{app:prompts}
In this section, we provide all \methodname prompts. We show individual prompt components in section~\ref{app:prompts_intro}-~\ref{app:prompts_evo}. We then describe we compose these prompts for different experiments in section~\ref{no_tool_instruct}-\ref{app:rlbench_instruct}. For details of their usage, please refer to Appendix \ref{app:implementation}. 

\subsubsection{Mission Introduction} \label{app:prompts_intro}

Initial sampling mission introduction prompt:
% \noindent
% \begin{tcolorbox}[colback=gray!5!white, colframe=gray!75!black, boxrule=0.5pt]
\begin{mdframed}[  
  backgroundcolor=gray!5,  % instead of colback
  linecolor=gray!75,       % instead of colframe
  linewidth=0.5pt,         % instead of boxrule
  skipabove=6pt, 
  skipbelow=6pt, 
]
\small
\texttt{You are a robotics hardware and controls expert. You operate with boldness and brilliance in the physical realm. You work with a robot arm that sits in the origin of your environment. You will be presented with some robotic tasks, and will be asked to design tools and actions to complete the task. Your goal is not to complete the task to perfection in one fell swoop. Instead, your meta-goal is to generate a wide range of differentiated good solutions over time, where one of them will inevitably succeed.}
% \end{tcolorbox}
\end{mdframed}

Evolution mission introduction prompt:
% \begin{tcolorbox}[colback=gray!5!white, colframe=gray!75!black, boxrule=0.5pt]
\begin{mdframed}[  
  backgroundcolor=gray!5,  % instead of colback
  linecolor=gray!75,       % instead of colframe
  linewidth=0.5pt,         % instead of boxrule
  skipabove=6pt, 
  skipbelow=6pt, 
]
\small
\texttt{You are a robotics hardware and controls expert. You operate with boldness and brilliance in the physical realm. The goal is to create tools and actions to complete a given task. You will be given a list of previously generated tool designs via JSON with URDF. Your goal is to evolve the tool designs via mutation and crossover, and generate the new best actions for the evolved tools. This will be done in a way that is similar to genetic algorithms, and will be specified in detail in the "Evolutionary Process" section below.}
% \end{tcolorbox}
\end{mdframed}

\subsubsection{Procedure Instruction} \label{app:prompts_procedure}

% \begin{tcolorbox}[colback=gray!5!white, colframe=gray!75!black, boxrule=0.5pt]
\begin{mdframed}[
  backgroundcolor=gray!5,  % instead of colback
  linecolor=gray!75,       % instead of colframe
  linewidth=0.5pt,         % instead of boxrule
  skipabove=6pt, 
  skipbelow=6pt, 
]
\small
\texttt{The procedure you will follow:\\
1. Receive Environment Descriptions: The user will provide some detailed environment descriptions, robotic task instructions, and an initial image of the workspace area from the overhead camera.\\
2. Describe the Scene: Analyze the environment. Write down the spatial relationship, including by not limited to the position, orientation, dimension, and geometry of all the objects in the scene. Use all the information provided to you, including all text, code, and images.\\
3. Create Strategies and Designs: You will need to create $n_{tool}$ tool that you can use to complete the task. For each of the tools you designed, you must generate $n_{action}$ set of action waypoints that you can use to complete the task. Specifically, for a total of $n_{tool}$ times, do the following steps:
\begin{itemize}[label={}]
  \item (a) First, write down a completely different, out-of-the-box tool design to tackle the task. Make it unlike any other tool design you made in your other strategies.
  \item (b) Create these tools following the "Tool Specification" section below.
  \item (c) For this tool, write the following down: (1) The spatial relationship (pose transformation) between the end-effector and each component of the tool; (2) The 3D space that each tool component will take up when connected to the robot; (3) The usage of each component of the tool when carrying out the task.
  \item (d) Use your previous analysis to tweak any obvious issues with the position, orientation, and dimension of your tool design.
  \item (e) Next, using your knowledge of the tool and your in depth analysis regarding the intricate 3D spatial relationships between the tool and its environment, create $n_{action}$ number of different step by step action plans to enable to effective tool use (See more in "Desired Action Criteria Definitions"). Be very wary about how objects interact with each other\!
  \item (f) Transform your step-by-step action plan into waypoints adhering to the "Action Specifications". During this transformation, think about the inherent nature of controlling robots with waypoint control and the difficulty that may present.
\end{itemize}}
% \end{tcolorbox}
\end{mdframed}

\subsubsection{Tool Specifications} \label{app:prompts_tool}
Tool specification prompt without the use of Franka Grippers:
% \begin{tcolorbox}[colback=gray!5!white, colframe=gray!75!black, boxrule=0.5pt]
\begin{mdframed}[
  backgroundcolor=gray!5,  % instead of colback
  linecolor=gray!75,       % instead of colframe
  linewidth=0.5pt,         % instead of boxrule
  skipabove=6pt, 
  skipbelow=6pt, 
]
\small
\texttt{(Tool Specifications) Your design of the tool must follow these rules: (1) You must only use 3D rectangles for each component; (2) Your tool will be outputted in a URDF block format, which should be directly added to the end of a panda URDF file, before the robot closing declaration; (3) Make sure your tools weigh very little in the URDF file, where each tool part should weigh no more than a few grams (these weights do not have to be realistic, it is just for the robot inverse kinematics to have a easier time converging). (4) Your design will be a single rigid tool, which should be attached directly to the "panda\_virtual" link, which you can safely assume to have the same orientation as the world frame. (5) Any attachments you design should geometrically be directly connected to their parent links in the URDF (there should be no gaps in between!) (6) As a general observation, you perform better when the tools you design are complex and intricate.}
% \end{tcolorbox}
\end{mdframed}
Tool specification prompt with the use of Franka Grippers:
% \begin{tcolorbox}[colback=gray!5!white, colframe=gray!75!black, boxrule=0.5pt]
\begin{mdframed}[
  backgroundcolor=gray!5,  % instead of colback
  linecolor=gray!75,       % instead of colframe
  linewidth=0.5pt,         % instead of boxrule
  skipabove=6pt, 
  skipbelow=6pt, 
]
\small
\texttt{(Tool Specifications) Your design of the tool must follow these rules: (1) You must only use 3D rectangles for each component; (2) Your tool will be outputted in a URDF block format, which should be directly added to the end of a panda URDF file, before the robot closing declaration; (3) Make sure your tools weigh very little in the URDF file, where each tool part should weigh no more than a few grams (these weights do not have to be realistic, it is just for the robot inverse kinematics to have a easier time converging).  (4) Your design will be a pair of attachments to the robot gripper fingers (which allows the tool to be actuated with the robot gripper); You should attach the left attachment to "panda\_leftfinger" and the right attachment to "panda\_rightfinger". (5) Any attachments you design should geometrically be directly connected to their parent links in the URDF (there should be no gaps in between!) (6) As a general observation, you perform better when the tools you design are complex and intricate.}
% \end{tcolorbox}
\end{mdframed}

\subsubsection{Action Specifications} \label{app:prompts_action}
Action specification prompt without the use of Franka Grippers:
% \begin{tcolorbox}[colback=gray!5!white, colframe=gray!75!black, boxrule=0.5pt]
\begin{mdframed}[
  backgroundcolor=gray!5,  % instead of colback
  linecolor=gray!75,       % instead of colframe
  linewidth=0.5pt,         % instead of boxrule
  skipabove=6pt, 
  skipbelow=6pt, 
]
\small
\texttt{(Action Specifications) Your tool-using action will be a Nx6 numpy array of action waypoints, where N is the number of waypoints, and each waypoint is of dimension 6 (xyz position + roll-pitch-yaw euler angle orientations). Your action needs to be precisely six numbers per waypoint. Your waypoints will be carried out by the EnvRunner class. It is important to stress this: the action waypoints are controlling the robot end-effector "panda\_virtual" link: this means you have to carefully take into account the dimensions of the tool and the thickness of its parts when designing effective waypoints. Again, you can safely assume the end-effector has the same orientation as the world frame upon initialization (see frame clarification again for details)!}
% \end{tcolorbox}
\end{mdframed}

Action specification prompt with the use of Franka Grippers:
% \begin{tcolorbox}[colback=gray!5!white, colframe=gray!75!black, boxrule=0.5pt]
\begin{mdframed}[
  backgroundcolor=gray!5,  % instead of colback
  linecolor=gray!75,       % instead of colframe
  linewidth=0.5pt,         % instead of boxrule
  skipabove=6pt, 
  skipbelow=6pt, 
]
\small
\texttt{(Action Specifications) Your tool-using action will be a Nx7 numpy array of action waypoints, where N is the number of waypoints, and each waypoint is of dimension 7 (xyz position + roll-pitch-yaw euler angle orientations + binary gripper open/close state in integers [0 for open, 1 for closed]). Your action needs to be precisely seven numbers per waypoint. Your waypoints will be carried out by the EnvRunner class. It is important to stress this: the action waypoints are controlling the robot end-effector "panda\_virtual" link: this means you have to carefully take into account the dimensions of the tool and the thickness of its parts when designing effective waypoints. Again, you can safely assume the end-effector has the same orientation as the world frame upon initialization (see frame clarification again for details)!}
% \end{tcolorbox}
\end{mdframed}

\subsubsection{Action Diversity Specification}\label{app:prompt_action_diversity}
% \begin{tcolorbox}[colback=gray!5!white, colframe=gray!75!black, boxrule=0.5pt]
\begin{mdframed}[
  backgroundcolor=gray!5,  % instead of colback
  linecolor=gray!75,       % instead of colframe
  linewidth=0.5pt,         % instead of boxrule
  skipabove=6pt, 
  skipbelow=6pt, 
]
\small
\texttt{(Desired Action Criteria Definitions) For the description below, we will call a single sequential set of waypoints in a single rollout as one "action set". For each tool you created, the goal is to generate $n_{action}$ action sets that optimize the task success and motion differentiation. Task success is optimized when an action set is able to complete the task successfully. Motion differentiation is optimized when there exists a large variance in the motion taken across all action sets you design for the same tool. A large variance in motion is defined the tool, at each time step, is located at a different location in the 3D space. Think about how a tool can be used to interact with the object from many different sides, angles, and ways. When both conditions are met, you have successfully designed a good set of actions sets.}
% \end{tcolorbox}
\end{mdframed}

\subsubsection{Frame Clarifications}\label{app:prompts_frame}
% \begin{tcolorbox}[colback=gray!5!white, colframe=gray!75!black, boxrule=0.5pt]
\begin{mdframed}[
  backgroundcolor=gray!5,  % instead of colback
  linecolor=gray!75,       % instead of colframe
  linewidth=0.5pt,         % instead of boxrule
  skipabove=6pt, 
  skipbelow=6pt, 
]
\small
\texttt{(Frame Clarification) In the world frame, front/back is along the x axis, left/right is along the y axis, and up/down is along the z axis with the following directions: Positive x: Towards the front of the table. Negative x: Towards the back of the table. Positive y: Towards the left. Negative y: Towards the right. Positive z: Up, towards the ceiling. Negative z: Down, towards the floor. In terms of orientation, starting from the origin frame, Positive rotation about the x-axis: tilting the end-effector head to the left. Negative rotation about the x-axis: tilting the end-effector head to the right. Positive rotation about the y-axis: tilting the end-effector head down. Negative rotation about the y-axis: tilting the end-effector head up. Positive rotation about the z-axis: rotating the end-effector head counter-clockwise. Negative rotation about the z-axis: rotating the end-effector head clockwise.}
% \end{tcolorbox}
\end{mdframed}

\begin{figure*}[h!]
    \centering
    \includegraphics[width=0.8\linewidth]{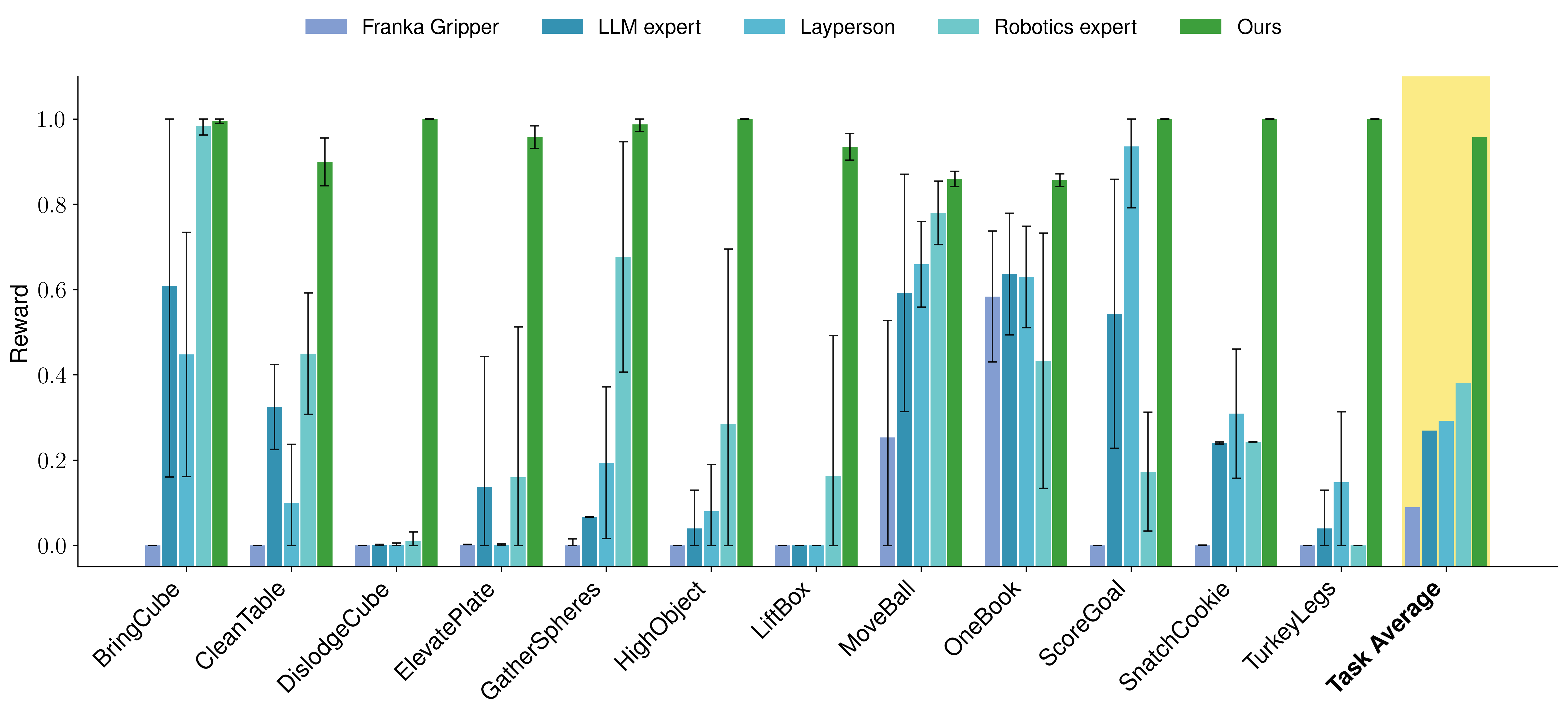}
    \caption{Comparison of the mean and standard deviation of reward generated by \methodname, human-prompted designs, and Franka Gripper across \NumBenchmarkTasks tasks. Error bars represent standard deviation (clipped to the range $[0,1]$).}
    \label{fig:avg_reward_with_error}
\end{figure*}

\subsubsection{Evolutionary Instructions} \label{app:prompts_evo}
% \begin{tcolorbox}[colback=gray!5!white, colframe=gray!75!black, boxrule=0.5pt]
\begin{mdframed}[
  backgroundcolor=gray!5,  % instead of colback
  linecolor=gray!75,       % instead of colframe
  linewidth=0.5pt,         % instead of boxrule
  skipabove=6pt, 
  skipbelow=6pt, 
]
\small
\texttt{(Evolutionary Process) Your design decision is a part of a tool design genetic algorithm. For each of the $n_{tool}$ tool designs, you can choose to either mutate or crossover. Specifically, tool mutation is defined as one change to a single randomly selected previous tool design. Mutation changes include:
\begin{itemize}[label={}]
  \item (1) Changing the dimension, location, or orientation of a single component of the tool.
  \item (2) Adding, removing, or replacing a single component of the tool.
\end{itemize}
Crossover is defined as the process of combining two randomly selected previous tool designs to create a new tool design. Combination is defined as:
\begin{itemize}[label={}]
  \item (1) Selecting components from two previous tool designs and combining them to form a new tool design. 
\end{itemize}
All mutation and crossover decisions must potentially increase the likelihood of task success, yet all decisions must be different and diverse.}
% \end{tcolorbox}
\end{mdframed}

\subsubsection{No Tool Instructions}\label{no_tool_instruct}
% \begin{tcolorbox}[colback=gray!5!white, colframe=gray!75!black, boxrule=0.5pt]
\begin{mdframed}[
  backgroundcolor=gray!5,  % instead of colback
  linecolor=gray!75,       % instead of colframe
  linewidth=0.5pt,         % instead of boxrule
  skipabove=6pt, 
  skipbelow=6pt, 
]
\small
\texttt{You are a robotics hardware and controls expert. You operate with boldness and brilliance in the physical realm. You work with a robot arm that sits in the origin of your environment. You will be presented with some robotic tasks, and will be asked to design actions to complete the task. \\
\\
...}
% \end{tcolorbox}
\end{mdframed}
The complete prompt is composed together with instructions from \ref{app:prompts_procedure}, \ref{app:prompts_action}, \ref{app:prompt_action_diversity}, and \ref{app:prompts_frame}.\\

\subsubsection{Human Specification Instructions}\label{human_spec_instruct}
% \begin{tcolorbox}[colback=gray!5!white, colframe=gray!75!black, boxrule=0.5pt]
\begin{mdframed}[
  backgroundcolor=gray!5,  % instead of colback
  linecolor=gray!75,       % instead of colframe
  linewidth=0.5pt,         % instead of boxrule
  skipabove=6pt, 
  skipbelow=6pt, 
]
\small
\texttt{You are a helpful robotics hardware and controls expert. You have a robot arm that sits in the origin of your environment. You are working with a colleague as a team to design tools and actions for a robot to complete a task. Your colleague will provide you with a design and action instructions in the form of natural language instructions. Your goal is to use your colleague's design and action instructions to output URDF and action waypoints for the robot to use. You should not use your own knowledge to design the tool and action, but rather follow your human colleague's instruction.
Here is the human colleague's prompt: \{human\_prompt\} \\
\\
...}
% \end{tcolorbox}
\end{mdframed}
The complete prompt is composed together with instructions from \ref{app:prompts_procedure},
\ref{app:prompts_tool},
\ref{app:prompts_action}, \ref{app:prompt_action_diversity}, and \ref{app:prompts_frame}.\\

\subsubsection{RLBench Instructions}\label{app:rlbench_instruct}
% \begin{tcolorbox}[colback=gray!5!white, colframe=gray!75!black, boxrule=0.5pt]
\begin{mdframed}[
  backgroundcolor=gray!5,  % instead of colback
  linecolor=gray!75,       % instead of colframe
  linewidth=0.5pt,         % instead of boxrule
  skipabove=6pt, 
  skipbelow=6pt, 
]
\small
\texttt{You are a helpful robotics hardware and controls expert. You have a robot arm that sits in the origin of your environment. You are working with a colleague as a team to design tools and actions for a robot to complete a task. Your colleague will provide you with a design in the format of a URDF, which is attached for you as tool.txt. Your goal is to use your colleague's URDF to come up with an action plan for the robot to use. \\
\\
...}
% \end{tcolorbox}
\end{mdframed}
The complete prompt is composed together with instructions from \ref{app:prompts_procedure}, \ref{app:prompts_action}, \ref{app:prompt_action_diversity}, and \ref{app:prompts_frame}.

\subsection{Statistical Significance Analysis}

% Fig.~\ref{fig:avg_reward_with_error} shows our main quantitative results presented with standard deviation across runs. Across \NumBenchmarkTasks tasks, \methodname not only consistently outperforms all baselines, but also consistently has minimal variations across runs, demonstrating that it can produce performant tool and action designs with high stability.

Fig.~\ref{fig:avg_reward_with_error} presents our primary quantitative results, including standard deviations across 5 runs. Across all \NumBenchmarkTasks tasks, \methodname consistently surpasses all baselines, exhibiting notably low variation between trials. This indicates that \methodname reliably produces high-performing and stable tool-action designs.

% In contrast, results from human prompts not only are much less performant but also show a considerably larger variation across runs. We hypothesize that this is due to the following reasons. First, the human tool design solutions tend to be still require more complex control strategies even if they can complete the task, and thus they are less robust to imperfect and suboptimal control strategies. In contrast, \methodname's tool designs usually are more robust towards action imperfections. Additionally, sometimes there is misalignment in terms of design specifications and physical understanding between the human and the VLM: humans sometimes also specify designs that are difficult to accurately or fully materialized or represented by the VLM. \methodname's automated design, on the other hand, does not suffer from this human-VLM misalignment issue. 

In contrast, results obtained from human-prompted designs not only yield significantly lower performance but also show greater variations across runs. We attribute this discrepancy to several factors. First, human-specified tools often require more intricate control strategies; even if capable of completing the task, these designs tend to be less resilient to suboptimal or imperfect executions. By comparison, \methodname-generated tools typically exhibit greater robustness to action imperfections. Second, human prompts sometimes suffer from specification ambiguity or misalignment with the VLM. There can be discrepancies between human intent and the VLM’s internal representation and physical modeling capabilities. By automating the design process, \methodname avoids these alignment issues, resulting in more effective and precisely realizable solutions.

\subsection{Tool Design Gallery} 

Table~\ref{tab:tool_gallery_part1} and~\ref{tab:tool_gallery_part2} show our tool gallery. In this tool design gallery, we take the opportunity to display tools from a few tasks that seemed to have allowed \methodname the most creative freedom. These are tool designs that are not presented elsewhere in the paper. We believe this illustrates \methodname's impressive physical creativity and problem-solving capabilities.

\begin{table*}[!t]
  \centering
  \setlength\tabcolsep{1pt}
  \renewcommand{\arraystretch}{0.5}
  \footnotesize
  \begin{tabular}{ M{2.6cm}  M{0.72\textwidth} }
    \toprule
      \textbf{Task Name}
    & \textbf{Example Tool Designs} \\
    \midrule

    \textsc{BringCube}
    &
      \begin{tabular}{*{3}{M{0.24\textwidth}}}
        \includegraphics[width=\linewidth]{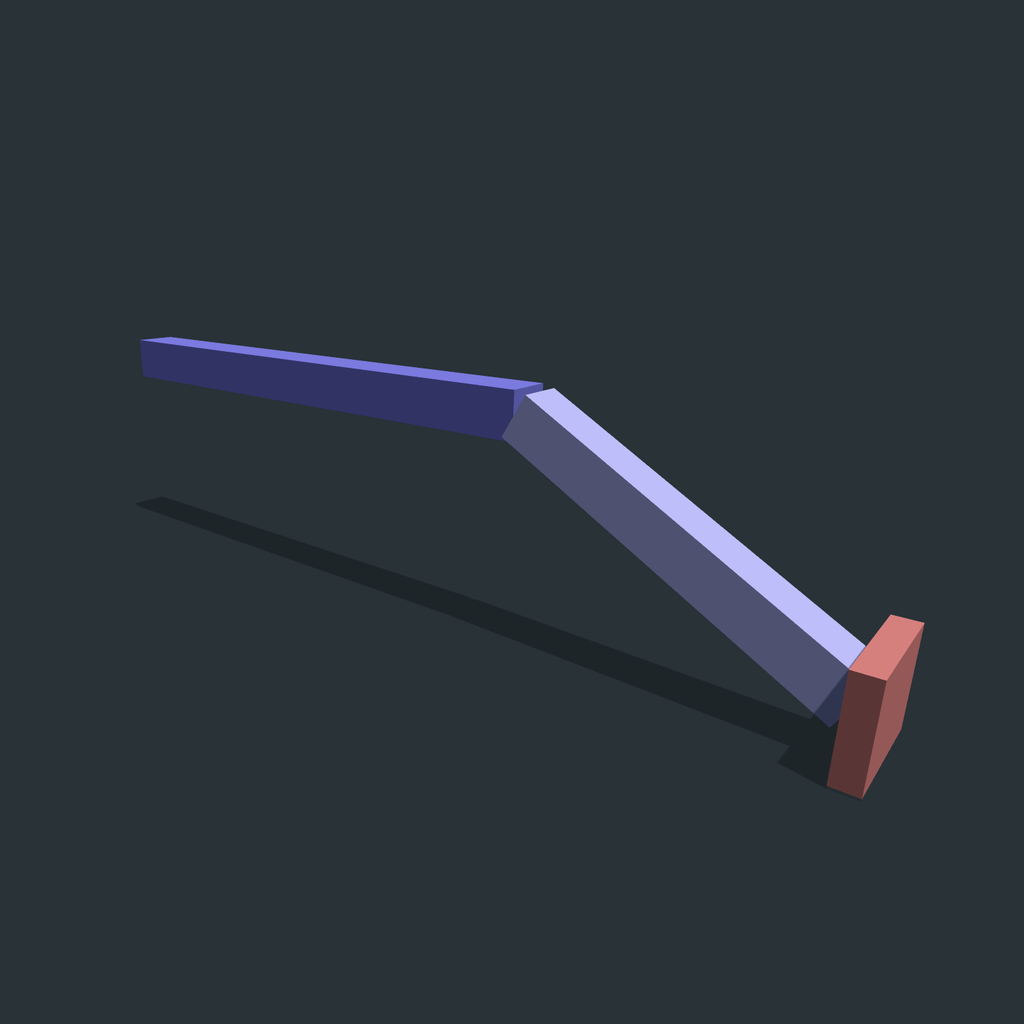} &
        \includegraphics[width=\linewidth]{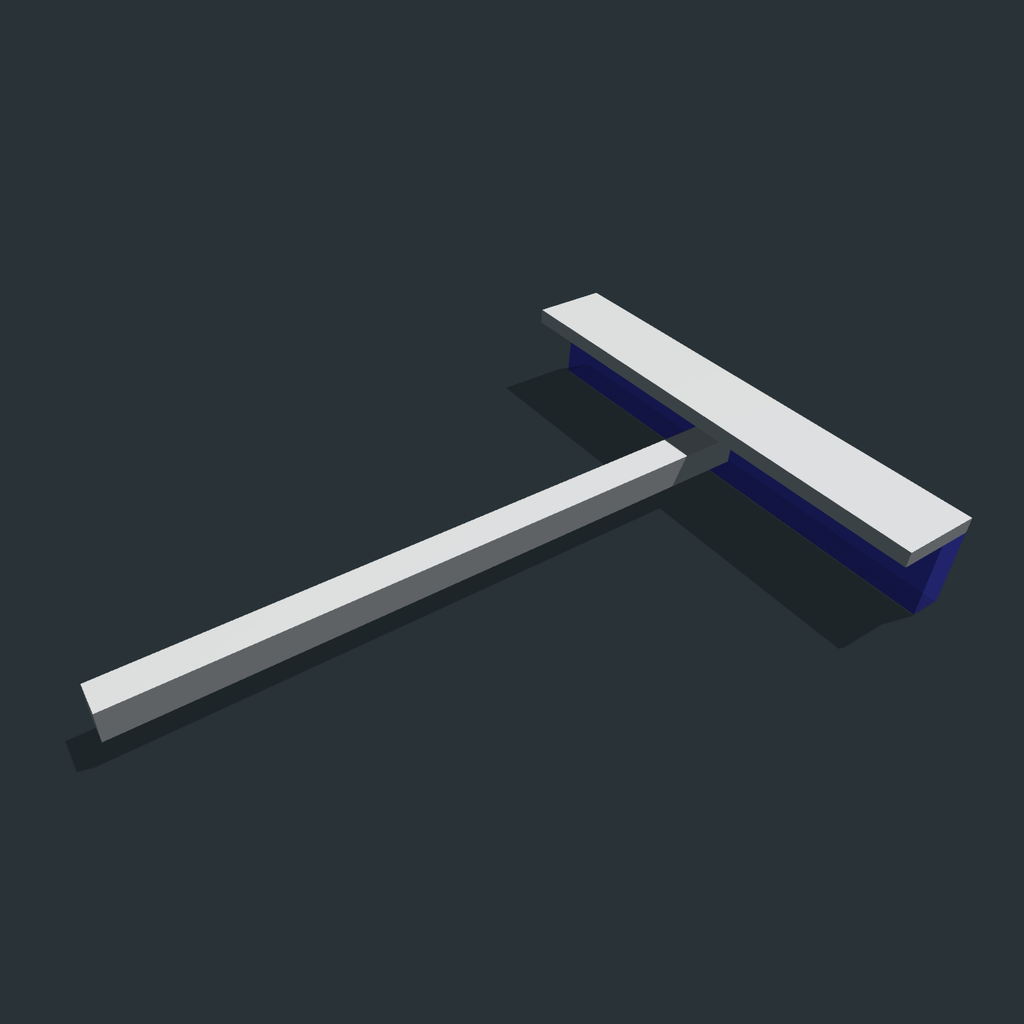} &
        \includegraphics[width=\linewidth]{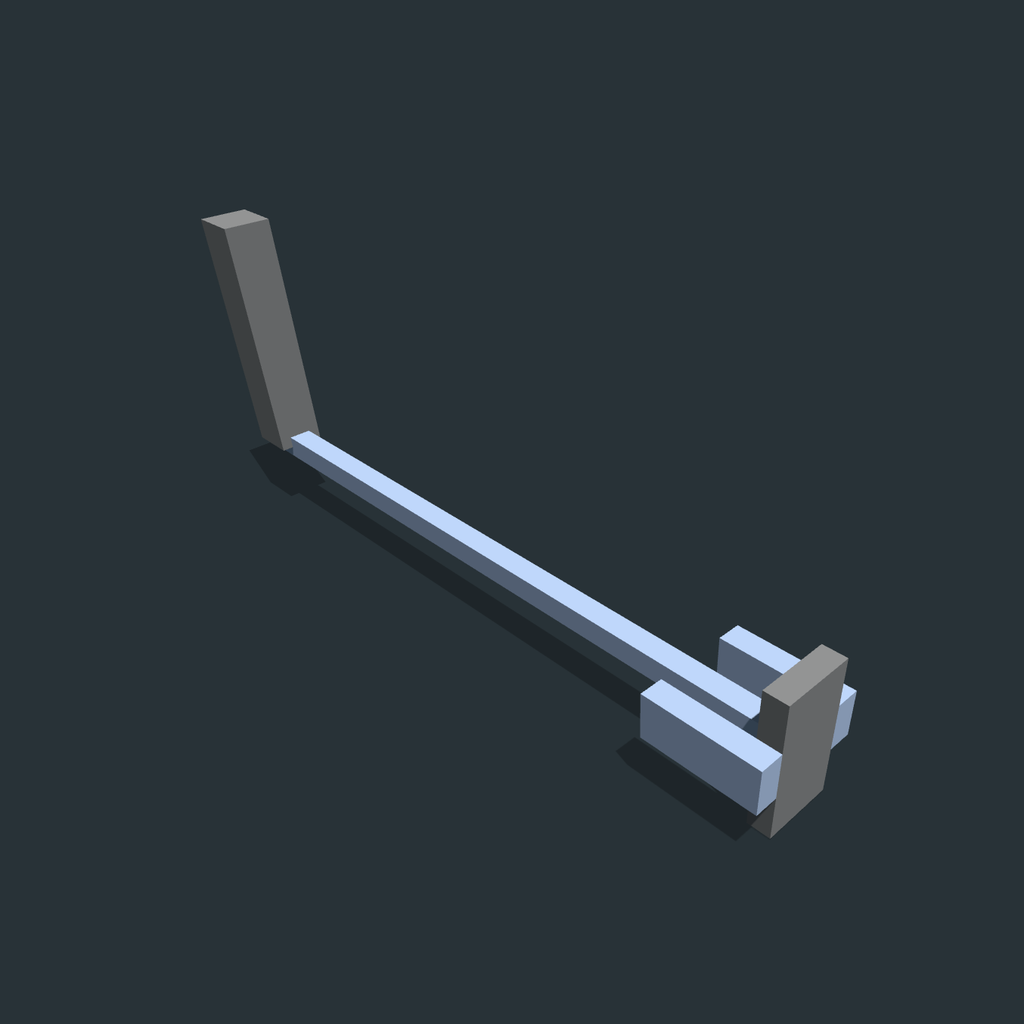} \\
      \end{tabular}
    \\[6pt]\midrule

    \textsc{CleanTable}
    &
      \begin{tabular}{*{3}{M{0.24\textwidth}}}
        \includegraphics[width=\linewidth]{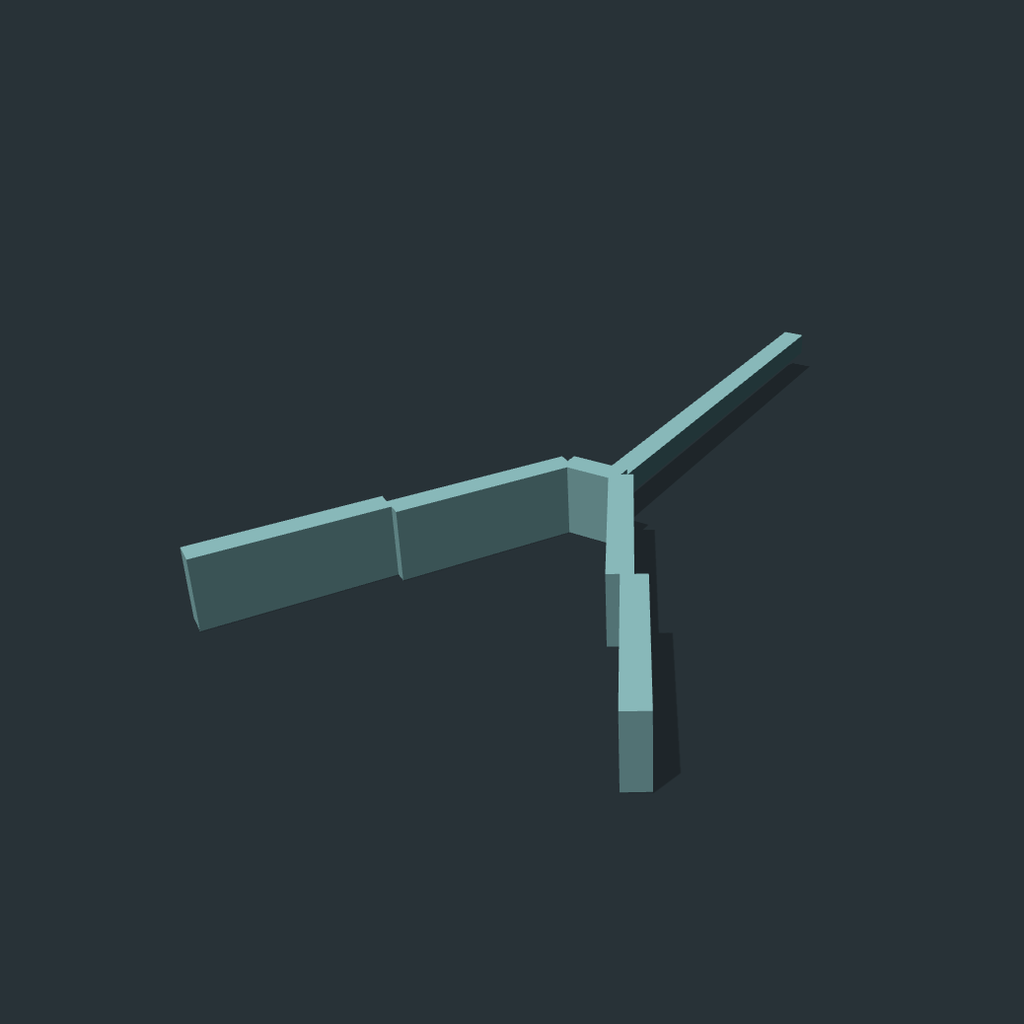} &
        \includegraphics[width=\linewidth]{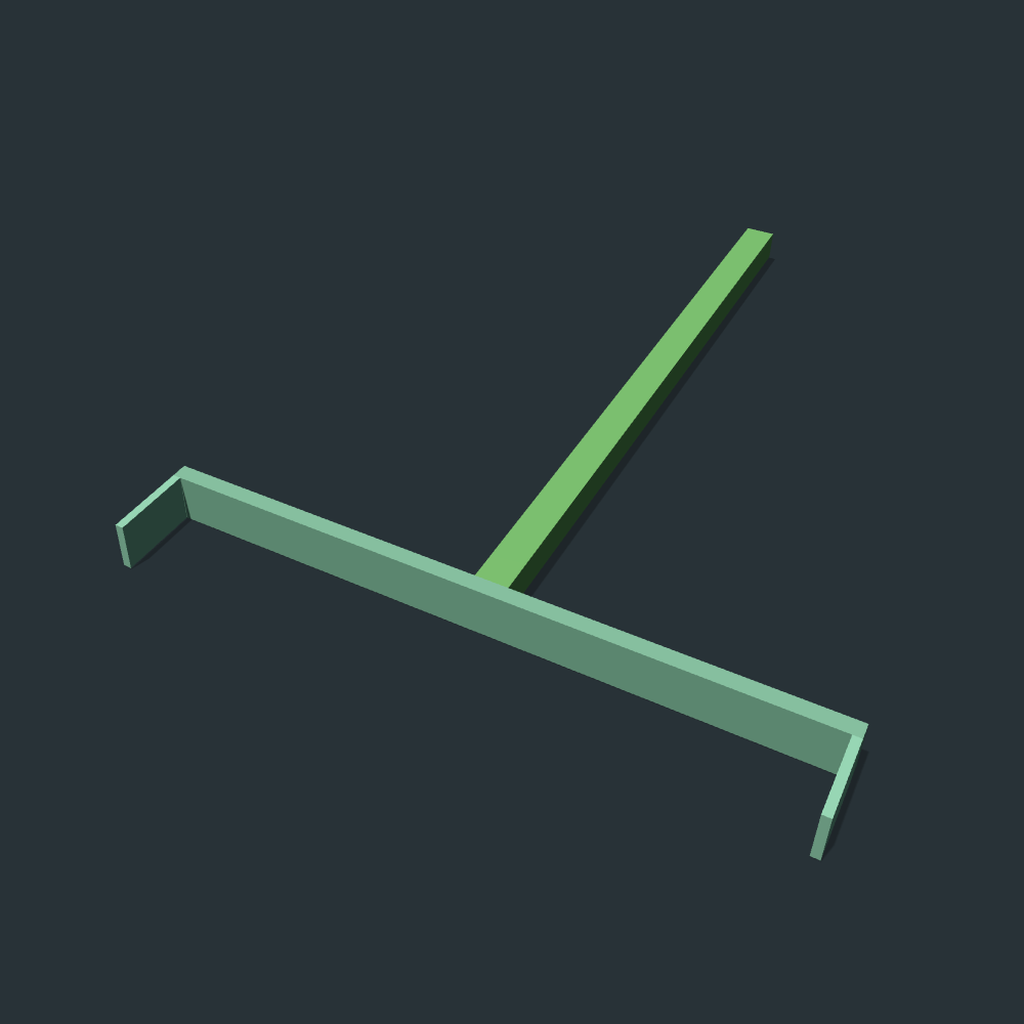} &
        \includegraphics[width=\linewidth]{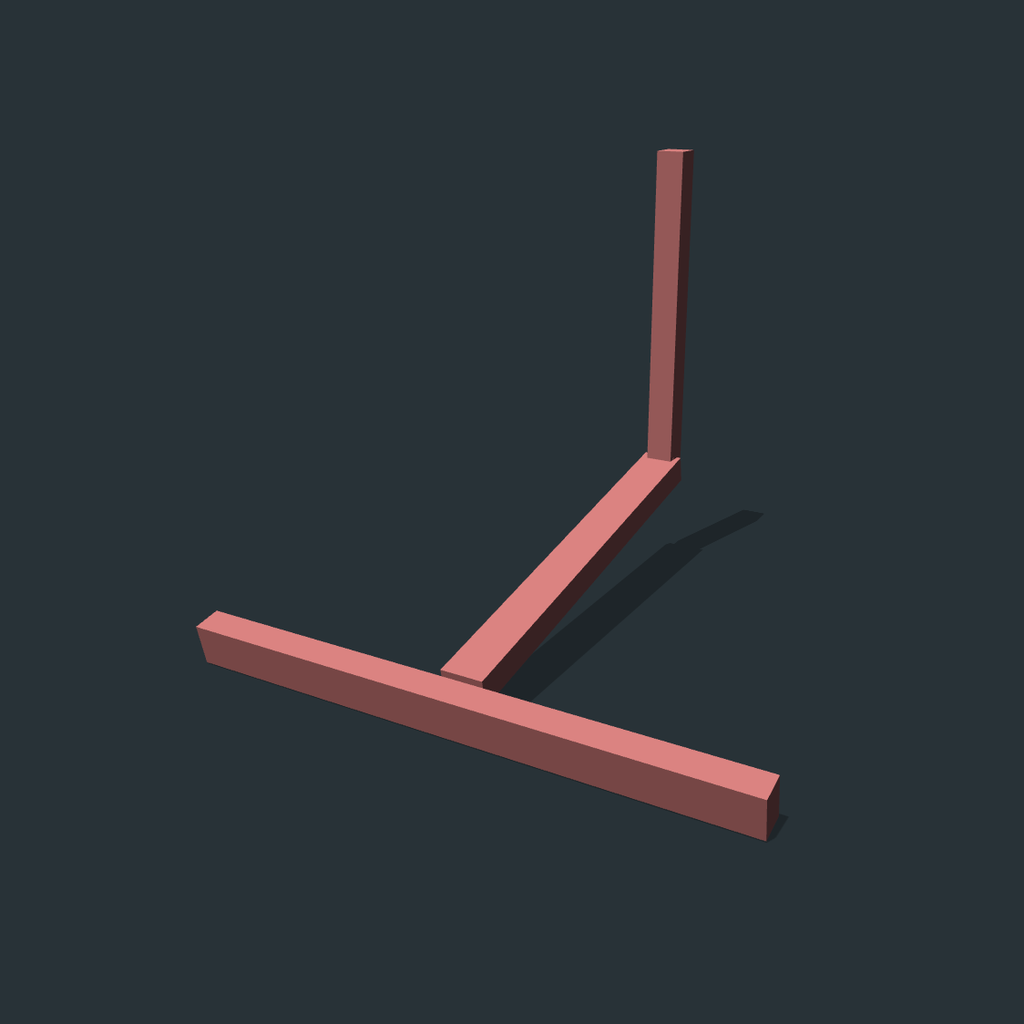} \\
      \end{tabular}
    \\[6pt]\midrule

    \textsc{ElevatePlate}
    &
      \begin{tabular}{*{3}{M{0.24\textwidth}}}
        \includegraphics[width=\linewidth]{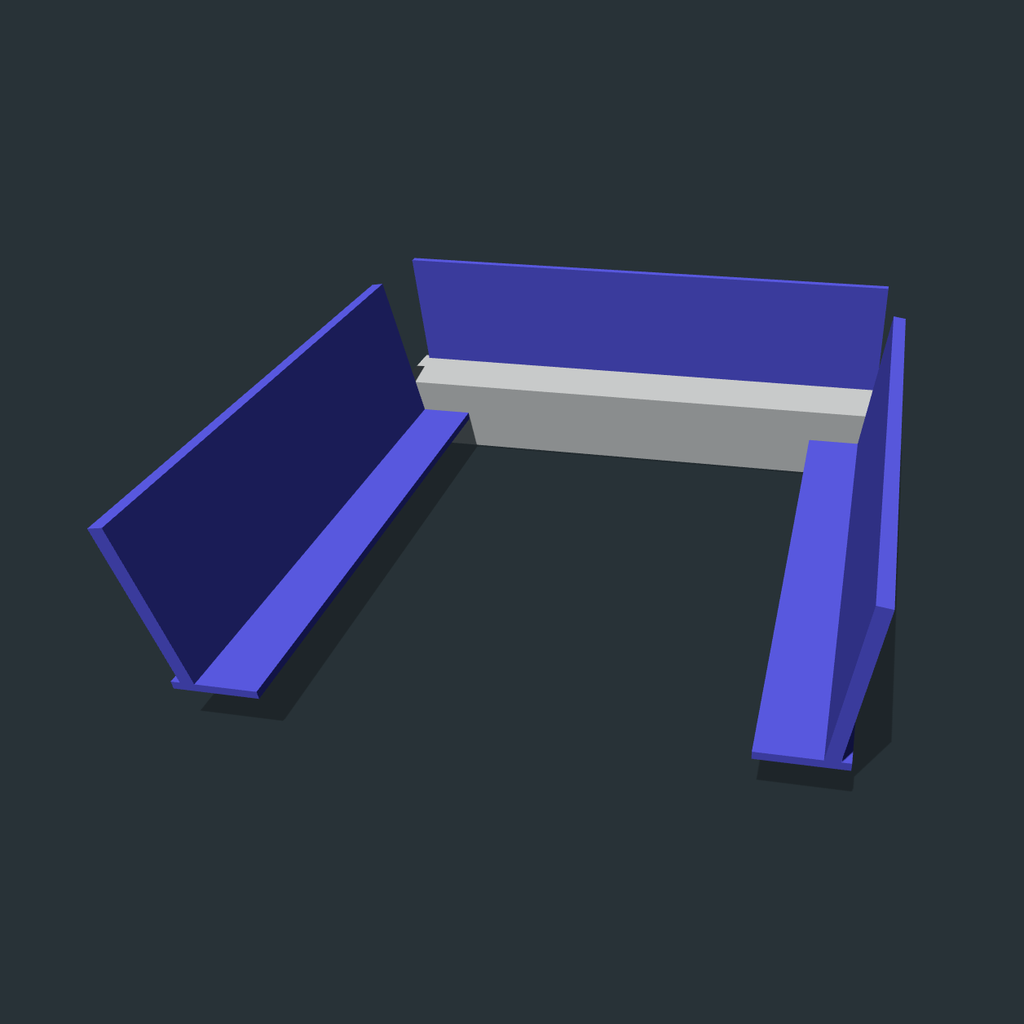} &
        \includegraphics[width=\linewidth]{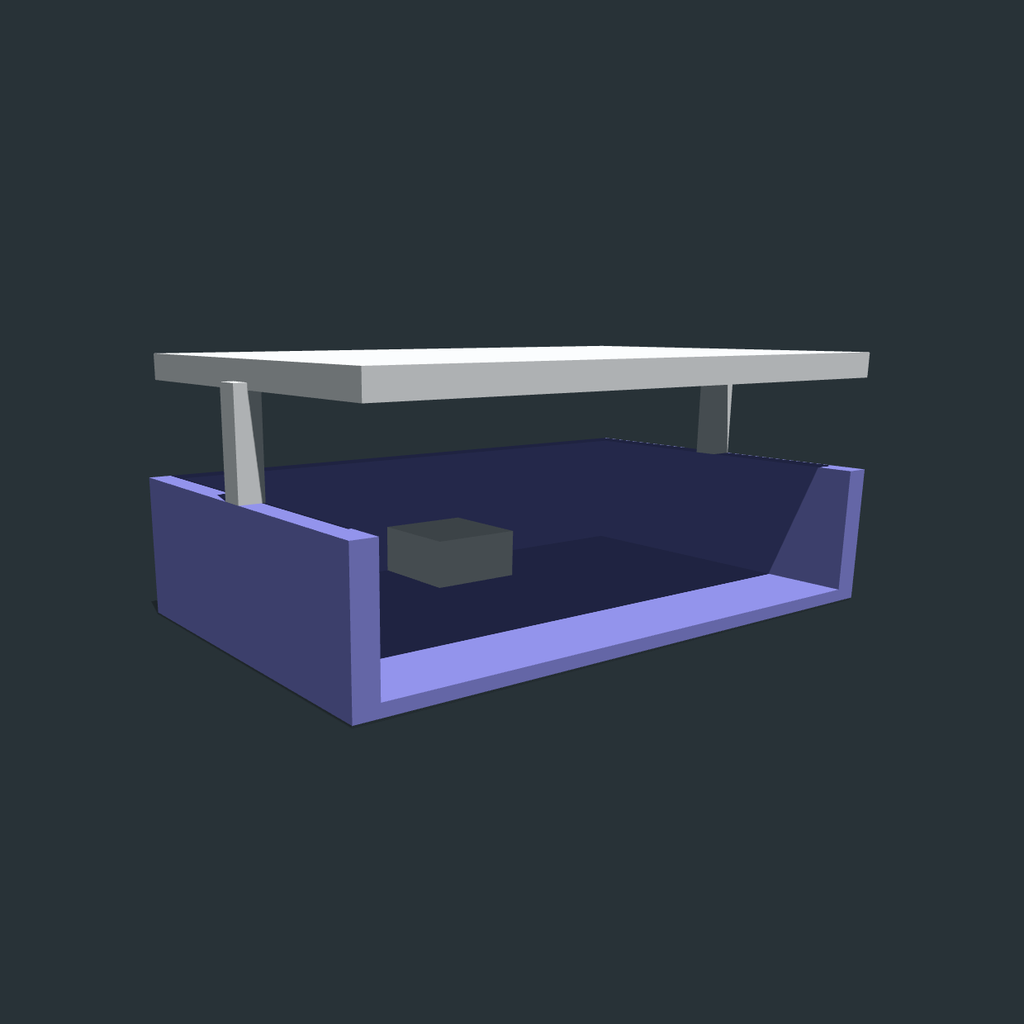} &
        \includegraphics[width=\linewidth]{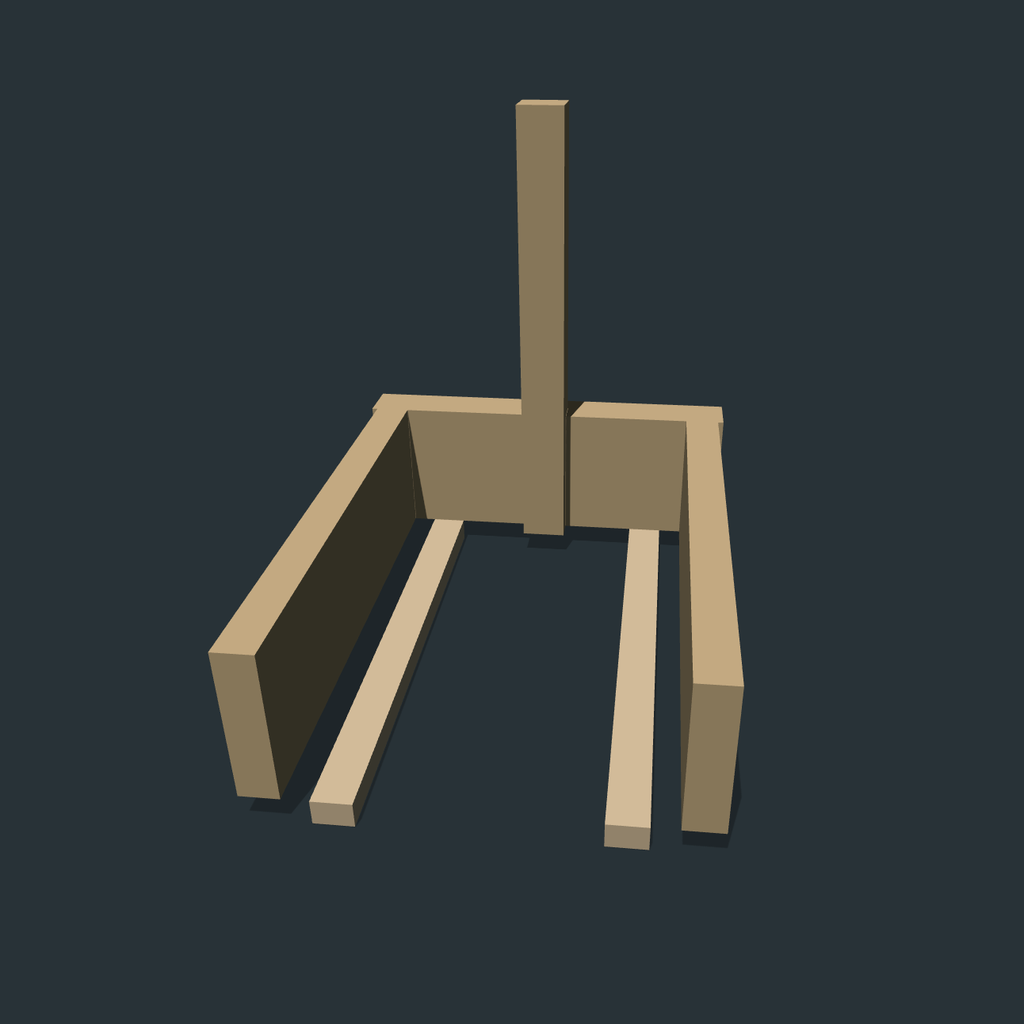} \\[-2pt]
        \includegraphics[width=\linewidth]{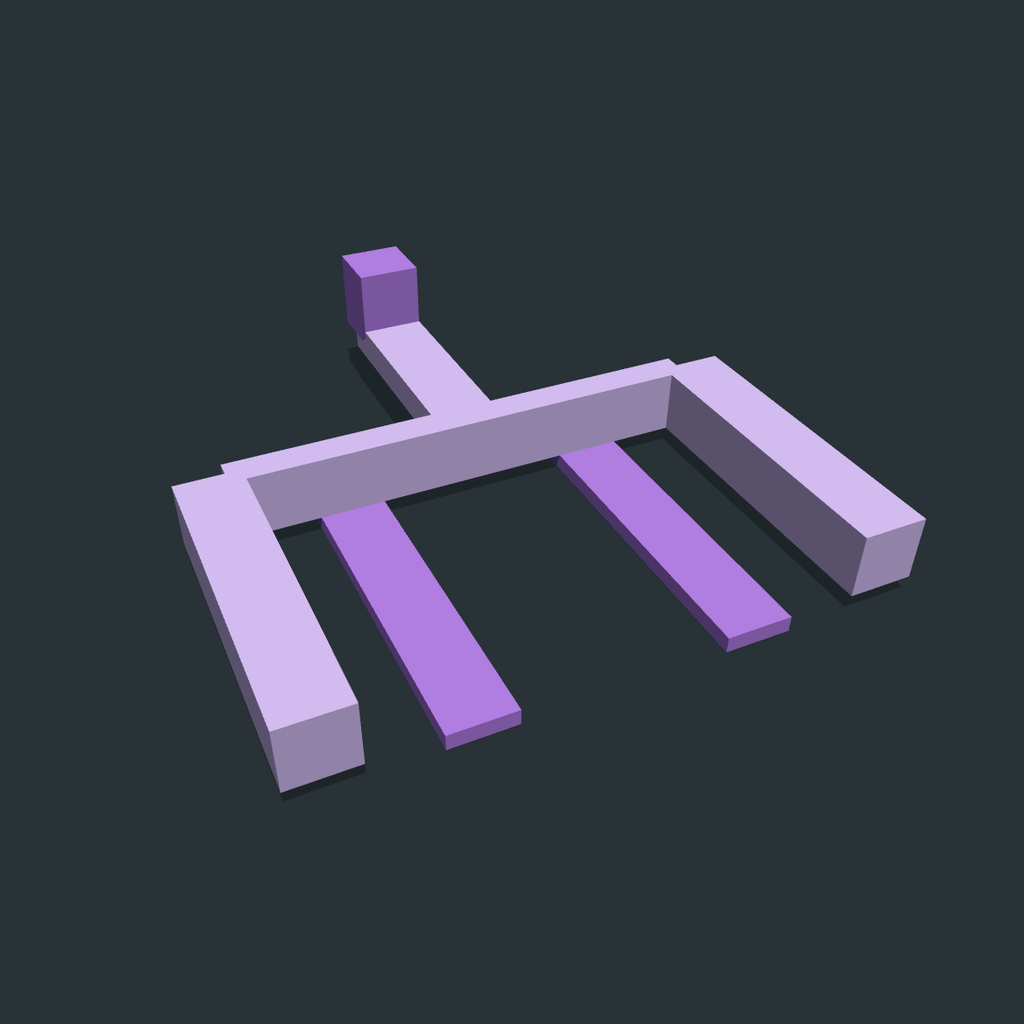} &
        \includegraphics[width=\linewidth]{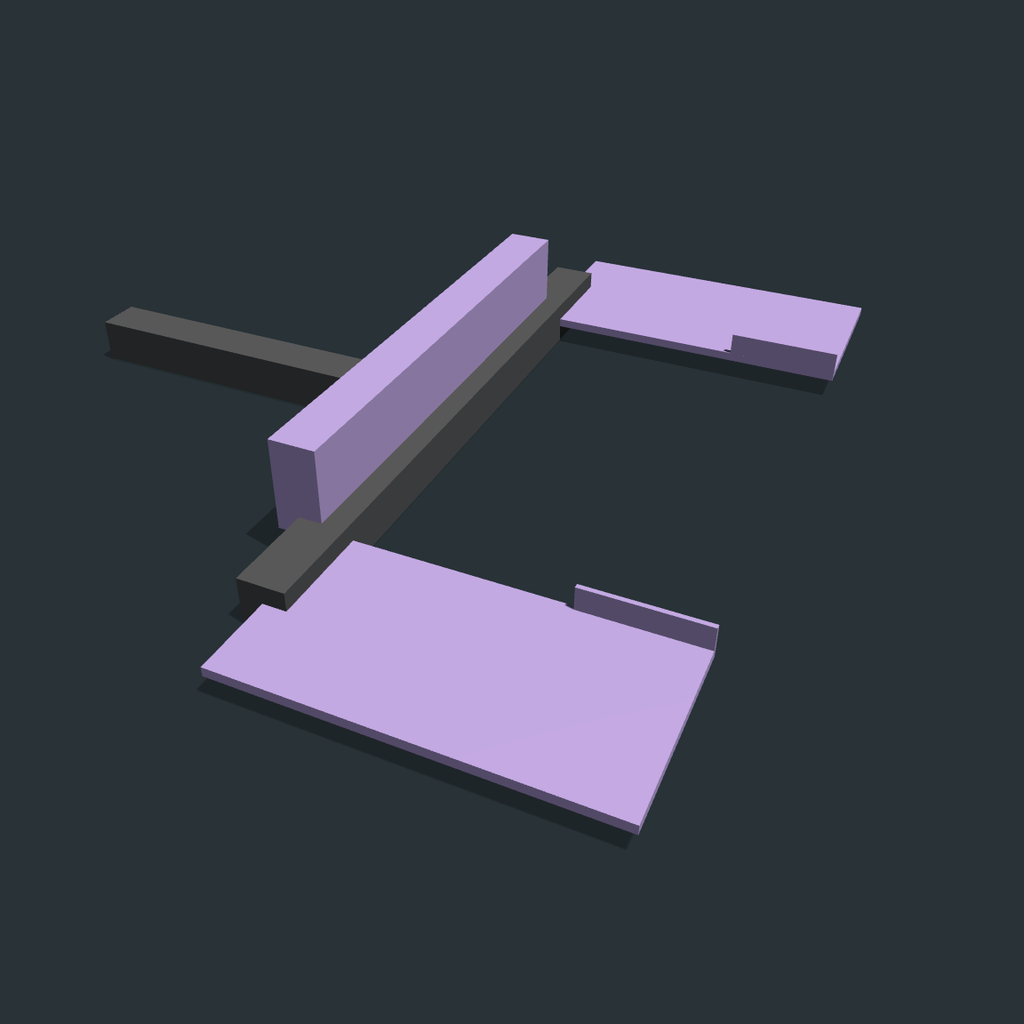} &
        \includegraphics[width=\linewidth]{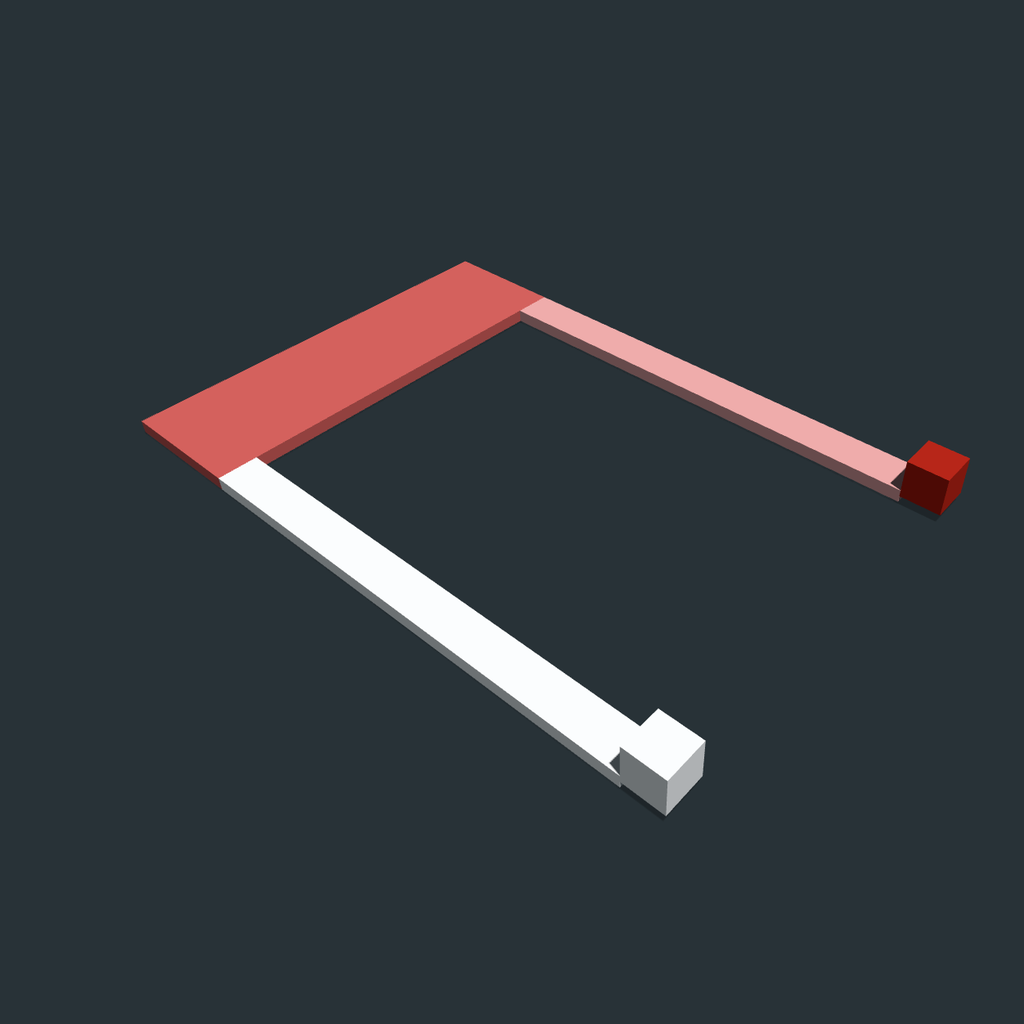} \\
      \end{tabular}
    \\

    \bottomrule
  \end{tabular}
  \caption{Tool-gallery for BringCube, CleanTable, and ElevatePlate.}
  \label{tab:tool_gallery_part1}
\end{table*}

%––––––––––––––––––––––––––––––––––––––––––––––––––––––––––––
% Table 2: GatherSpheres, MoveBall
\begin{table*}[!t]
  \centering
  \setlength\tabcolsep{1pt}
  \renewcommand{\arraystretch}{0.5}
  \footnotesize
  \begin{tabular}{ M{2.6cm}  M{0.72\textwidth} }
    \toprule
      \textbf{Task Name}
    & \textbf{Example Tool Designs} \\
    \midrule

    \textsc{GatherSpheres}
    &
      \begin{tabular}{*{3}{M{0.24\textwidth}}}
        \includegraphics[width=\linewidth]{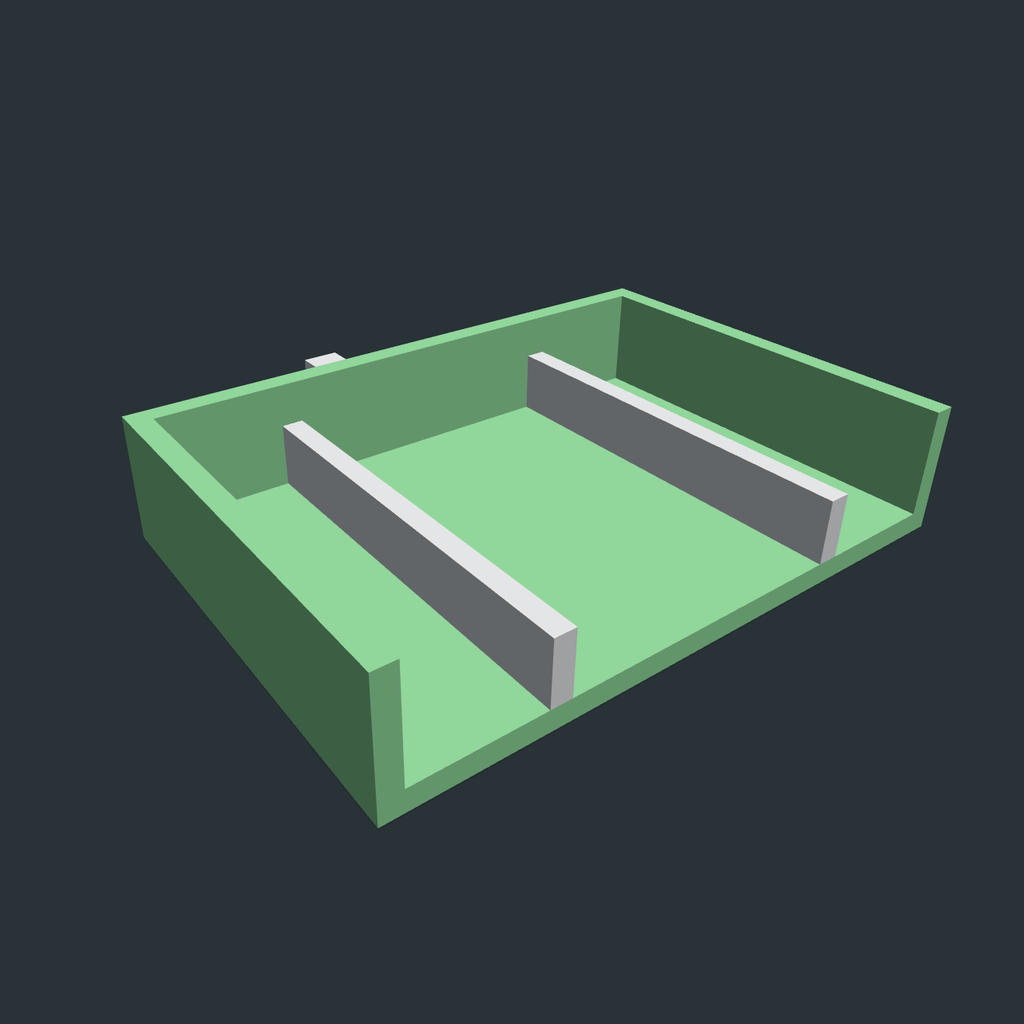} &
        \includegraphics[width=\linewidth]{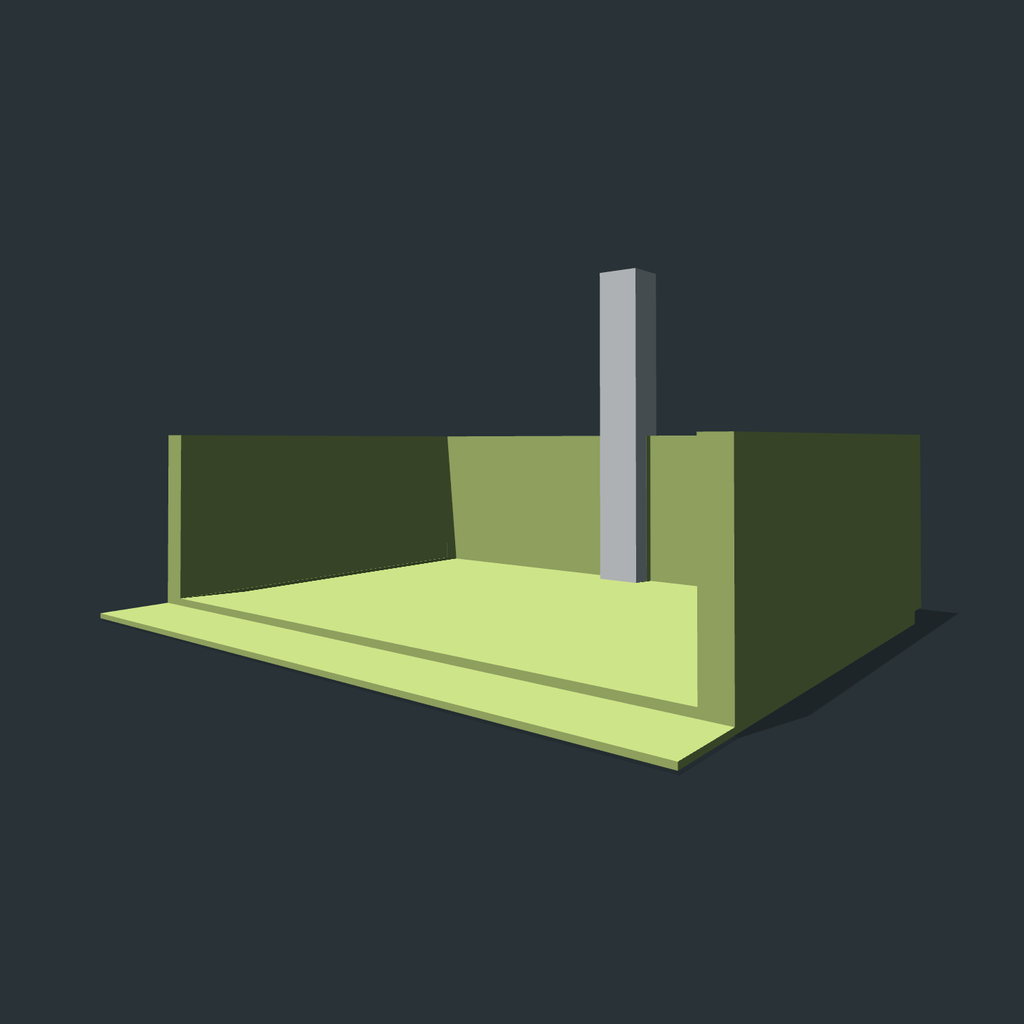} &
        \includegraphics[width=\linewidth]{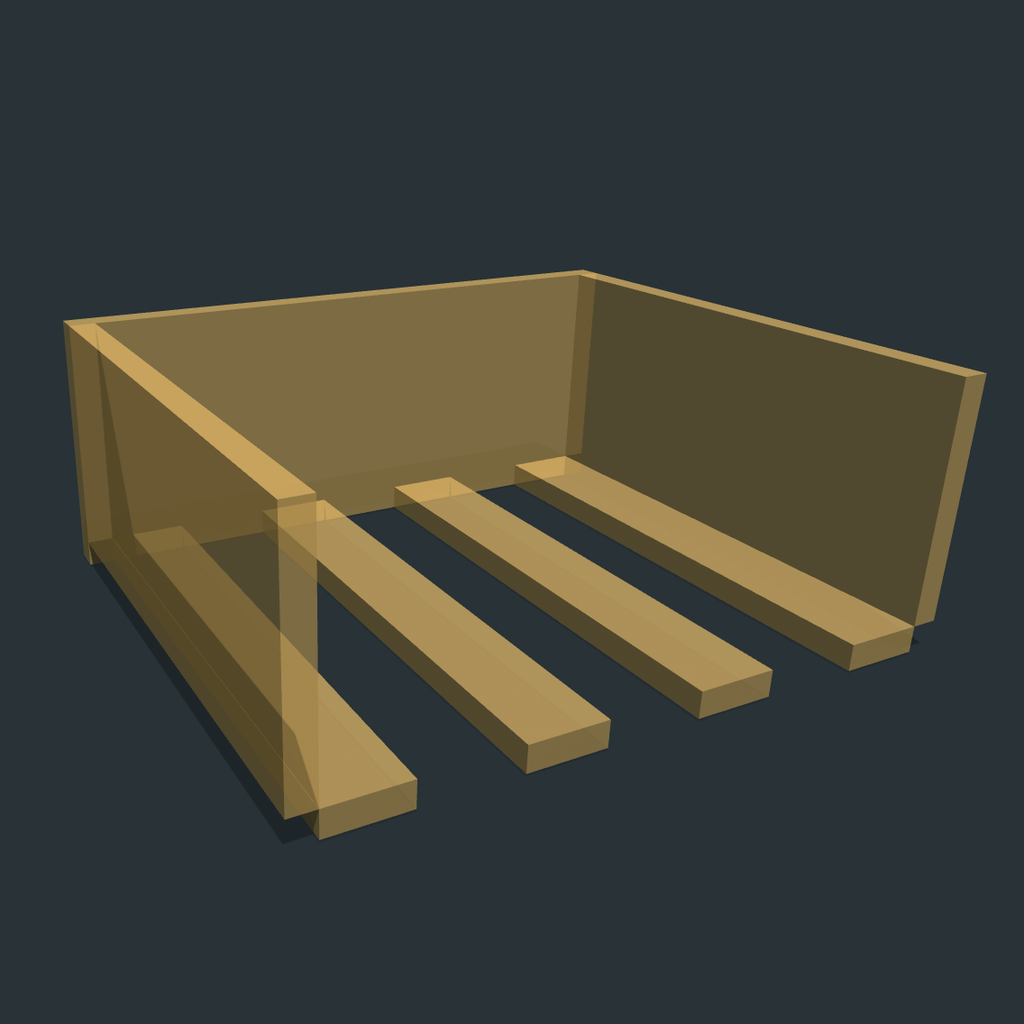} \\
      \end{tabular}
    \\[6pt]\midrule

    \textsc{MoveBall}
    &
      \begin{tabular}{*{3}{M{0.24\textwidth}}}
        \includegraphics[width=\linewidth]{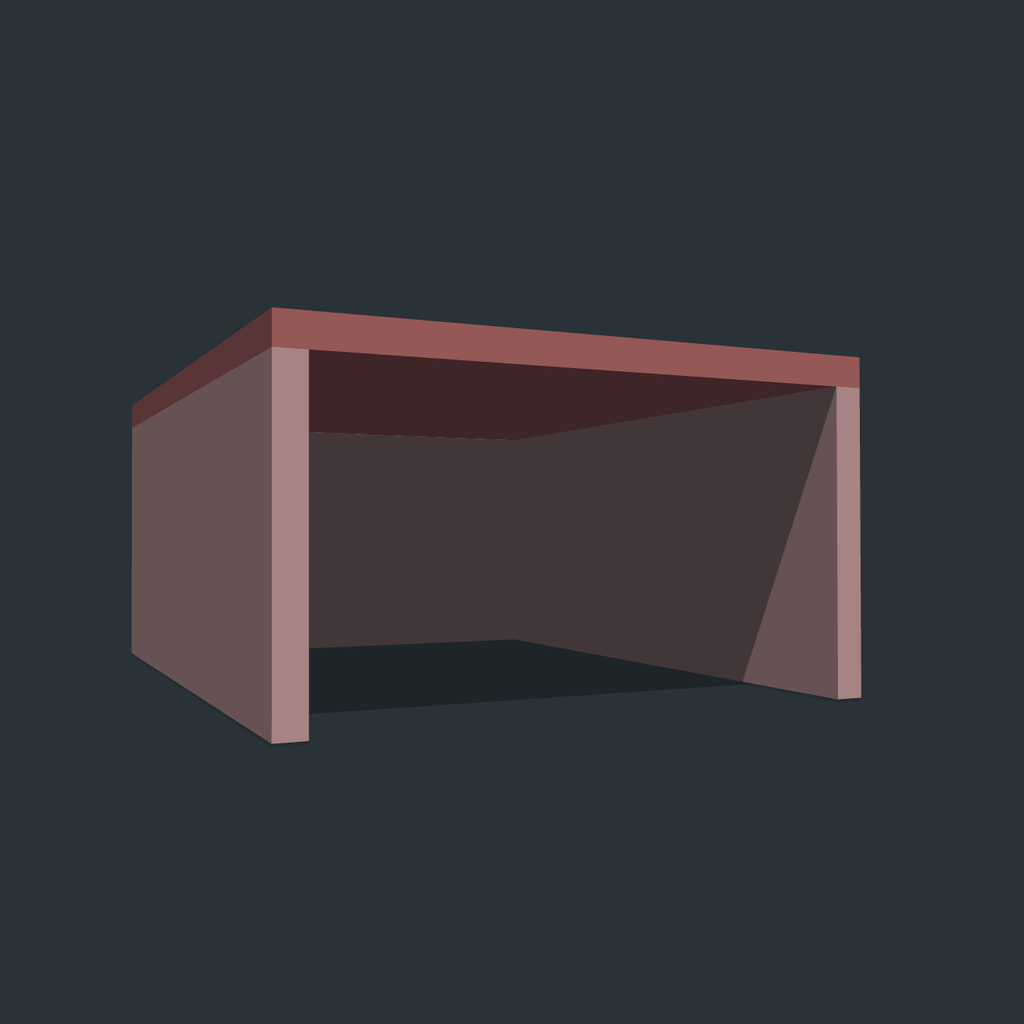} &
        \includegraphics[width=\linewidth]{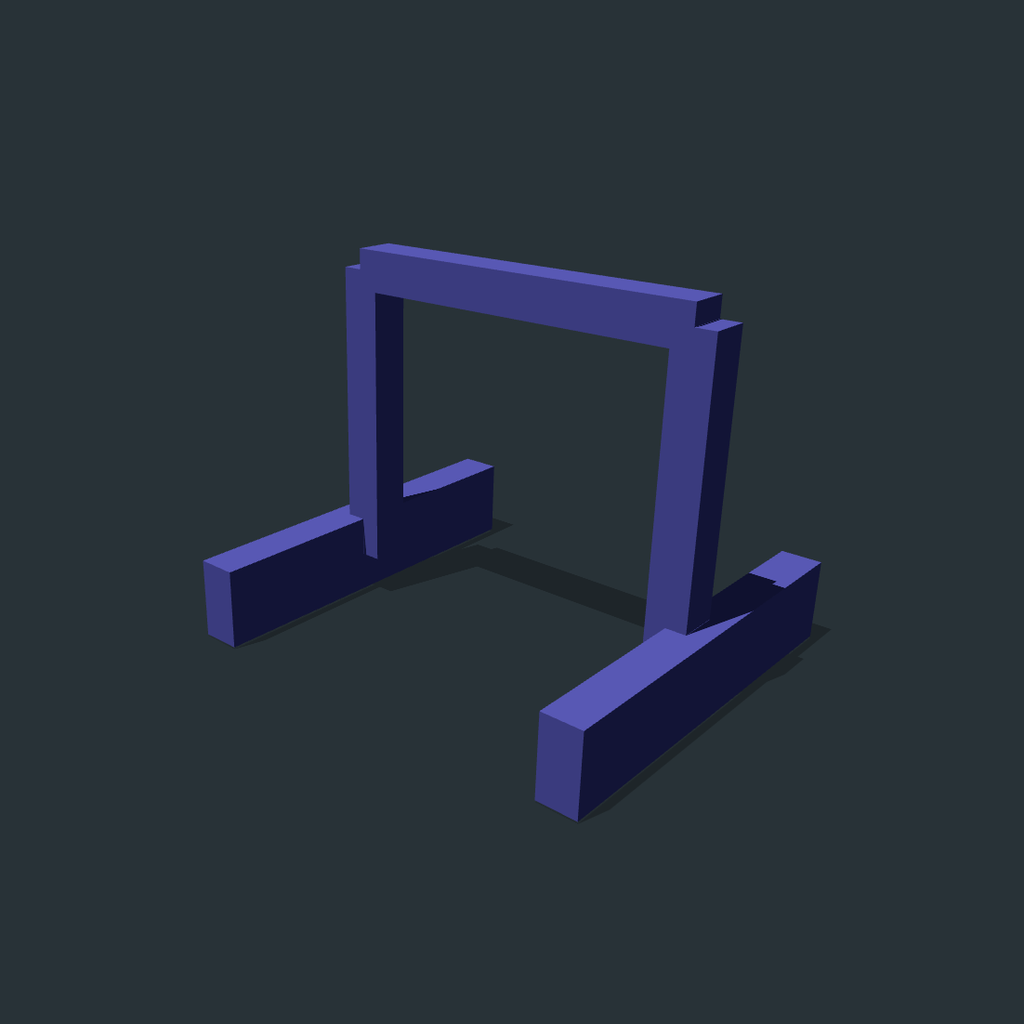} &
        \includegraphics[width=\linewidth]{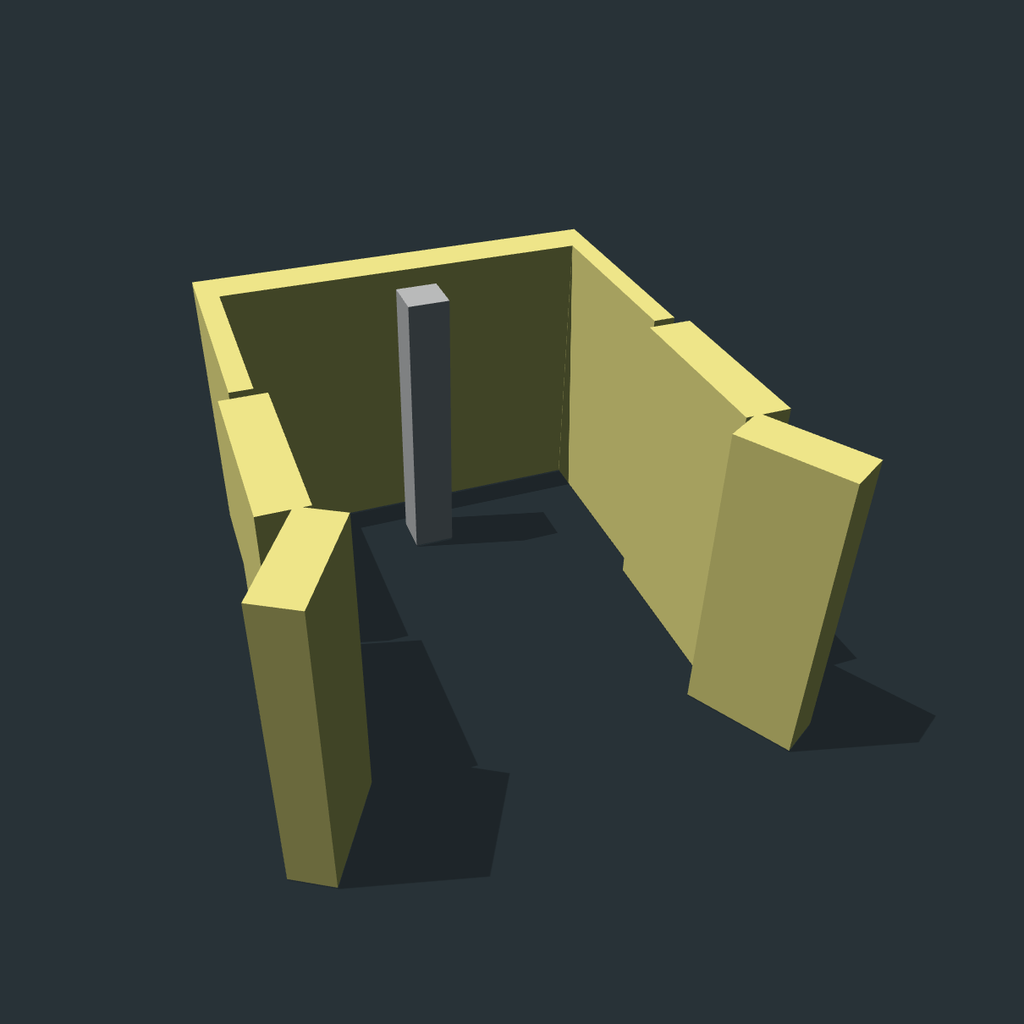} \\[-2pt]
        \includegraphics[width=\linewidth]{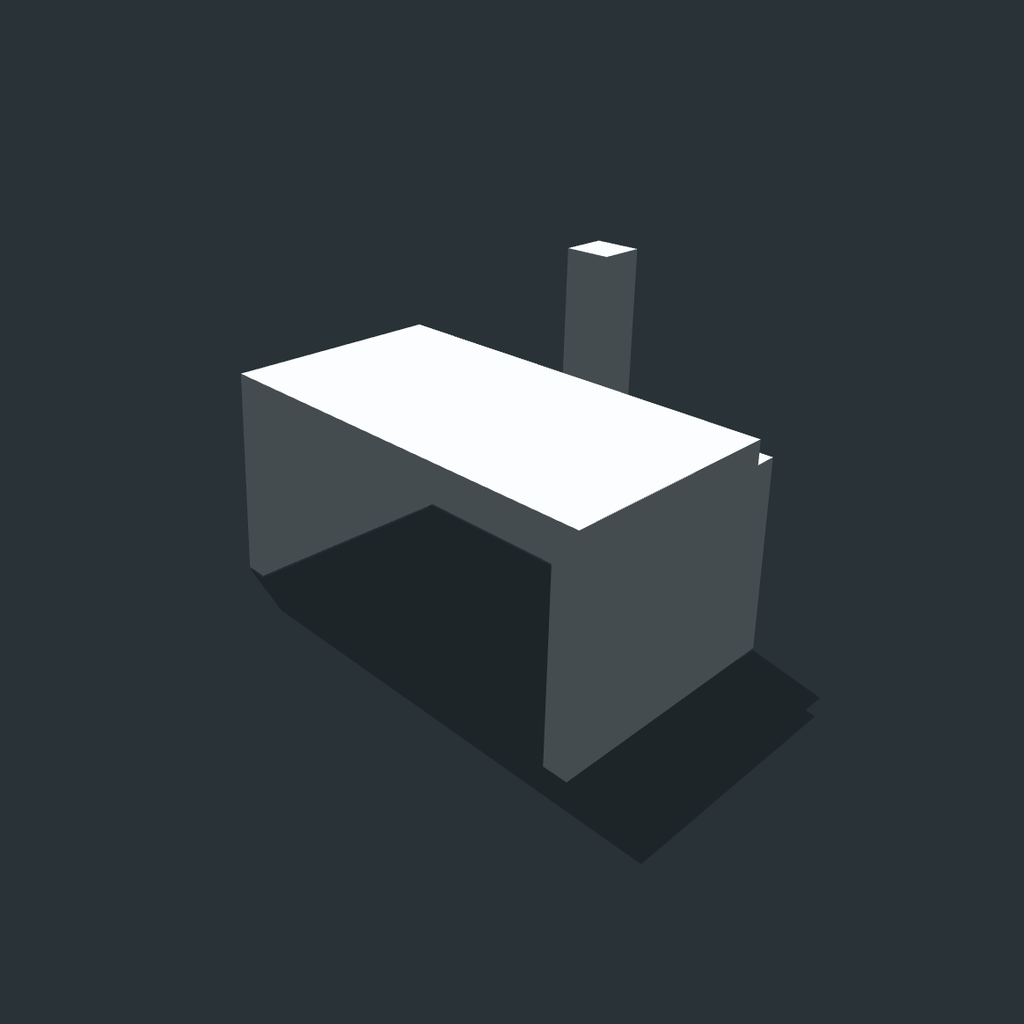} &
        \includegraphics[width=\linewidth]{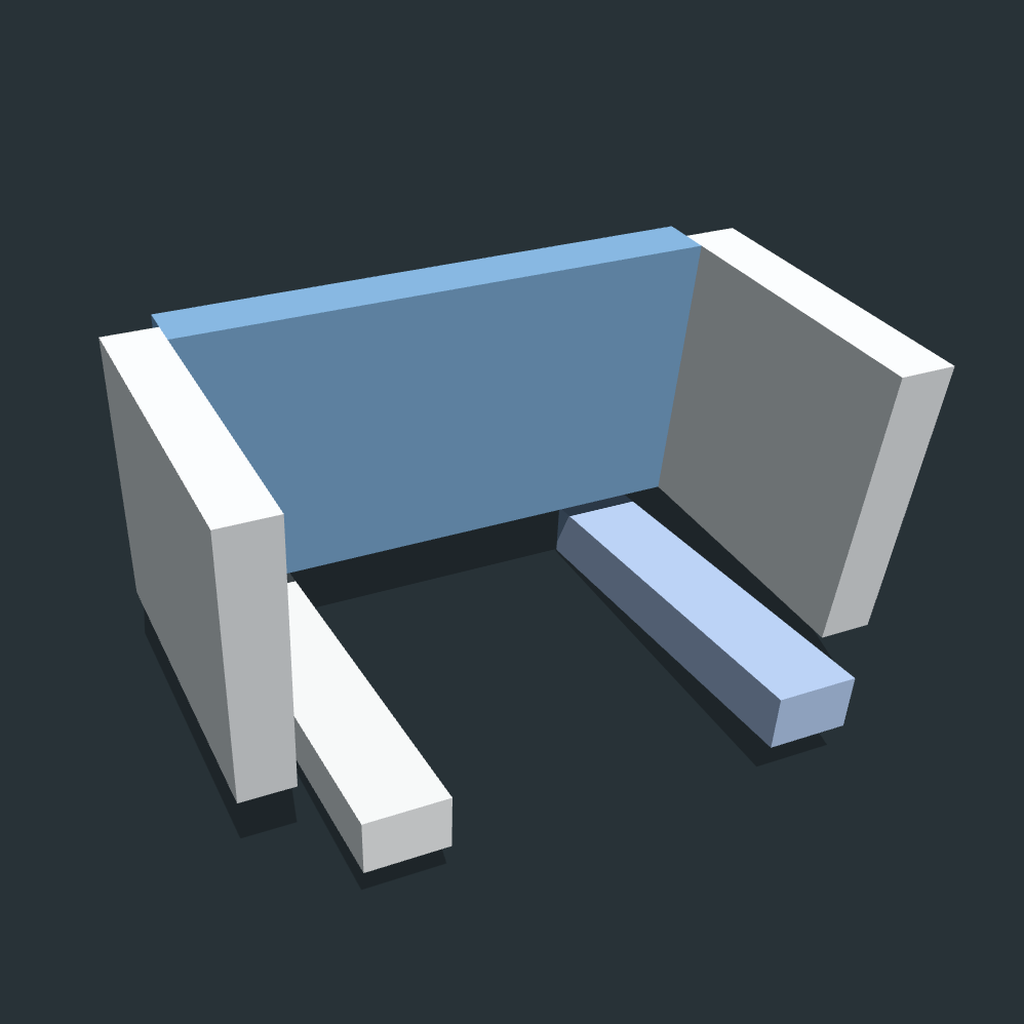} &
        \includegraphics[width=\linewidth]{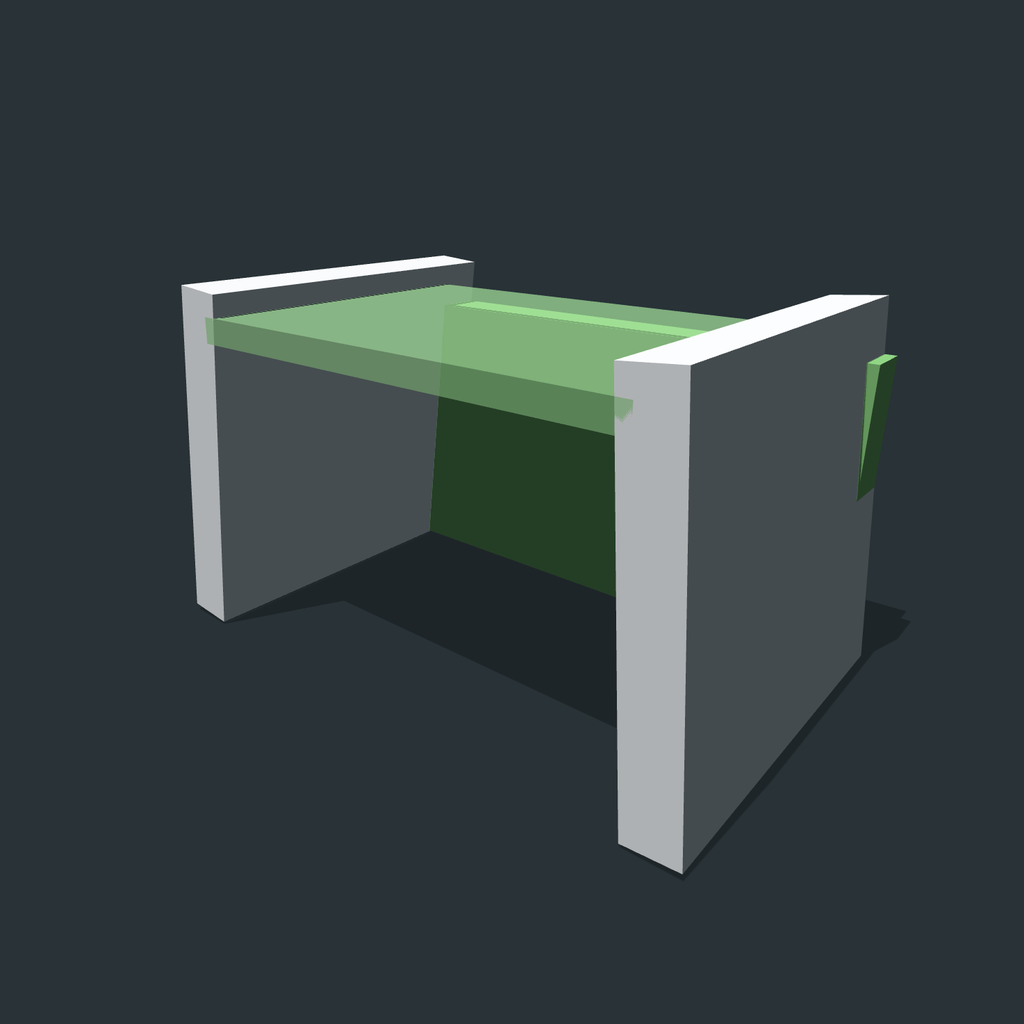} \\
      \end{tabular}
    \\

    \bottomrule
  \end{tabular}
  \caption{Tool-gallery for GatherSpheres and MoveBall.}
  \label{tab:tool_gallery_part2}
\end{table*}

\subsection{Licenses}
The cardboard box asset used in \textsc{LiftBox} environment is from PartNet-Mobility Dataset~\cite{Xiang_2020_SAPIEN}. Their terms of use are stated here: \href{https://sapien.ucsd.edu/about}{sapien.ucsd.edu/about}.

The book assets and the book holder used in the \textsc{OneBook} environment came from the YCB Dataset~\cite{Calli2015-ju}. This dataset is under the CC BY 4.0 license.

The goal frame and net assets used in the \textsc{ScoreGoal} environment came from the Meta-World Benchmark~\cite{yu2019meta}. This benchmark is under the MIT License.

The transparent jar asset used in the \textsc{SnatchCookie} environment came from \href{https://www.cgtrader.com/free-3d-models/household/kitchenware/glass-mason-jar-16oz}{cgtrader}, a 3D CAD model website. This asset is under the "Royalty Free No Ai License", detailed \href{https://www.cgtrader.com/pages/terms-and-conditions#royalty-free-license}{here}.

The cookie assets used in the \textsc{SnatchCookie} environment came from \href{https://sketchfab.com/3d-models/cookie-778c9c225d904e60b890cc43875a7aad}{sketchfab}, a 3D CAD model website. This asset is under the CC BY 4.0 license.

The turkey leg assets used in the \textsc{TurkeyLegs} environment came from \href{https://sketchfab.com/3d-models/turkey-leg-3c889b5ee6b64a2aafb9b8977f8a8219}{sketchfab}, a 3D CAD model website. This asset is under the CC BY 4.0 license.

\end{document}